\documentclass[a4paper,12pt]{book} 

\usepackage{algorithm}
\usepackage{algpseudocode}

\usepackage{amsmath, amsthm, amssymb, amsfonts}
\usepackage{etoolbox}
\usepackage{comment}
\usepackage{booktabs}
\usepackage{scalerel}

\usepackage{mathtools}
\usepackage{xcolor}
\definecolor{tableShade}{gray}{0.9}
\definecolor{lightblue}{RGB}{230,247,254}
\definecolor{lightgreen}{RGB}{222,251,152}
\definecolor{lightorange}{RGB}{254,239,224}
\usepackage{colortbl}
  \usepackage{multirow}
  \usepackage{graphicx}
  \usepackage{soul,color,url}  

  \usepackage{bbding}
  \usepackage{colortbl}
  \usepackage{bm} 
  \usepackage{multirow}
  \usepackage{subcaption}
  \usepackage{listings}             
  \usepackage{wrapfig}
\usepackage{physics}
  \usepackage{framed}
\definecolor{shadecolor}{RGB}{230,247,254} 

  \newenvironment{SnugshadeB}[1][230,247,254]{
  \begin{snugshade}%
}{%
    \end{snugshade}%
}

\newcounter{MethodF}
{\end{upshape} \par}

  \newcounter{MethodB}
\newenvironment{MethodB}[1][]{\refstepcounter{MethodB}\par\noindent\textbf{Model \theMethodB: #1\\}\begin{upshape}}
{\end{upshape} \par}
\newtheorem{theorem}{Theorem}[chapter]
\newtheorem{proposition}[theorem]{Proposition}
\newtheorem{lemma}[theorem]{Lemma}
\theoremstyle{definition}
\newtheorem{definition}[theorem]{Definition}
\newtheorem{remark}[theorem]{Remark}
\newtheorem{example}[theorem]{Example}
 
\newcommand{\bx}{{\bf x}}
\newcommand{\by}{{\bf y}}
\newcommand{\bz}{{\bf z}}
\newcommand{\Qset}{\mathcal{Q}}

\newcommand{\bxi}{\bm{\xi}}
\newcommand{\bgamma}{\bm{\gamma}}

\AtEndEnvironment{theorem}{\qed}
\AtEndEnvironment{proposition}{\qed}
\AtEndEnvironment{definition}{\qed}
\AtEndEnvironment{lemma}{\qed}
\AtEndEnvironment{remark}{\qed}
\AtEndEnvironment{example}{\qed}

\usepackage{natbib}

\title{A tutorial on learning from preferences and choices with Gaussian Processes}

\author{
    Alessio Benavoli\thanks{School of Computer Science and Statistics, Trinity College Dublin, Dublin, Ireland; \texttt{alessio.benavoli@tcd.ie}}
    \and
    Dario Azzimonti\thanks{SUPSI, Dalle Molle Institute for Artificial Intelligence (IDSIA), Lugano, Switzerland; \texttt{dario.azzimonti@idsia.ch}}
}

\date{2026\\~\\~\\~\\
\begin{quotation}
\noindent \textbf{Suggested Citation:} \\
Benavoli A, Azzimonti D (2026), ``A tutorial on learning from preferences and choices with Gaussian processes''. Foundations and Trends in Machine Learning, Vol. 19 No. 1 pp. 1–120, doi: \url{https://doi.org/10.1108/FTMAL-03-2025-0124}
\end{quotation}
} 

\begin{document}

\maketitle

\chapter*{Abstract}
Preference modelling  lies at the intersection of  economics, decision theory, machine learning and statistics. By understanding individuals' preferences and how they make choices, we can build  products  that closely match their expectations, paving the way for more efficient and personalised applications across a wide range of domains. The objective of this tutorial is to present a cohesive and comprehensive framework for preference learning with Gaussian Processes (GPs), demonstrating how to seamlessly incorporate rationality principles (from economics and decision theory) into the learning process. By suitably tailoring the likelihood function, this framework enables the construction of preference learning models that encompass random utility models, limits of discernment, and scenarios with multiple conflicting utilities for both object- and label-preference. This tutorial starts by reviewing the mathematical concept of preference, its properties and the representation via utility functions. We then organise preferential data into three types: object preferences, label preferences and choices. We present five GP-models for object preferences showing in which situations each one is better adapted. For label preferences we present three models based on different noise assumptions. Finally for choice data we present a model for rational choice functions and one for pseudo-rational choice functions. The last part of the tutorial presents five applications of GP-based preference learning. This tutorial builds upon established research while simultaneously introducing some novel GP-based models to address specific gaps in the existing literature.


\chapter{Introduction}
\label{sec:intro}
The objective of preference learning \citep{furnkranz2010preference} is to use a statistical model to learn and  predict human preferences, conditioned on data.  Unlike regression or classification, where the target variable is a scalar, preference data is in the form of pairwise comparisons, which express a  subject's preference between alternative options.

Depending on the nature of these options, we distinguish  two  cases: \textit{object preferences} and \textit{label preferences} \citep{furnkranz2010preference}. In \textit{object preferences}, the training data consists of pairwise comparisons among  the characteristics or attributes associated with the objects. Example:

\begin{itemize}
    \item \textit{Food preference}: the goal is to learn user preferences among various foods by analysing preference data such as `I prefer food A over food B'. Implicitly, the user is indicating preferences for specific attributes of the food, such as ingredients and cooking methods. The model learns from these preferences and can then suggest similar dishes to the user based on attributes (ingredients and cooking methods) associated with those dishes.
\end{itemize}

In  \textit{label preferences}, the training data consists of pairwise comparisons among labels associated with the objects. Example:

\begin{itemize}
    \item \textit{Travel preference:} the objective is to gain insights and to predict user preferences for different modes of transportation (in general referred to as labels) such as cars, trains, buses, and bicycles, based on data collected in the form `I prefer using a train to go from A to B'. The attributes considered in this context pertain to the user: age, education, and occupation, for example. The aim is to predict transportation modes for new users.
\end{itemize}
There are also applications where we encounter mixed-type problems, exhibiting characteristics of both label and object preference. We will provide an example in the application section of this tutorial.

There are two main approaches for learning preferences: 
\begin{enumerate}
\itemsep0em 
    \item \textit{utility function} based; 
    \item \textit{two-argument function} based. 
\end{enumerate}

\section{Utility function based learning.}
This approach exploits a fundamental result in economics \citep{debreu1954representation}:  a  preference relation that satisfies certain \textit{rationality} properties is representable by a  \textit{utility function} (and vice versa), see Section~\ref{sec:pref} for a review of these results. This means that, when presented with various alternatives, a \textit{rational} subject tends to prefer the option with the highest utility. In preference learning, the utility function is latent (not directly observable) and the objective is to learn this latent function based on the observed subject's preferences. Then, we can use the inferred utility function to predict the subject's preference among options by choosing the alternative with the highest expected utility.

However, in real-world scenarios, individuals often deviate from \textit{rationality} for different reasons. 
To account for this behaviour, one modelling approach is to employ \textit{random utility models} \citep{mcfadden1974,mcfadden1978}. These models  assume that a subject's preference is determined by a noisy utility function. The diversity of random utility models arises from different assumptions about the distribution of this noise. Examples of proposed distributions for the noise are Gaussian \citep{thurstone2017law} and Gumbel \citep{luce2012individual}. There are other reasons why individuals can deviate from rationality, such as, for instance, a \textit{limit of discernibility} \citep{luce1956semiorders} between options having close utilities as well as the presence of \textit{multiple conflicting utilities} \citep{moulin1985choice}. 
In this tutorial, we will review and implement various learning models designed to address the aforementioned source of irrationality in preferences.

Finally, we must mention that in many applications we do not observe preferences, but only the choices of a subject, that is the options the subject has selected among the given alternatives. For example, in e-commerce, we may only observe which item a user clicks on among several displayed options, without knowing their full preference ordering. Therefore, we must also be able to \textit{learn from choice data}. Under some \textit{rationality} assumptions in the way individuals make choices, it is still possible to learn the underlying utility function(s) which determines the subject's choices. In economics, stating  the conditions of rationality that make this possible is referred to as the   \textit{revealed preference theory} \citep{samuelson1938note}.

\section{Preference learning with two-argument functions.} 
{An alternative approach to learning preferences is to model the preference relation directly as a two-argument function, rather than learning a utility function over individual options. In this framework, instead of assigning a numerical value (utility) to each option and comparing those values, the model learns a function that takes two options as input and directly predicts which one is preferred.}
Expressing preferences through two-argument functions has been studied  in economics to model non-transitive preferences, see, for instance, \citet{shafer1974nontransitive,fishburn1988nonlinear}. 
In machine learning, this is the predominant method \citep{furnkranz2010preference} for preference learning, because it allows the practitioner to frame the problem as a classification task. This approach \citep{van2022choice} works without imposing specific model assumptions, relying solely on a data-driven methodology. However, the effectiveness of this method in learning properties like (quasi-)transitivity from data hinges on a substantial amount of data, making it particularly well-suited for applications with large datasets \citep{wong2020bi,SIFRINGER2020236,LEDERREY2021100226}.

\section{Objective of this tutorial}
Our goal is to establish a  modelling framework for preference learning that can integrate utility-based models, random utility models, limits of discernibility, and scenarios involving multiple conflicting utilities. We aim to achieve this by introducing a Gaussian Process (GP) based approach (GPs are defined in Section \ref{sec:GP}) for both object and label preference learning based on the  existing literature \citep{ChuGhahramani_preference2005,houlsby2011bayesian,chau2022learning,nguyen2021top,benavoli2023a}, as well as developing novel GP-based models  to address specific gaps in this literature. 
The presented models come with a Python library \textit{prefGP} \citep{prefGP}, developed for this tutorial, enabling practitioners to seamlessly apply these models to their data, facilitating comparison and analysis. 

\subsection{Why Gaussian processes?}

GPs are a class of nonparametric, kernel-based models used for learning functions from data. Instead of assuming a fixed parametric form for the function (e.g., linear or neural network-based), GPs define a distribution over functions, where any finite set of function values follows a joint Gaussian distribution. The choice of the covariance (kernel) function of the GP encodes assumptions about the smoothness and structure of the underlying function, allowing GPs to flexibly model complex relationships with principled uncertainty estimates \citep{rasmussen2006gaussian}.  We provide a brief introduction to GPs in Section~\ref{sec:GP}. \emph{Why use GPs for preference learning?}
First, in comparison to traditional statistics literature on preference learning, GPs enable us to move beyond the assumption of a linear utility function in the covariates. In contrast to the machine learning literature, GPs allow us to directly model the utility function without the need to parameterise it, for instance, through a neural network. This simplifies both the modelling treatment and the model inference process. Second,  for several preference models, GPs offer an efficient (rejection-free) method  to sample from the posterior. Third, we seek a model that can automatically adjust its own complexity to the data -- a crucial aspect in preference learning for achieving `exact interpolation', essential for modelling the utility function of a rational subject. Fourth,  a kernel-based approach gives greater versatility in handling various types of data; for instance, it allows the definition of kernels on images and graphs as well as defining specific preference kernels \citep{houlsby2011bayesian,lomeli2019antithetic,chau2022learning}.
Finally, GPs provide an estimate of their own uncertainty, a valuable aspect in decision-making. For example, it can be employed for active preference learning \citep{eric2007active,zoghi2014relative} and Preferential Bayesian optimisation \citep{shahriari2015taking,gonzalez2017preferential,siivola2021preferential,benavoli2021preferential,nguyen2021top,benavoli2023d,xu2024principled,takeno2023towards, Adachi_etal_2024_LoopExplain,zhang2021bayesian}.

It is important to note that our modelling approach involves specifying different likelihoods, while using the same prior, a GP. Hence, this tutorial is also valuable for practitioners who aim to model preferences using the same likelihoods (cost functions) while employing alternative machine learning models to fit the utility.

We will present and derive \text{9  models} to learn from preference and choice data:
\begin{description}
\itemsep0em 
    \item[Model \ref{Model:1}:] Consistent Preferences.
    \item[Model \ref{Model:JND}:] Just Noticeable Difference.
    \item[Model \ref{Model:Probit}:] Probit for Erroneous Preferences.
        \item[Model \ref{model:Gaussian}:] Preferences with Gaussian Noise Error.
    \item[Model \ref{model:Probitclassification}:] Probit  for Erroneous Preferences as a Classification Problem.
    \item[Model \ref{model:thurston}:] Thurstonian Model for Label Preferences.
    \item[Model \ref{model:PL}:] Plackett-Luce Model for Label Ordering Data.
    \item[Model \ref{model:pairwiselabel}:] Paired Comparison for Label Preferences.
        \item[Model \ref{model:choice}:] Rational and Pseudo-rational Models for Choices.
\end{description}

\section{Literature overview}
The literature in preference learning is vast. The following books provide an overview of the main models and challenges: \cite{marden1996analyzing, train2009discrete,furnkranz2010preference,alvo2014statistical}.

The most popular \textit{Random Utility Models} (RUMs) for preference learning are
variants of the Bradley-Terry model \citep{bradley1952rank}, Babington-Smith model \citep{smith1950discussion,mallows1957non},  and  Thurstone-Mosteller model \citep{thurstone2017law,mosteller1951experimental}. These are probabilistic models used to describe the outcome of pairwise comparisons between objects. The underlying probabilistic models (likelihoods) can be derived assuming the pairwise comparisons are noisy. They assume either  Gumbel noise, resulting in a \textit{logit}  likelihood, or  Gaussian noise, giving rise to a \textit{probit} likelihood. Traditionally, these models have been designed for label-preference and they are fitted via \textit{maximum likelihood estimation}
using different methods 
\citep{hunter2004mm,shah2015estimation,maystre2015fast,vojnovic2016parameter,ragain2016pairwise,agarwal2018accelerated}. Bayesian models and inference are instead proposed by  \cite{guiver2009bayesian,caron2012efficient,azari2012random}.

The extensions of the above RUMs to include covariates  are known as  \textit{multinomial logit models} \citep{mcfadden1974,mcfadden1978} and  \textit{multinomial probit models}. The main difference between these two models lies in their treatment of the noise. The first model assumes independent noise, whereas the second model assumes dependent noise, specifically in the form of Gaussian correlated noise. \textit{Apollo} \citep{hess2019apollo} and \textit{mlogit} \citep{croissant2020estimation} are R packages which allow to fit multinomial logit/probit models using maximum likelihood. \textit{Biogeme} \citep{bierlaire2018pandasbiogeme}, \textit{Pylogit} \citep{brathwaite2018asymmetric}, \textit{torch-choice} \citep{du2023torch} are similar Python-based libraries.

These models have been extended within the framework of generalised linear models  \citep{critchlow1991paired}, generalised additive models with splines \citep{abe1999generalized,kneib2007semiparametric} and,   support vector machines \citep{evgeniou2005generalized,maldonado2015advanced}. 

A Bayesian nonparametric approach based on \textit{Gaussian Processes} for preference learning was first proposed by \citet{ChuGhahramani_preference2005} and by \citet{houlsby2011bayesian} for both object- and label-preference with a \textit{probit} likelihood. This approach offers two advantages: a nonlinear utility in the covariates and the representation of uncertainty through the posterior. Since the posterior is not Gaussian, \citet{ChuGhahramani_preference2005} proposed the Laplace's approximation for inference. Other approximations were considered by \citet{houlsby2011bayesian}. Recently, \citet{benavoli2021preferential,benavoli2021} showed that the posterior is a Skew GP and employed a rejection-free procedure to efficiently sample from the posterior.  

Apart from random utilities,  there are other reasons why individuals can deviate from rationality, such as for instance a \textit{limit of discernibility} \citep{luce1956semiorders} between options having close utilities. 
Furthermore, deviations may arise in cases where the subject's preferences are determined by \textit{multiple utility functions}. Rationalising preferences and, more in general, choices based on multiple utilities has been extensively studied in economics  \citep{moulin1985choice,eliaz2006indifference}. An extension of logit-based RUM to deal with multiple utilities has been proposed in \citet{benson2018discrete}. A more general model derived to learn choice functions  via multiple utilities (through a Pareto embedding) was  derived in  \citet{pfannschmidt2020learning},  using a hinge-loss and a neural network based model.  A GP and probit version of the above model was formulated  in \citet{benavoli2023a}.
 
The approach to preference learning not based on utility functions aims to directly learn a \textit{two-argument function}. This problem can be formulated as an augmented binary classification
problem or constrained classification (SVM), see, e.g., \citep{cohen1997learning,herbrich1998learning, aiolli2004learning,har2002constraint,fiechter2000learning,hullermeier2008label,furnkranz2010preference}.
For Gaussian Processes, \citet{houlsby2011bayesian} derived the theoretical  link between the GP-based probit model for rational preference learning based on RUM and the two-argument function. Indeed, a GP prior on the latent utility  induces a GP prior on the two-argument function  by linearity, see Section~\ref{sec:twoargument} for more details. This allowed \citet{houlsby2011bayesian} to derive the so called \textit{preference kernel}.  \citet{pahikkala2010learning} instead, derived the kernel of a two-argument function by using a feature map view, which allows modelling preferences that satisfy asymmetry  but not  transitivity in general (we will provide a definition of asymmetry and transitivity in Section \ref{sec:preliminary}). This kernel is known as  \textit{intransitive preference kernel}. A preference-learning model which employs a GP prior  based on this intransitive kernel was proposed by \citet{chau2022learning}. \citet{Hu_etal_2022_PrefShapley} used this approach to propose a Shapley-based explainability algorithm for preferences.

In this tutorial, we will not cover the discussion of techniques for deducing the consensus preference within a group of subjects or about aggregation methods. It is worth noting that certain methods we will introduce in this tutorial may be applicable to such scenarios, such as interpreting multiple utilities as arising from distinct subjects (multi-self). Nonetheless, it is important to stress that the aggregation problem has been extensively investigated in the literature, see, for instance, \citep{arrow2012social,fligner1990posterior,Meila_Phadnis_Patterson_Bilmes_2012,negahban2012iterative,azari2013generalized,Caragiannis_Procaccia_Shah_2016}.
A related problem to aggregation is ranking and learning from ranking data,  in this case the input-data is a complete or partial ordering of labels. For example, a partial ordering appears when users are asked to order (rank) their top-10 movies or the result of a web-search.  A linear RUM used to learn from ordering data is the \textit{Plackett-Luce} model \citep{luce2012individual,plackett1975analysis}. There have been many models developed for \textit{recommendation systems} \citep{ricci2015recommender}, for instance based on neural networks (``learning to rank'', \citet{burges2005learning,burges2010ranknet,cao2007learning}). A GP-based, nonlinear extension of  Plackett-Luce was derived by \citet{nguyen2021top} and we will discuss it in this tutorial.

It is important to highlight that preference learning is accompanied by its own computational challenges. The potential number of preferences among $r$ objects is  $r(r-1)/2$ and it can become large for big datasets. In the case of choice data, computational complexity can increase factorially based on the choice-set's size, depending on the specific problem and the subject's decisions. In label-preference, the number of utilities is equal to the number of labels, leading to large number of utilities  in many applications. 
This implies that computational considerations must also be taken into account when choosing a model. Related to this, an important area of research pertains to the elicitation of preferences, see, e.g., \citep{boutilier2006constraint,boutilier2013computational,bourdache2019incremental,adam2021possibilistic,adam2021incremental}.

Finally, in this tutorial, we will not cover   \textit{preference-based reinforcement learning}, a framework that enables learning from non-numerical feedback in sequential domains. Compared to standard reinforcement learning, preference-based learning offers the  advantage of defining feedback that is independent of arbitrary reward choices/shaping/engineering \citep{wirth2017survey}. This approach  has recently found applications in training \textit{large language models} \citep{christiano2017deep,stiennon2020learning,ouyang2022training,Rafailov2023llmDPO}. Note that the techniques used for preference modeling are similar to the models discussed in this tutorial for object preference. The main difference is that, in preference-based reinforcement learning, the objects are trajectories  actions-states. These approaches are also known as \textit{Reinforcement Learning from Human Feedback} (RLHF).
In this dynamic context, a crucial consideration that has not received sufficient attention is the extension of consistency principles for preferences over time  \citep{troffaes2013note}.

\section{Tutorial organisation}
Chapter~\ref{sec:preliminary}  introduces the mathematical objects and concepts that will be used in the rest of the tutorial. Chapter~\ref{sec:objpref} will focus on object preference. In particular, a model to learn from consistent preference is introduced in Section~\ref{sec:consistent}. We will then modify this model to account for possible sources of irrationality: limit of discernibility in Section~\ref{sec:discern}; additive Gaussian noise in Section~\ref{sec:gaussiannoise} and multiple conflicting utilities in Section~\ref{sec:multipleu}. All the above models aim to learn the utility function of the subject. Section~\ref{sec:twoargument} delves into directly learning the preference relation as a two-argument function.
We will then move to label-preference in Chapter~\ref{sec:labelpref} discussing two RUMs based on Gaussian noise in Section~\ref{sec:thurstone} and Gumbel noise in Section~\ref{sec:PL}. We then present a model for label-preference learning based on pairwise comparisons in Section~\ref{sec:labelPW}. We will then move to choice functions in Chapter~\ref{sec:choice} introducing two models in Section~\ref{sec:chocieslearning}. Chapter~\ref{sec:applications} will discuss  applications. Finally, Chapter~\ref{sec:Conclusions} concludes the tutorial.

\chapter{Preliminaries}
\label{sec:preliminary}
 In this section, we introduce the main mathematical objects and concepts that we will use in the next sections. We start by defining preferences, utility and choice functions. We will then move to Gaussian Processes, which is the statistical model we will use to learn from preference and choice data.

\section{Preferences and Utilities}
\label{sec:pref}
Consider a finite set $\mathcal{X}$ and a binary relation $R$ on $\mathcal{X}$ expressed by a subject. $R$ is a subset of $\mathcal{X} \times \mathcal{X}$. For any two elements of $\mathcal{X}$, the statement $(x, y) \in R$, denoted by $xRy$, tells us that, according to the subject, `x is R-related to y' in some way.

The following is a list of common properties of binary relations.

\begin{itemize}
\itemsep0.1em 
\item \textit{Reflexive}: $\forall x \in \mathcal{X}$ $x R  x$;
\item \textit{Irreflexive}: $\forall x \in \mathcal{X}$ $\neg x R  x$, where $\neg$ stands for logical negation;
\item \textit{Symmetric}: $\forall x,y \in \mathcal{X}$ if $x R  y$ then $y R  x$;
\item \textit{Asymmetric}: $\forall x,y \in \mathcal{X}$ if $x R  y$ then not $y R x$;
\item \textit{Negatively transitive}: if $x R y$ then for any other element $z \in \mathcal{X}$ either $x R z$ or $z R y$ or both.
\item \textit{Transitive}: $\forall x,y,z \in \mathcal{X}$ if $x R y$ and $y R z$ then $x R z$.
\item \textit{Acyclicity}: If, for any finite number $n$, $x_1 R x_2$, $x_2 R x_3$, $\dots$, $x_{n-1} R x_n$ then $x_n \neq x_1$.
\item \textit{Completeness}: $\forall x,y \in \mathcal{X}$ either $x R y$ or $y R  x$ or both.
\end{itemize}
Preferences are particular types of relations and they are for instance used to model a consumer. Given two objects, Alice (the subject/consumer) may say that ``$x$ is better than $y$'' and, in this case, we write the relation $x R y$ as $x \succ y$, read as $x$  is strictly preferred to $y$. 

\begin{definition}
A strict preference is a binary relation, denoted by $\succ$, which is asymmetric and negatively transitive. 
\end{definition}

The asymmetry implies  that the subject cannot say that $x$ is better than $y$ and also that $y$ is better than $x$. The property \textit{negatively transitive} is clearer when formulated as ``if $x \nsucc z$ and $z \nsucc y$ then $x \nsucc y$'', which means that if $x$ is not better than $z$ and $z$ not better than $y$ then $x$ is not better than $y$.  A strict preference relation is said to be \textit{consistent} -- Alice is  \textit{rational} -- when it satisfies the above two properties.

\begin{proposition}(\citet[Prop.~2.1]{kreps1990course})
Any strict preference is \textit{irreflexive, transitive and acyclic}. 
\end{proposition}
We point the reader to \citet[Ch.\ 2]{kreps1990course} for the proofs of the results in this section. From a strict preference, $\succ$, we can define two new relations called \textit{weak preference} and, respectively, \textit{indifference}.

\begin{definition}
For each $x,y \in \mathcal{X}$, $x \succeq y$ (read as $x$ weakly preferred to $y$), if it is not the case that $y \succ x$.
For each $x,y \in \mathcal{X}$, $x \sim y$ (read as $x$ is indifferent to $y$), if it is not the case that either $x \succ y$ or $y \succ x$.
\end{definition}

\begin{proposition}(\citet[Prop.~2.2]{kreps1990course})
The above weak preference relation is complete and transitive.  The above indifference relation is reflexive, symmetric and transitive.
\end{proposition}

It is worth noticing that:
\begin{proposition}
\label{prop:completeasym}
~
\begin{enumerate}
\itemsep0em 
    \item $\succ$ is asymmetric iff $\succeq$ is complete;
    \item $\succ$ is negatively transitive iff $\succeq$ is transitive.
\end{enumerate}
\end{proposition}

\section{Representation via utility function}
Strict preferences can be represented by a \textit{value function}. We refer to a \textit{value function} that represents
preferences as a \textit{utility function}.

\begin{definition}
For any set $\mathcal{X}$ and preference relation $\succ$ (resp. $\succeq$) on $\mathcal{X}$, the function $u : \mathcal{X} \rightarrow \mathbb{R}$
represents $\succ$ (resp. $\succeq$) if
$$
x \succ y ~~\textit{ iff }~~ u (x) > u (y) ~~~~\textit{(resp. $x \succeq y ~~\textit{ iff }~~ u (x) \geq u (y)$)}~~ \forall x,y \in \mathcal{X}
$$
We say that $u$ is a utility function for $\succ$ (resp. $\succeq$).
\end{definition}
This means that, when presented with various alternatives, a rational subject tends to prefer the option with the highest utility. When does $\succ$ admit a utility function representation? 
If $\mathcal{X}$ is finite, $\succ$  admits a utility function representation iff it is asymmetric and negatively transitive \citep[Ch.\ 2]{kreps1990course}.\footnote{When $\mathcal{X}$ is infinite, we need additional topological requirements for $\mathcal{X}$ to prove a similar result \citep{debreu1954representation}.} However, this representation is not unique. Indeed, utility functions are invariant under increasing transformations. We can define a new utility function $g(u(x))$ for any  increasing function $g$, that is, $\forall x,y \in \mathcal{X}$ we have
$$
x \succ y ~\textit{ iff }~ u (x) > u (y) ~\textit{ iff }~g(u (x)) > g(u (y)).
$$
So $u(\cdot)$ and $g(u(\cdot))$ represent the same strict preference relation. What this means is that we treat utility functions as ordinal rather than cardinal. That is, only the relative ranking of two items is important, not the magnitude of their utility difference. The reader should keep this in mind for the next sections.

\section{Choice functions}
\label{sec:choicefunintro}
In the previous sections, we have assumed that Alice expresses her judgements  as preferences. 
 However, we often only observe her choices. For instance, given a set of products on an internet commerce site, she may decide to click on/buy some of them.
 
Alice's choices can be formalised through the concept of choice functions.  Let  $\Qset$ denote the set of all  finite subsets of $\mathcal{X}$.\footnote{We assume that $\mathcal{X}$  is finite.}
\begin{definition}[\cite{kreps1990course}]
 A choice function $C$ is a set-valued operator on sets of objects. More precisely, it is a map $C: \Qset \rightarrow \Qset$ such that, for any set of objects $A \in \Qset$, the corresponding value of $C$ is a subset $C(A)$ of $A$.
\end{definition}
It will be assumed throughout this paper that Alice is
able to find a choosable option in every set she is presented with, and therefore $C(A) \neq \emptyset$ for all $A$. It is convenient to introduce the set of rejected options, denoted by $R(A)$, and  equal to $A\backslash C(A)$.

In other words, given a set of objects $A = \{o_1, o_2, \dots, o_n\}$, we denote by $C(A)$ the subset of objects that Alice chooses, and by $R(A)=A\backslash C(A)$ the subset she rejects. Alice's choices are mathematically represented by the choice function $C$.

There are two main interpretations of choice functions. 
In both interpretations, for a given option set $A \in \Qset$, the statement that an object $o_j \in A$ is rejected from $A$ (that is, $o_j  \notin C(A)$) means that there is at least one object  $o_i \in A$ that a subject  prefers over $o_j$. Alice is not required to tell us which object(s) in $C(A)$ she  prefers to $o_j$. This makes choice functions a very easy-to-use tool to express choices. 

The two interpretations differ instead in the meaning of the statement  $o_i  \in C(A)$.
\begin{enumerate}
    \item In the traditional interpretation, one reads the statement $o_i  \in C(A)$ as ``$o_i$ is considered to be at least as good as all other options in $A$,” and thus infers from a statement like $\{o_i, o_j\} \subseteq C(A)$ that Alice is indifferent between $o_i$ and $o_j$.
    \item The alternative and more general interpretation of  $\{o_i, o_j\} \subseteq C(A)$ is that $o_i$ and $o_j$ are incomparable for Alice. Equivalently, $\{o_i, o_j\} \subseteq C(A)$ should be read as ``$o_i$ and $o_j$ are undominated in $A$'' in a sense that will be clarified hereafter.
\end{enumerate}

\subsection{Choice functions and utilities}
In Section \ref{sec:pref}, we have discussed the properties that a preference relation should have to be considered consistent. There are three important properties for choice functions, see  \citep{chernoff1954rational,sen1971choice,aizerman1981general,moulin1985choice,Schwartz1986} for a detailed exposition.

The first property states that if an object  is selected from a set of alternatives $A$, it should also be selected in any subset $A'$ that contains it.

\begin{description}
 \item[Chernoff:] if $A' \subseteq A$ then $(C(A) \cap A')\subseteq C(A')$.
\end{description}
The second property states that every object that is selected from two different sets $A,A'$ should also be selected from their union $A \cup A'$.
\begin{description}
 \item[Expansion:] $\forall A,A'$,~ $(C(A) \cap C(A'))\subseteq C(A \cup A')$.
\end{description}
The third property states that the subset of
objects chosen from a set cannot expand when non selected
objects are removed from that set.
\begin{description}
 \item[Aizerman:] $\forall A,A'$, if $C(A) \subseteq A' \subseteq A$ then $C(A') \subseteq C(A)$.
\end{description}
Chernoff and Expansion are also known as $\alpha$ and $\gamma$ consistency rules. Similarly to what was done for preferences, we can ask if a consistent choice function defines or is defined by some utility functions. We first introduce to concept of \textit{Pareto
rationalisability}.

\begin{definition}[Pareto
rationalisability]
Given a collection of strict orders  $\{\succ_k\}_{k=1}^d$, they are said to Pareto rationalise $C$ when the
following condition holds:
\begin{description}
 \item[]
$\forall A \in \mathcal{Q}$, $\forall o_i \in A$, $o_i \in C (A)$ if and
only $\nexists o_j \in A$ such that $o_j \succ_k o_i$ $\forall k\in\{1,2\dots,d\}$.
\end{description}
\end{definition}
Pareto Rationalisability means that only the objects that are not dominated by any other objects in the same set are chosen.

\begin{proposition}[\cite{moulin1985choice}]
\label{prop:moulin}
 Let $C: \mathcal{Q} \rightarrow \mathcal{Q}$ be a choice function. There exists a set of strict preferences $\{\succ_k\}_{k=1}^d$ that Pareto
rationalises $C$ if and only if $C$ satisfies  Chernoff, Expansion and Aizerman.
\end{proposition}
As discussed in Section \ref{sec:pref}, any strict order $\succ_i$ can be represented by a utility function $u_i$ and vice versa.
Proposition \ref{prop:moulin} tells us  that there is a latent vector of utility functions ${\bf u}(o)=[u_1(o),\dots,u_{d}(o)]^\top$, for some finite dimension $d$, which represents the choice function $C$ if and only if $C$ satisfies  Chernoff, Expansion and Aizerman. ${\bf u}$ embeds the options $o$ into a  space $\mathbb{R}^{d}$ and the choice set $C(A)$ expresses a Pareto set of strongly undominated options with respect the utilities ${\bf u}$.\footnote{It is actually possible to prove Proposition \ref{prop:moulin} even assuming weak-orders $\{\succeq_k\}_{k=1}^d$ \citep{arlegi2022pareto}, which leads to a weak-Pareto form of rationalisability.}

We may question if Chernoff, Expansion and Aizerman are always desirable properties. Thinking in terms of a vector of utilities, we may think of situations where Alice, in case of options with conflicting quality attributes (utilities), might pick those options that are optimal for at least one of these attributes. Then, this way of choosing is called pseudo-rational.
\begin{proposition}[\cite{moulin1985choice}]
\label{prop:moulin2}
A choice function is pseudo-rationalisable if and only if it satisfies Chernoff and Aizerman.
\end{proposition}
 Pseudo-rationalisability  is a  weaker property than Pareto 
rationalisability. It represents
 a choice function in the form 
 \begin{equation}
     \label{eq:pseudoratio}
    C(A)= \bigcup\limits_{k=1,\dots,d} \arg\max_{o \in A} u_k(o),
 \end{equation}
 and, therefore, it requires less rationality (less
consistency of the choice sets) than representing it in the Pareto
sense.\footnote{It is worth mentioning that Chernoff and Expansion characterise
rationalisability by an acyclic relation \citep{moulin1985choice}, which we will not further explore in this paper.}

We point the reader to \citet{kreps1990course,moulin1985choice} for the connection between preferences and choice functions.

Representing rational and pseudo-rational choice functions through utility functions is important because it allows us to learn choice behavior by learning the underlying utility functions. This is particularly useful, as learning functions from observed choice data is a well-established task in machine learning and statistics.

\section{Gaussian process}
\label{sec:GP}
 As we will see in the next sections, there are two main ways to learn from preference data. We can learn either   the preference-relation (the binary-relation) or the utility function that represents it.
Since a binary-relation is a function that assigns truth values (true or false) to two elements of $\mathcal{X}$, in both cases learning preferences reduces to learning a function.

An elegant way to learn functions is by defining a prior probability over functions and then compute  the posterior distribution  given the data. 
Gaussian Processes (GPs) are prior over functions \citep{o1978curve,rasmussen2006gaussian}, that  have attractive advantages over parametric (including neural networks) models.\footnote{GPs can be seen as single-layer neural networks  with an infinite number of hidden units \citep{williams1996computing}.} They are   nonparametric, meaning that their complexity automatically grows as more data are observed and, therefore, they have the ability  to exactly `interpolate' the data, which is important to model \textit{consistent} preference relations. 

They have a small number of tunable hyperparameters (and so they can be trained on  small  datasets), and naturally provide an uncertainty measure on their predictions. Moreover, by being kernel based, they provide a framework to learn utility functions defined on any domain $\mathcal{X}$ on which we can define a kernel function. 

To define a prior over a function $u: \mathcal{X} \rightarrow \mathbb{R}$, a GP assumes that, for every $n$,  the distribution of the vector $(u({\bf x}_1),\dots, u({\bf x}_n))$ is jointly Gaussian, with mean  $[\mu({\bf x}_1),\dots, \mu({\bf x}_n)]$ and covariance  $Cov(u({\bf x}_i),u({\bf x}_j)) = k({\bf x}_i,{\bf x}_j)$, for $i,j =1, \ldots, n$.
$\mu({\bf x})$ and $k({\bf x},{\bf x}')$ are the mean function and, respectively, the (positive definite) kernel function of the GP.  A GP is usually parametrised with a zero mean function $\mu({\bf x})=0$ and a covariance kernel $k_{\boldsymbol{\theta}}({\bf x},{\bf x}')$ which depends on hyperparameters  $\boldsymbol{\theta} \in \Theta$. A typical example is the automatic relevance determination (ARD) Squared-Exponential (SE) kernel on $\mathbb{R}^c$, $c \in \mathbb{N}$. For ${\bf x},{\bf x}' \in \mathbb{R}^c$ it is defined as
\begin{equation}
\label{eq:RBF}
k_{\boldsymbol{\theta}}({\bf x},{\bf x}')= \displaystyle{\sigma_k^2 \,\exp\left(-\sum_{i=1}^c\frac{ (x_i-x_i')^2}{2 \ell_i^2}\right)},
\end{equation}
where $\boldsymbol{\theta}=[\ell_1,\dots,\ell_c,\sigma_k^2]$ includes the lengthscale hyperparameters $\ell_i >0$ (one for each dimension) and the scale parameter $\sigma_k^2$.
GPs have a natural Bayesian interpretation that makes them ideal for regression problems. For didactic purposes, we review GP regression (with Gaussian likelihood) before moving to general GP  inference (with non-Gaussian likelihood), which will be the main focus in this tutorial.

\textit{Observations and likelihood.} Consider the dataset $\mathcal{D}_m=\{{\bf x}_i,y_i\}_{i=1}^m$ and assume that the output $y_i \in \mathbb{R}$ is equal to the sum of a function $u$ evaluated at the input ${\bf x}_i$  plus Gaussian noise, i.e. $y_i = u({\bf x}_i) + \varepsilon_i$ with $\varepsilon_i\sim N(0, \sigma^2)$ for $i=1, \ldots,m$. Assuming the noise variables are independent, we can write the observation model more compactly as  the likelihood
\begin{equation}
\label{eq:GPliketheor}
p(\mathcal{D}_m| u(X),\sigma^2) =N(\mathbf{y}; {u}(X), \sigma^2I_m),
\end{equation}
where $\mathbf{y}=[y_1,\dots,y_m]^\top$,  $X=[{\bf x}_1,\dots,{\bf x}_m]^\top$ and $I_m$ is the identity matrix of dimension $m$.\footnote{Note that $p(\mathbf{z} 
| \mathbf{m}, \Sigma)=N(\mathbf{z}; \mathbf{m}, \Sigma)$ denotes the probability density function of a multivariate Gaussian with mean $\mathbf{m}$ and covariance $\Sigma$, evaluated at $\mathbf{z}$. In contrast, we use the notation $\mathbf{z} \sim N(\mathbf{m}, \Sigma)$ to indicate that $\mathbf{z}$ is  `distributed as'.}  The variance of the Gaussian likelihood $\sigma^2$ is  considered to be a hyperparameter in GP regression and included in $\boldsymbol{\theta}$.

\textit{Prior.} We assume a GP prior on $u$, that is $u \sim GP(0,k_{\boldsymbol{\theta}})$. This implies that the function evaluated at observed inputs (covariates) ${\bf x}_i$ is a multivariate Gaussian, distributed as:
\begin{equation}
\label{eq:GPpriortheor}
p(u(X)|{\boldsymbol{\theta}})=
 N\left(\begin{bmatrix} u(\mathbf{x}_1) \\ \vdots \\ u(\mathbf{x}_m) \end{bmatrix}; 
\begin{bmatrix}
0 \\
\vdots \\
0
\end{bmatrix},
\begin{bmatrix}
k_{\boldsymbol{\theta}}(\mathbf{x}_1, \mathbf{x}_1) & \cdots & k_{\boldsymbol{\theta}}(\mathbf{x}_1, \mathbf{x}_m) \\
\vdots & \ddots & \vdots \\
k_{\boldsymbol{\theta}}(\mathbf{x}_m, \mathbf{x}_1) & \cdots & k_{\boldsymbol{\theta}}(\mathbf{x}_m, \mathbf{x}_m)
\end{bmatrix}
\right),
\end{equation}
which we can write more compactly as $p(u(X)|{\boldsymbol{\theta}})= N(u(X);0,K_{\boldsymbol{\theta}}(X,X))$, where   $K_{\boldsymbol{\theta}}(X,X)$ is a matrix whose ij-th element is defined as  $(K_{\boldsymbol{\theta}}(X,X))_{ij}=k_{\boldsymbol{\theta}}({\bf x}_i,{\bf x}_j)$. The prior depends on the values of the kernel hyperparameters $\boldsymbol{\theta}$.

\textit{Posterior.} The posterior, obtained combining the likelihood \eqref{eq:GPliketheor} and the prior \eqref{eq:GPpriortheor}, is also a multivariate Gaussian distribution $p(u(X)|\mathcal{D}_m,\boldsymbol{\theta})= N(u(X);\mu_p(X),K_p(X,X))$ with mean and covariance given by:
\begin{align}
    \mu_p(X) &= K_{\boldsymbol{\theta}}(X,X) (K_{\boldsymbol{\theta}}(X,X)+\sigma^2I_n)^{-1}\mathbf{y} \\
    k_p(X, X) &= K_{\boldsymbol{\theta}}(X,X) - K_{\boldsymbol{\theta}}(X,X)  (K_{\boldsymbol{\theta}}(X,X)+\sigma^2I_n)^{-1}K_{\boldsymbol{\theta}}(X,X).
\end{align}

\textit{Predictive posterior.} The predictive posterior of the function $u$ evaluated at a new test point ${\bf x}^*\in \mathcal{X}$ can be computed as follows:
\begin{equation}
  p(u({\bf x}^*)|\mathcal{D}_m,{\boldsymbol{\theta}}) = \int p(u({\bf x}^*)|u(X)) N(u(X);\mu_p(X),K_p(X,X)) du(X).
\end{equation}
Since the joint $p(u({\bf x}^*),u(X))$ is Gaussian (with an expression similar to \eqref{eq:GPpriortheor}), then the conditional probability density function  $p(u({\bf x}^*)|u(X))$ is also Gaussian. Therefore, $p(u({\bf x}^*)|\mathcal{D}_m,{\boldsymbol{\theta}})$ is also a Gaussian probability density function. This holds true for all the conditionals\\  $p(u({\bf x}_1^*),\dots,u({\bf x}_r^*)|\mathcal{D}_m,{\boldsymbol{\theta}})$ for every $r$, and, therefore, the predictive posterior is a Gaussian process $GP(\mu_p,k_p)$, with mean and covariance (kernel) function given by:
\begin{align}
\label{eq:posterpredmean}
    \mu_p({\bf x}^*) &= K_{\boldsymbol{\theta}}({\bf x}^*,X) (K_{\boldsymbol{\theta}}(X,X)+\sigma^2I_n)^{-1}\mathbf{y}, \\
\label{eq:posterpredcov}
    k_p({\bf x}^*, {\bf x}^*) &= K_{\boldsymbol{\theta}}({\bf x}^*,{\bf x}^*) - K_{\boldsymbol{\theta}}({\bf x}^*,X)  (K_{\boldsymbol{\theta}}(X,X)+\sigma^2I_n)^{-1}K_{\boldsymbol{\theta}}(X,{\bf x}^*).
\end{align}

\textit{Hyperparameters selection.} The hyperparameters of the model $\boldsymbol{\theta}$ (including the variance of the likelihood $\sigma^2$) are commonly estimated by maximising the marginal likelihood:
$$
p(\mathcal{D}_m|\boldsymbol{\theta})= N(\mathbf{y};0,K_{\boldsymbol{\theta}}(X,X)+\sigma^2 I_m).
$$

In tasks with likelihoods different from the Gaussian, the posterior is not a multivariate Gaussian and, therefore, the predictive posterior is not a GP. For  Probit  (classification/preference learning) and skew-Gaussian   likelihoods, the posterior is Skew Gaussian  and the predictive posterior is a Skew GP \citep{benavoli2021}.\footnote{A Skew Gaussian distribution (and also a Skew GP) is an extension of the normal distribution that includes additional parameters (respectively functions) to control asymmetry, allowing the distribution to be skewed rather than being perfectly symmetric.} We point the reader to Appendix \ref{app:skewgp} for details. For other likelihoods, the posterior does not have a closed-form and is approximated  using four main approaches: (i) Laplace's approximation (LP) \citep{mackay1996bayesian,williams1998bayesian}; (ii) Expectation Propagation (EP) \citep{minka2001family}; (iii) Kullback-Leibler divergence (KL) minimization \citep{opper2009variational}, comprising Variational Bounding (VB) \citep{gibbs2000variational} as a particular case; (iv) Markov Chain Monte Carlo (MCMC) methods such as elliptical-slice sampling \citep{pmlr-v9-murray10a}.

For all the approaches discussed above, the process of learning  a function with a non-Gaussian likelihood can be summarised by the following Algorithm \ref{alg:1}. It addresses the general case where the goal is to learn a vector of unknown (in our case utility) functions from vector-valued inputs (covariates) and  outputs (responses). We will later demonstrate that GP-based models for learning from preference and choice data can be naturally expressed within this learning framework.
\begin{algorithm}
\caption{Generic inference algorithm for GPs}\label{alg:1}
\begin{description}
\item[Data:] $\mathcal{D}_m=({\bf x}_i,{\bf y}_i)_{i=1}^m$;
\item[Define the likelihood:]  $p(\mathcal{D}_m|{\bf u})$;
\item[Unknown functions vector:] ${\bf u}=(u_1,\dots,u_d)$; 
\item[Define the GP prior on ${\bf u}$:]  $p({\bf u}|\boldsymbol{\theta})=GP({\bf u};{\bf 0},K_{\boldsymbol{\theta}})$;
\item[Algorithm:] ~
\begin{description}
    \item[A1:] Select a numeric method to approximate the posterior $p({\bf u}(X)|\mathcal{D}_m,\boldsymbol{\theta})$;
    \item[A2:]  Based on the select approximation method, define an approximation of the marginal likelihood $p(\mathcal{D}_m|\boldsymbol{\theta})$ and maximise it to learn the optimal hyperparameters $\hat{\boldsymbol{\theta}}$.
    \item[A3:]  Use the approximate posterior $\tilde{p}({\bf u}(X)|\mathcal{D}_m,\hat{\boldsymbol{\theta}})$ and the Gaussian marginal $p({\bf u}({\bf x}^*)|{\bf u}(X))$  to compute the approximated predictive posterior for the value of the functions evaluated at ${\bf x}^*$, that is for ${\bf u}({\bf x}^*)$.  
\end{description}
\end{description}
\end{algorithm}
\newpage
Note that, the inference steps for the Gaussian likelihood case discussed earlier are a special instance of Algorithm~\ref{alg:1}, where both the posterior and the marginal likelihood can be computed in closed form, as they are Gaussian.

The predictive posterior, produced as the output of Algorithm~\ref{alg:1}, serves as the basis for decision-making. It can be used to compute expected values, credible intervals, and other decision criteria, typically via sampling.

\subsubsection{Computational aspects}

A drawback of GPs is their computational complexity, $\mathcal{O}(m^3)$ time complexity and
$\mathcal{O}(m^2)$ memory complexity, where $m$ is a size of the training data.\footnote{The matrix $K(X,X)$  has $m \times m$ elements and needs to be inverted to compute the mean and covariance, as shown in \eqref{eq:posterpredmean}--\eqref{eq:posterpredcov}.} However, there are a  number of well established ways to scale up GPs (to $\mathcal{O}(m)$) that can be easily applied to the above models by using inducing points \citep{quinonero2005unifying,snelson2006sparse}, in combination with variational inference \citep{pmlrv5titsias09a,Hensman2013,hernandez2016scalable} or other well-established approaches \citep{bauer2016understanding,SCHURCH2020,schuch2023correlated}. In Section \ref{sec:movie}, we will demonstrate how the variational approximation enables to scale  Model~\ref{model:pairwiselabel} to a large dataset for a movie recommender system.

\chapter{Learning from object preferences}
\label{sec:objpref}
As we discussed in Section \ref{sec:intro}, there are two types of preferences: preferences on objects and preferences on labels. In this section, we will focus on the first type.

Therefore, our aim is to learn Alice's preference relation $\succ$  on a set of objects $\mathcal{X}$  from a finite set of examples, the training set.  We represent each object by the feature vector ${\bf x}\in \mathbb{R}^c$ of its characteristics  and denote the objects in $\mathcal{X}$ as  $\bx_i \in \mathcal{X}$.
We will explore different variants of this problem that allow for some deviations from rationality in the subject, which can help address practical challenges when comparing objects.

\section{Learning consistent preferences}
\label{sec:consistent}
First, we will focus on learning consistent (rational) preferences. Consider a vector of objects $X=[\bx_1,\bx_2,\dots,\bx_r]^\top$ such that $\bx_i \in \mathbb{R}^c$ for all $i=1,\dots,r$, we denote the training set of $m$ pairwise preferences as
$$
 \mathcal{D}_m = \{ \bx_{i_s} \succ \bx_{j_s}:~~ s = 1,\dots,m\},
 $$ 
where $\bx_{i_s} \succ \bx_{j_s}$ is the s-th pairwise preference,  with $\bx_{i_s} \neq \bx_{j_s}$, $\bx_{i_s},\bx_{j_s} \in X$, 
$i_s,j_s \in\{1,2,\dots,r\}$ for each $s = 1,\dots,m$. As discussed in Section \ref{sec:pref}, whenever Alice's preference relation is \textit{consistent}, that is, it satisfies asymmetry and negative transitivity, we can assume that Alice has in mind a function $u$ that attaches to each alternative a numerical value; the
higher the value, the better Alice likes the alternative.  We can then rewrite $\mathcal{D}_m$ as $\mathcal{D}_m = \{ u(\bx_{i_s})-u(\bx_{j_s})>0:~~ s = 1,\dots,m\}$. 
Preference learning is then the task of learning the function $u$ from $\mathcal{D}_m$.

\begin{example}[Thermal comfort]
\label{ex:1}
To introduce the concepts, we make use of a simple  running example: learning Alice's  preferred home temperature.
We present Alice with pairs of temperatures (in Celsius) and ask her preferred value. Imagine Alice's utility is as depicted in Figure \ref{fig:hometemp1}, namely, she has three locally most preferred temperature depending on her activity in the house (home-fitness, working, relaxing). 
\begin{figure}[h]
\centering
\includegraphics[width=7cm]{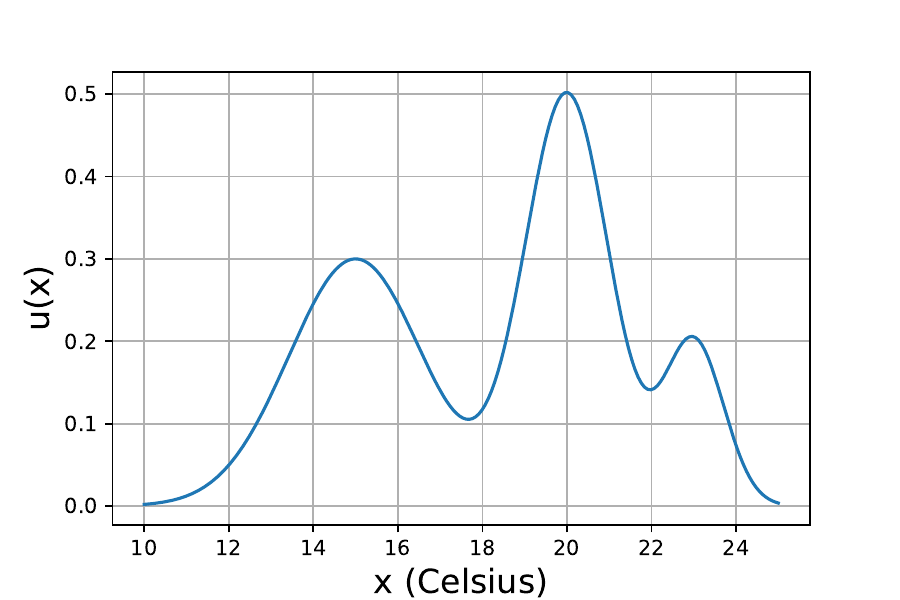}
\caption{Alice's utility for home temperature.}
\label{fig:hometemp1}
\end{figure}
In order to infer her utility function, we consider the set of temperatures (objects) $\mathcal{X}=\{10,11,12,\dots,25\}$ and asks her preferences for $19$ pairs of temperatures. Based on the utility in Figure \ref{fig:hometemp1}, this results in the dataset: 
\begin{align}
\nonumber
\mathcal{D}_{19}=&\{12\succ 10,13\succ 25,14\succ 13,15\succ 22,15\succ 23,16\succ 11,19\succ 15, \\ \nonumber
&19\succ 21,19\succ 22, 19\succ 24,20\succ 10,20\succ 14,20\succ 21,20\succ 24,\\ 
\label{eq:tempdataset}
&20\succ 25, 21\succ 13, 21\succ 25, 23\succ 25, 24\succ 25\}.
\end{align}
We aim to learn Alice's utility function for home-temperature from $\mathcal{D}_{19}$ and predict her preferences for any other pairs of temperatures.
\end{example}

To this purpose, it is useful to represent the statements in $\mathcal{D}_m$ in matrix form. Define $u(X)=[u(\bx_1),u(\bx_2),\dots,u(\bx_r)]^\top$ and $W$, a $m \times r$  matrix whose s-th row is all zero apart from $W_{i_s}=1,W_{j_s}=-1$. Then we can equivalently write the preference statements using a \textit{matrix representation} $\mathcal{D}_m = \{ Wu(X)>0\}$,  where the inequality is applied element-wise.

\begin{example}
\label{ex:2}
For the dataset $\mathcal{D}_{19}$ in Example \ref{ex:1}, the constraints
$ Wu(X)>0$ are as follows:

$$
\begin{bsmallmatrix*}[r]
  -1 & 0 & 1 & 0 & 0 & 0 & 0 & 0 & 0 & 0 & 0 & 0 & 0 & 0 & 0\\
  0 & 0 & 0 & 1 & 0 & 0 & 0 & 0 & 0 & 0 & 0 & 0 & 0 & 0 & -1\\
  0 & 0 & 0 & -1 & 1 & 0 & 0 & 0 & 0 & 0 & 0 & 0 & 0 & 0 & 0\\
  0 & 0 & 0 & 0 & 0 & 1 & 0 & 0 & 0 & 0 & 0 & -1 & 0 & 0 & 0\\
  0 & 0 & 0 & 0 & 0 & 1 & 0 & 0 & 0 & 0 & 0 & 0 & -1 & 0 & 0\\
  0 & -1 & 0 & 0 & 0 & 0 & 1 & 0 & 0 & 0 & 0 & 0 & 0 & 0 & 0\\
  0 & 0 & 0 & 0 & 0 & -1 & 0 & 0 & 1 & 0 & 0 & 0 & 0 & 0 & 0\\
  0 & 0 & 0 & 0 & 0 & 0 & 0 & 0 & 1 & 0 & -1 & 0 & 0 & 0 & 0\\
  0 & 0 & 0 & 0 & 0 & 0 & 0 & 0 & 1 & 0 & 0 & -1 & 0 & 0 & 0\\
  0 & 0 & 0 & 0 & 0 & 0 & 0 & 0 & 1 & 0 & 0 & 0 & 0 & -1 & 0\\
  -1 & 0 & 0 & 0 & 0 & 0 & 0 & 0 & 0 & 1 & 0 & 0 & 0 & 0 & 0\\
  0 & 0 & 0 & 0 & -1 & 0 & 0 & 0 & 0 & 1 & 0 & 0 & 0 & 0 & 0\\
  0 & 0 & 0 & 0 & 0 & 0 & 0 & 0 & 0 & 1 & -1 & 0 & 0 & 0 & 0\\
  0 & 0 & 0 & 0 & 0 & 0 & 0 & 0 & 0 & 1 & 0 & 0 & 0 & -1 & 0\\
  0 & 0 & 0 & 0 & 0 & 0 & 0 & 0 & 0 & 1 & 0 & 0 & 0 & 0 & -1\\
  0 & 0 & 0 & -1 & 0 & 0 & 0 & 0 & 0 & 0 & 1 & 0 & 0 & 0 & 0\\
  0 & 0 & 0 & 0 & 0 & 0 & 0 & 0 & 0 & 0 & 1 & 0 & 0 & 0 & -1\\
  0 & 0 & 0 & 0 & 0 & 0 & 0 & 0 & 0 & 0 & 0 & 0 & 1 & 0 & -1\\
  0 & 0 & 0 & 0 & 0 & 0 & 0 & 0 & 0 & 0 & 0 & 0 & 0 & 1 & -1\\
\end{bsmallmatrix*}\begin{bsmallmatrix*}
  u(10)\\
  u(11)\\
  u(12)\\
  u(13)\\
  u(14)\\
  u(15)\\
  u(16)\\
  u(18)\\
  u(19)\\
  u(20)\\
  u(21)\\
  u(22)\\
  u(23)\\
  u(24)\\
  u(25)\\
\end{bsmallmatrix*}>0.$$ 
It can be noticed that the 8-th column is all zero, since the temperature 18 is not included in $\mathcal{D}_{19}$. What can we infer about Alice's preference $20 \stackrel{?}{\succ} 18$ from the data in $\mathcal{D}_{19}$? We will answer this question in Example \ref{ex:4}.
\end{example}

We can express the statements in  $ Wu(X)>0$ as a likelihood:
\begin{equation}
\label{eq:likeind0}
    p(\mathcal{D}_m|u(X))=\prod_{s=1}^m I_{\{u(\bx_{i_s})-u(\bx_{j_s})>0\}}(u(X))=I_{\{W u(X) >0\}}(u(X)),
\end{equation}
where $I_{A}(x)$ is the indicator function: it is equal to one when $x \in A$, equivalently when $x$ satisfies the statement $A$, and zero otherwise.

In GP-based preference learning, we assume a GP prior over $u(X)$, the likelihood \eqref{eq:likeind0} and then compute  the posterior distribution for $u(X)$, by following the steps described in Algorithm \ref{alg:1}.

\begin{lemma}
Assuming $u \sim \mathrm{GP}(0,k_{\boldsymbol{\theta}})$ with  a strictly positive definite kernel\footnote{A kernel function is strictly positive definite if the matrix $K_{\boldsymbol{\theta}}(X,X)$ is positive definite for all $X$.} $k_{\boldsymbol{\theta}}$, the posterior is equal to:
\begin{equation}
    \label{eq:posteriorexact}
p(u(X)|\mathcal{D}_m,\boldsymbol{\theta})=TN_{\{W u(X) >0\}}(u(X);\mathbf{0},K_{\boldsymbol{\theta}}(X,X)),
\end{equation}
where the term on the right hand side denotes a multivariate normal distribution with zero mean and covariance matrix $K_{\boldsymbol{\theta}}(X,X)$ truncated in $\{u(X) \in \mathbb{R}^r: W u(X) > 0\}$. The posterior is well-defined\footnote{The normalisation constant is different from zero.} provided that $\succ$ satisfies asymmetry and negative transitivity.
\end{lemma}

\begin{remark}
Since the set $\{u(X)\in \mathbb{R}^r: W u(X) \geq 0\}\backslash \{u(X)\in \mathbb{R}^r: W u(X) > 0\}$ has zero measure with respect to the Gaussian distribution $N(u(X);\mathbf{0},K_{\boldsymbol{\theta}}(X,X))$, we can equivalently write the posterior as:
\begin{equation}
    \label{eq:posteriorexact1}
p(u(X)|\mathcal{D}_m,\boldsymbol{\theta})=TN_{\{W u(X) \geq 0\}}(u(X);\mathbf{0},K_{\boldsymbol{\theta}}(X,X)).
\end{equation}
 Observe that, this is different from assuming that the underlying preference relation is \textit{weak} (instead of \textit{strict}). Indeed we cannot learn a weak preference relation using a GP-based model. To see that, consider for instance the training set  $\mathcal{D}_2 = \{ u(\bx_1) - u( \bx_2)\geq 0 , u(\bx_2) - u( \bx_1)\geq 0 \}$ (which violates asymmetry), this implies that  $u(\bx_1)=u(\bx_2)$ but the probability of the evidence:
 $$
p(\mathcal{D}_2|\boldsymbol{\theta})= \int I_{\{u(\bx_1)=u( \bx_2)\}}(u(X))N(u(X);\mathbf{0},K_{\boldsymbol{\theta}}(X,X)) du(X)=0,
 $$
 and, therefore, the posterior cannot be defined via Bayes' rule (the denominator is zero).
\end{remark}

Given the posterior \eqref{eq:posteriorexact1}, we can compute the predictive posterior as follows.

\begin{proposition}
\label{eq:prepost}
The posterior predictive distribution for $u(X^*)$ given the test-points $X^*=[\bx_1,\dots,\bx_p]$ is 
\begin{align}
   \nonumber
&p(u(X^*)|\mathcal{D}_m)\\
 \label{eq:posteriorexactpred}
&=\int p(u(X^*)|u(X))\;TN_{\{W u(X) \geq 0\}}(u(X);\mathbf{0},K_{\boldsymbol{\theta}}(X,X))du(X),
\end{align}
where
\begin{equation}
\label{eq:prediGP}
\begin{aligned}
p(u(X^*)|u(X),\boldsymbol{\theta})
=&~N(u(X^*);K_{\boldsymbol{\theta}}(X^*,X)K^{-1}_{\boldsymbol{\theta}}(X,X)u(X),\\
&~K_{\boldsymbol{\theta}}(X^*,X^*)-K_{\boldsymbol{\theta}}(X^*,X)K^{-1}_{\boldsymbol{\theta}}(X,X)K_{\boldsymbol{\theta}}(X,X^*)).
\end{aligned}
\end{equation}
\end{proposition}
The computation of posterior and predictive posterior expectations is nontrivial due to the analytically intractable normalising constant of the truncated normal distribution.  However, we can easily approximate the  posterior  by Monte Carlo sampling. Efficient sampling of truncated multivariate normal distributions under linear inequality constraints can be performed  by using Gibbs sampling \citep{kotecha1999gibbs,taylor2016restrictedmvn} linear elliptical slice sampler (\textit{lin-ess}) \citep{gessner2019integrals},  minimax tilting method accept-reject sampler \citep{botev2017normal} and Hamiltonian Monte-Carlo \citep{pakman2014exact}.

We will use \textit{lin-ess} because it only needs the Cholesky factorisation of $K_{\boldsymbol{\theta}}(X,X)$, details can be found in \citet{gessner2019integrals}. We provide a visual illustration of this sampling procedure in the next example.

\begin{example}[Lin-ess]
\label{ex:3}
Assume for simplicity we have only one preference in the dataset $\mathcal{D} = \{ u(\bx_2) - u( \bx_1)\geq 0\}$. The above constraint on $u$ defines a half-space region depicted in Figure \ref{fig:lin-ess} (shaded blue region).
Let $s_0$ be the sample of the vector $[ u(\bx_1), u(\bx_2)]^\top$ accepted at the previous step that is satisfying the constraint. Lin-ess works as follows: (i) sample a new point from the GP (Multivariate-Normal) prior, which in this case is equal to:
$$
\nu \sim N\left(\begin{bmatrix}
0\\
0\\
\end{bmatrix},\begin{bmatrix}
k_{\boldsymbol{\theta}}(\bx_1,\bx_1)&k_{\boldsymbol{\theta}}(\bx_1,\bx_2)\\
k_{\boldsymbol{\theta}}(\bx_2,\bx_1)& k_{\boldsymbol{\theta}}(\bx_2,\bx_2) \\
\end{bmatrix}\right).
$$
(ii) define the ellipse $\{\cos(\phi)s_0+\sin(\phi)\nu:~ \phi \in[0,2\pi)\}$ shown in Figure \ref{fig:lin-ess}. (iii) calculate analytically the interval $[\phi_{min},\phi_{max}]$ including the values of $\phi$ corresponding to the slice of the ellipse inside the constraint region. (iv) finally generate a new sample $s_1=\cos({\tilde \phi})s_0+\sin(\tilde{\phi})\nu$ with $\tilde{\phi}$ sampled from the Uniform distribution $ \mathcal{U}(\phi_{min},\phi_{max})$. (v) set $s_0=s_1$ and continue iteratively. The algorithm is  rejection-free, that is each new sample is accepted.  

In the general case, the constraint region is $\{u(X)\in \mathbb{R}^r: W u(X) \geq 0\}$, but the algorithm can deal with this general linear constraint in a similar way.
\begin{figure}[h]
\centering
\includegraphics[width=0.4\linewidth]{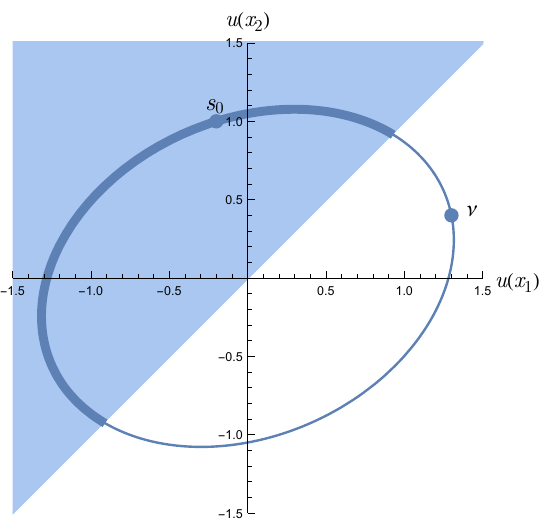}
\caption{Sampling from a constrained region 
using lin-ess,}
\label{fig:lin-ess}
\end{figure}
\end{example}

The only parameters in the model are the hyperparameters $\boldsymbol{\theta}$ in the kernel function. Since the utility function is invariant under increasing transformations, the scale  (variance) parameter $\sigma_k^2$ in \eqref{eq:RBF} is not identifiable. Therefore,  we  assume  that $\sigma_k^2=1$. 
The other hyperparameters can be selected by maximising the \textit{marginal likelihood}:
\begin{align}
\nonumber
\hat{\boldsymbol{\theta}} &= \arg\max_{\boldsymbol{\theta}\in \Theta} p(\mathcal{D}_m|\boldsymbol{\theta}) ~\text{ with } \\ \label{eq:ML}
&p(\mathcal{D}_m|\boldsymbol{\theta}) =\int I_{\{W u(X) \geq0\}}(u(X))N(u(X);\mathbf{0},K_{\boldsymbol{\theta}}(X,X)) du(X). 
\end{align}
The integral in \eqref{eq:ML} is intractable. There are a variety of approximate inference techniques that one can resort to. We use the following approach: (a) we approximate the indicator function with a sigmoid-type function; (b) we perform a local expansion of the integrand around its maximum and then analytically compute the integral. Note that, the step (b) corresponds to the Laplace's approximation  \citep{rasmussen2006gaussian}.

\begin{example}
\label{ex:4}
Our aim is to infer Alice's utility $u(\bx)$ from the dataset of pairwise preferences  $\mathcal{D}_{19}$ in Example \ref{ex:1}.  We will then use the
learned model to predict Alice's preference for $20 \stackrel{?}{\succ} 18$. As described in this section, we place a GP prior on $u$ and use the SE kernel in \eqref{eq:RBF} with $\sigma_k^2=1$. The value of the lengthscale is fixed to $\hat{\ell}=1.5$. We then sample 60,000 samples of $u(X)$ from the posterior with \textit{lin-ess}, and compute the predictive posterior for $200$ equally spaced temperatures in the interval $[10,25]$. Figure \ref{fig:th20}-left shows the corresponding mean and 95\% credible interval. The uncertainty is high due to the 
small  dataset. 
\begin{figure}[h]
\centering
\begin{subfigure}[b]{.85\linewidth}
\includegraphics[width=0.5\linewidth]{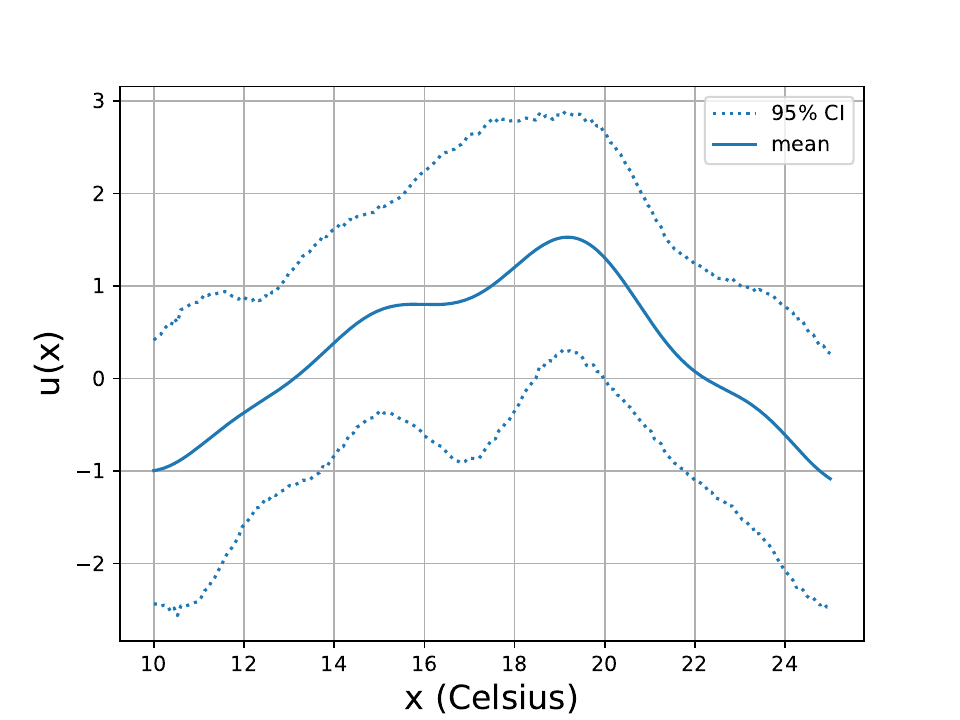}\includegraphics[width=0.5\linewidth]{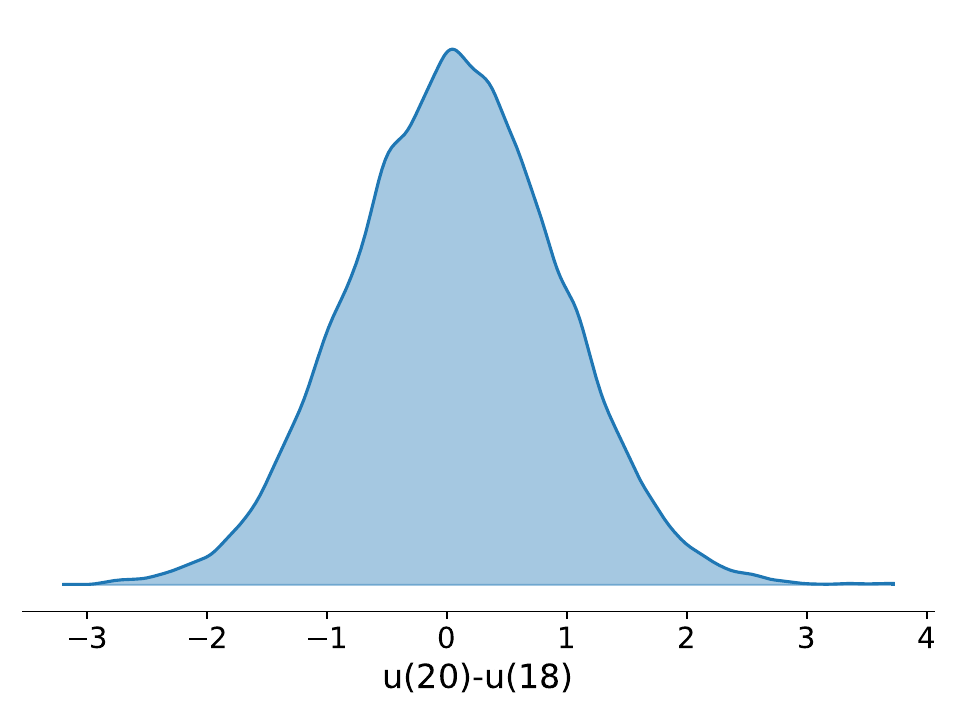}
\caption{20 pairwise preferences}\label{fig:th20}
\end{subfigure}\\
\begin{subfigure}[b]{.85\linewidth}
\includegraphics[width=0.5\linewidth]{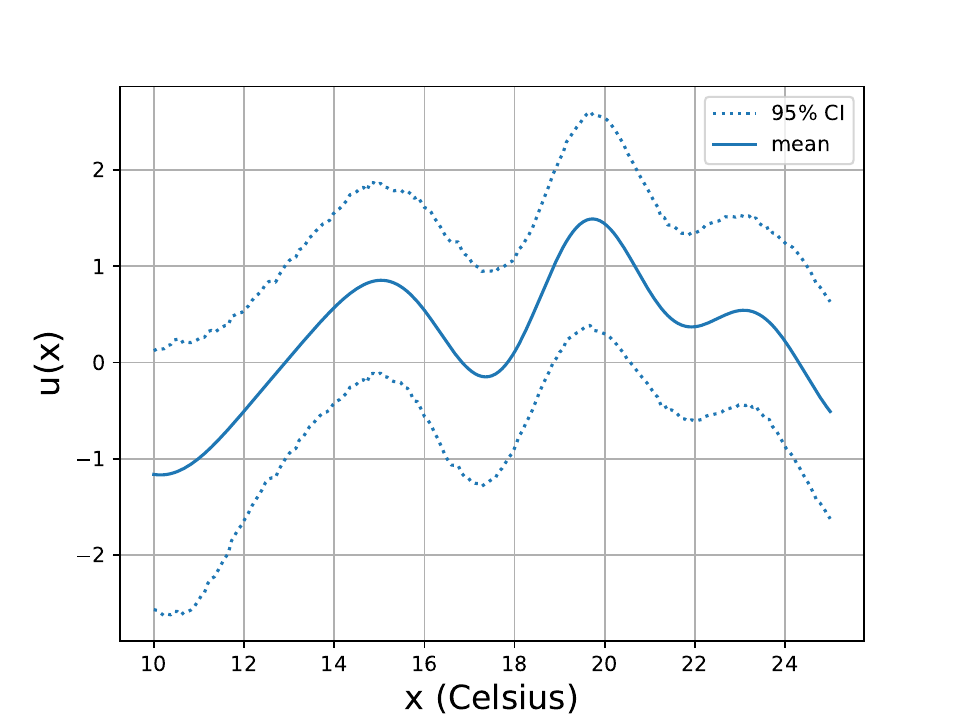}\includegraphics[width=0.5\linewidth]{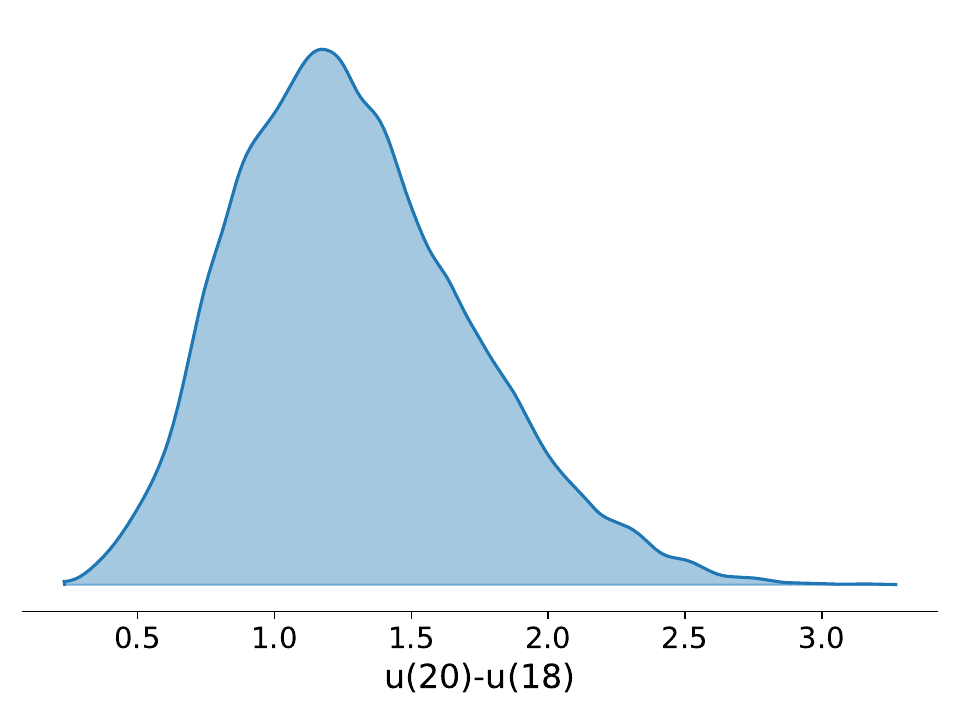}
\caption{40 pairwise preferences}\label{fig:th40}
\end{subfigure}\\
\caption{Left: learned utility functions, the mean is the continuous line and the region between the dotted-lines is the 95\% credible region. Right: posterior marginal of the difference $u(20)-u(18)$.}
\label{fig:th2040}
\end{figure}
This is the advantage of using a Bayesian approach to estimate the utility function $u$: it quantifies the uncertainty of its own estimate. Indeed, the trained model is undecided about $u(20)-u(18)\stackrel{?}{>} 0$ as shown by the distribution in Figure \ref{fig:th20}-right (the pair $18,20$ is not included in the training dataset). Note that, the predictive posterior for $u(20)-u(18)$ has mass over both the positive and negative values. The uncertainty decreases if we consider 40 pairwise preferences as shown in Figure \ref{fig:th40}-left (compare it with Figure \ref{fig:hometemp1}). From Figure \ref{fig:th40}-right, we conclude that $u(20)-u(18)> 0$ with probability $\approx 1$, that is Alice prefers the temperature $20$ to $18$.  Note that, also in this case, the pair $18,20$ is not included in the dataset.
\end{example}

The model is summarised in the following box.

\begin{SnugshadeB}
\begin{MethodB}[Consistent Preferences]
\label{Model:1}
Consider a vector $X=[\bx_1,\bx_2,\dots,\bx_r]^\top$ of objects such that $\bx_i \in \mathbb{R}^c$ for all $i=1,\dots,r$ and a training set of $m$ preferences $
 \mathcal{D}_m = \{ \bx_{i_s} \succ \bx_{j_s}:~~ s = 1,\dots,m\}
 $ with $\bx_{i_s} \neq \bx_{j_s}$, $\bx_{i_s},\bx_{j_s} \in X$, 
$i_s,j_s \in\{1,2,\dots,r\}$ for each $s = 1,\dots,m$. Assuming the preferences are \textbf{consistent}, there exists a utility function $u$ which represents them. Then, the probability of observing the data $ \mathcal{D}_m $ given the utility function vector $u(X)$ is equal to the likelihood function:
\begin{equation}
\label{eq:likeind0box}
    p(\mathcal{D}_m|u(X))=\prod_{s=1}^m I_{\{u(\bx_{i_s})-u(\bx_{j_s})>0\}}(u(X))=I_{\{W u(X) >0\}}(u(X)).
\end{equation}
Assuming a GP prior $p(u) = \mathrm{GP}(u;0,k_{\boldsymbol{\theta}})$, the predictive posterior for $u(X^*)$ given the test-points $X^*=[\bx_1,\dots,\bx_p]$ is
\begin{equation}
\label{eq:posteriorexactpredbox}
\begin{aligned}
&p(u(X^*)|\mathcal{D}_m)\\
&=\int p(u(X^*)|u(X))\;TN_{\{W u(X) \geq 0\}}(u(X);\mathbf{0},K_{\boldsymbol{\theta}}(X,X))du(X),
\end{aligned}
\end{equation}
where $p(u(X^*)|u(X))$ is defined in Proposition \ref{eq:prepost}. Inference is carried out according to Algorithm \ref{alg:1}, where  the Laplace's approximation is used in  A2  and MCMC (lin-ess) in A3.
\end{MethodB}
\end{SnugshadeB}

\section{Problems arising in the modelling of preference relations}
\label{sec:problems}
What if the preferences of a subject are irrational? Alice's preferences may fail to satisfy \textit{asymmetry} and/or \textit{negative transitivity} for a number of reasons.  We consider here the following three common reasons.

\begin{enumerate}
\itemsep0.1em
    \item  \textit{Limit of discernibility}. As discussed in Section \ref{sec:pref}, all asymmetric and negatively transitive preference relations can be thought to be originated by a utility function according to:
$$
\bx \succ \by ~\textit{ iff }~ u(\bx)>u(\by).
$$
However, consider two objects $\bx,\by$  whose utilities can barely be discerned by Alice, and a third object  $\bz$  whose utility is between that of  $\bx$ and $\by$. In this case, Alice may either not be able to state her preference
for $\bz$ compared to $\bx,\by$ (violating negative transitivity), or state a wrong preference (violating asymmetry or negative transitivity).
\item \textit{Additive noise.} Another potential issue arises in the case the observed utility function differs from the true utility function (due to disturbances, measurement errors).  In this situation, Alice states $\bx \succ \by$ if the observed  utility of $\bx$ is greater than the observed utility of $\by$, i.e.,\ $o(\bx)>o(\by)$.
In random-utility models \citep{mcfadden1974,mcfadden1978}, the observation error is commonly assumed to be additive Gaussian, that is  $o(\bx)=u(\bx)+v_x$ and $o(\by)=u(\by)+v_y$, where $v_x,v_y$ are  Gaussian distributed variables with zero mean and variance $\sigma^2_v$. Due to the observation error, preferences may violate asymmetry or negative transitivity or both.  Another common distribution for the error is the Gumbel distribution, which we will discuss in Section \ref{sec:PL}.
\item \textit{Multiple utilities.} Often the apparent violation of asymmetry and negative transitivity can be explained as the result of the intersection of several primitive preferences. For instance, in the home-temperature example, Alice may compare temperatures based on two  utility functions: relax $u_1$ and fitness $u_2$. Given two temperatures $\bx,\by$, Alice may prefer $\bx \succ \by$ under $u_1$ and $\by \succ \bx$ under $u_2$ leading to conflictual preferences. 
\end{enumerate}

We will now discuss how we can change the previous model to account for these issues.

Note that, many other reasons (and corresponding models) for violations of rationality in preferences have been explored in the behavioural economics literature, such as, for instance,  anchoring effects, framing, loss aversion, and cognitive capacity limits (see, for instance, \citet{rabin1998psychology}). While our goal in this tutorial is not to provide a comprehensive overview of these models, we aim to equip the reader with the tools needed to explore such extensions.\footnote{After going through this tutorial paper, the reader should be able to develop GP-based implementations of these other sources of irrationality, and adapt the code provided in the \textit{prefGP} \citep{prefGP} library to implement these models.}

\section{Accounting for the limit of discernibility}
\label{sec:discern}
In order to account for the \textit{limit of discernibility}, we can employ a cardinal utility function, that is the difference $u(\bx)-u(\by)$ is used as a measure of the similarity between $\bx,\by$. We will consider two models.

\paragraph{Luce's model:}
Following  a model introduced by \cite{luce1956semiorders}, the relation $\succ$ is represented by
a pair $( u , \delta)$ where $u$ is a utility function  and $\delta>0$ is a threshold -- called the \textit{just noticeable difference} -- such that
\begin{equation}
    \label{eq:luce}
   \bx \succ \by ~~\textit{ iff }~~ u(\bx)>u(\by)+\delta. 
\end{equation}
In other words, Alice states  $\bx \succ \by$ only if the difference between the utility functions of the two items is larger than the threshold $\delta$ (which is the limit of discernibility). The  relation defined in \eqref{eq:luce} satisfies \textit{asymmetry} and  \textit{transitivity} (but not \textit{negative transitivity}). 

In the absence of strict preference between two alternatives  $\bx , \by$, that is, if neither $\bx \succ \by$  nor $\by \succ \bx$ holds, then Alice will state that there is no noticeable difference between the two objects, denoted as  $\bx \smile \by$, that is
\begin{equation}
    \label{eq:NND}
   \bx \smile \by ~~\textit{ iff }~~ | u(\bx)-u(\by)| \leq \delta.
\end{equation}
The relation $\smile$ is not an indifference relation because it is reflexive, symmetric but it is not transitive. We can easily modify the  \textit{consistent preferences learning model} presented in Section \ref{sec:consistent} to account for the limit of discernibility under Luce’s model.

  \begin{SnugshadeB}
\begin{MethodB}[Just Noticeable Difference]
\label{Model:JND}
Consider a vector $X=[\bx_1,\bx_2,\dots,\bx_r]^\top$ of objects such that $\bx_i \in \mathbb{R}^c$ for all $i=1,\dots,r$, a training set of $m_a$ preferences denoted as $\mathcal{D}_{m_a} = \{ \bx_{i_s}\succ \bx_{j_s}:~~ s = 1,\dots,m_a\}$ and $m_b$ indiscernibility statements denoted as $\mathcal{D}_{m_b} = \{ \bx_{i_s}\smile \bx_{j_s}:~~ s = 1,\dots,m_b\}$ with $\bx_{i_s} \neq \bx_{j_s}$, $\bx_{i_s},\bx_{j_s} \in X$, 
$i_s,j_s \in\{1,2,\dots,r\}$ for each $s$.
 Under \eqref{eq:luce}, we can rewrite these sets as $\mathcal{D}_{m_a} = \{ \frac{1}{\delta}(u(\bx_{i_s})-u(\bx_{j_s}))>1:~~ s = 1,\dots,m_a\}$ and $\mathcal{D}_{m_b} = \{ \frac{1}{\delta}|u(\bx_{i_s})-u(\bx_{j_s})|\leq 1:~~ s = 1,\dots,m_b\}$.
 Then, the probability of observing the
data $\mathcal{D}_{m_a},\mathcal{D}_{m_b}$ given the utility function vector $u(X)$ is equal to the likelihood function:
\begin{equation}
\label{eq:likeindluce}    p(\mathcal{D}_{m_a},\mathcal{D}_{m_b}|u(X))=p(\mathcal{D}_{m_a}|u(X))p(\mathcal{D}_{m_b}|u(X)),
\end{equation}
\begin{align}
p(\mathcal{D}_{m_a}|u(X))&=\prod_{s=1}^{m_a} I_{\left\{\frac{1}{\delta}(u(\bx_{i_s})-u(\bx_{j_s}))>1\right\}}=I_{\left\{-W_a \frac{u(X)}{\delta}  < -\mathbf{1}_{m_a}\right\}}(u(X)),\\ \nonumber
p(\mathcal{D}_{m_b}|u(X))&=\prod_{s=1}^{m_b} I_{\left\{\frac{1}{\delta}\left(u(\bx_{i_s})-u(\bx_{j_s})\right)\leq 1\right\}}I_{\left\{\frac{1}{\delta}\left(u(\bx_{j_s})-u(\bx_{i_s})\right)\leq 1\right\}} \\ 
&=I_{\left\{W_{b} \frac{ u(X)}{\delta} \leq \mathbf{1}_{2m_b}\right\}}(u(X)),
\end{align}
where $\mathbf{1}_{d}$ is the unit-vector of dimension $d$, $W_a \in \mathbb{R}^{m_a \times r},~W_b \in \mathbb{R}^{2m_b \times r}$ are matrices representing the linear constraints.
By defining $W=[-W_a^\top,W_b^\top]^\top$, $\mathbf{c}=[-\mathbf{1}_{m_a}^\top,\mathbf{1}_{2m_b}^\top]^\top$,
$u'(\cdot)=\frac{u(\cdot)}{\delta}$ and
assuming a GP prior $p(u') = \mathrm{GP}(u';0,k_{\boldsymbol{\theta}})$, the predictive posterior for $u'(X^*)$ given the test-points $X^*=[\bx_1,\dots,\bx_p]$ is
\begin{align}
    \label{eq:posteriorindis}
&p(u'(X^*)|\mathcal{D}_{m_a},\mathcal{D}_{m_b}) \\ \nonumber
&=\int p(u'(X^*)|u'(X))\;TN_{\{W u'(X) \leq \mathbf{c}\}}(u'(X);\mathbf{0},K_{\boldsymbol{\theta}}(X,X))du'(X),
\end{align}
with $p(u'(X^*)|u'(X))$ is defined  in Proposition \eqref{eq:prepost}. Inference is carried out according to Algorithm \ref{alg:1}, where  the Laplace's approximation is used in  A2  and MCMC (lin-ess) in A3.
\end{MethodB}
\end{SnugshadeB}
After  rescaling $u'(\cdot)=\frac{u(\cdot)}{\delta}$, the  only parameters in the model are the hyperparameters $\boldsymbol{\theta}$ in the kernel function. The scale  (variance) parameter $\sigma_k^2$ in \eqref{eq:RBF} is weakly identifiable in this case, because the inequality $|u'(\bx_{i})-u'(\bx_{j})|\leq 1$ determines the level of discernibility. 
The  hyperparameters  $\boldsymbol{\theta}$ can be selected by maximising the \textit{marginal likelihood} using the same approach described in Section \ref{sec:consistent}. The computation of posterior and predictive posterior inferences can again be performed via sampling using \textit{lin-ess}.

\begin{example}
\label{ex:5}
We consider the home-temperature  example, whose true utility function is shown in Figure \ref{fig:hometemp1}.  We assume that two temperatures ${\bf x}_i,{\bf x}_j$ are indiscernible by Alice whenever $|u({\bf x}_i)-u({\bf x}_j)|\leq 0.025$. 
In order to infer Alice's utility function, we consider the set of temperatures (objects) $\mathcal{X}=\{10,11,12,\dots,25\}$ and asks her preferences for $19$ pairs of temperatures. This results in the datasets: 
\begin{align}
\nonumber
\mathcal{D}_{17}=\{&12\succ 10, 13\succ 25, 14\succ 13, 15\succ 22, 15\succ 23, 16\succ 11, 19\succ 22, \\ \nonumber
&19\succ 24,20\succ 10, 20\succ 14,20\succ 21,20\succ 24,20\succ 25, \\ 
\label{eq:tempdatasetLuce}
&21\succ 13, 21\succ 25, 23\succ 25, 24\succ 25\},\\
\label{eq:tempdatasetLucesim}
\mathcal{D}_{2}=&\{15 \smile 19, 19 \smile 21\},
\end{align}
where $\mathcal{D}_{2}$ includes two indiscernible pairs of temperatures. We aim to learn Alice's utility $u'$  from the above datasets and predict her preferences for any other pairs of temperatures.

We place a GP prior on $u'$ with a SE kernel.  Figure \ref{fig:th20JND} shows in the Cartesian-plot the predictive posterior mean and 95\% credible interval for $u'$. The uncertainty is high due to the 
small dataset.  Indeed, note that the model is undecided  about $|u(13)-u(18)|\stackrel{?}{\leq} 1$  as shown by the distribution in Figure \ref{fig:th20JND}. Instead, we can conclude with certainty that $15,19$ are indiscernible and $20 \succ 18$ with high probability.  The uncertainty decreases if we consider 40 pairwise preferences as shown in Figure \ref{fig:th40JND}. From Figure \ref{fig:th40JND}, we conclude that $u(20)-u(18)> 0$ with probability $\approx 1$, that is Alice prefers the temperature $20$ to $18$.
There is also high probability that the temperatures $13$ to $18$ are indiscernible for Alice. 

\begin{figure}[h!]
\centering
\begin{subfigure}[b]{.99\linewidth}
\centering
\includegraphics[width=0.4\linewidth]{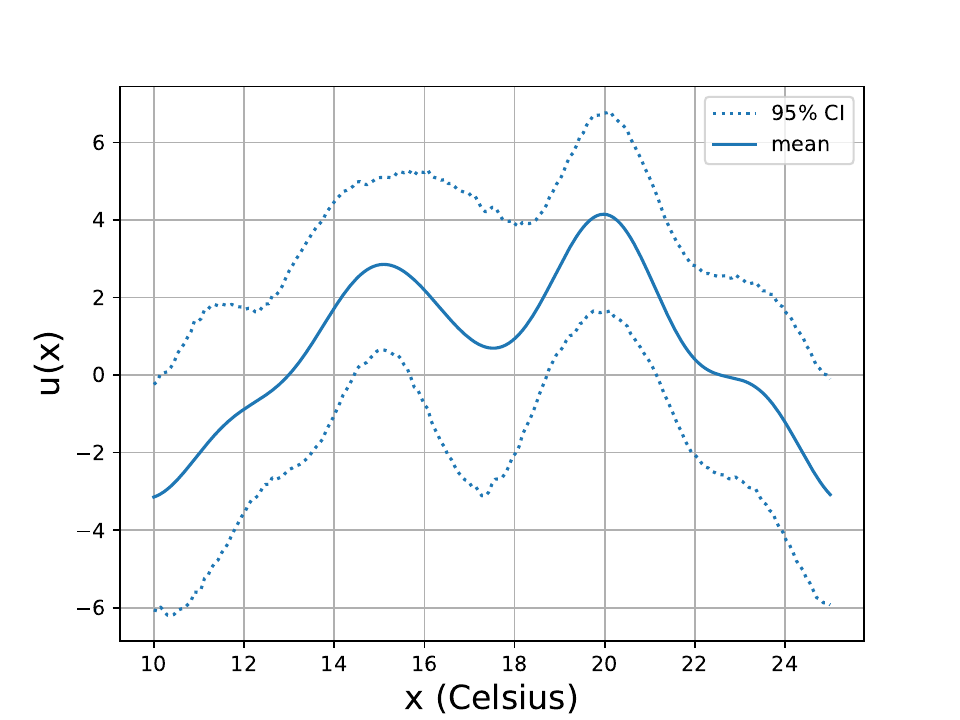}\includegraphics[width=0.4\linewidth]{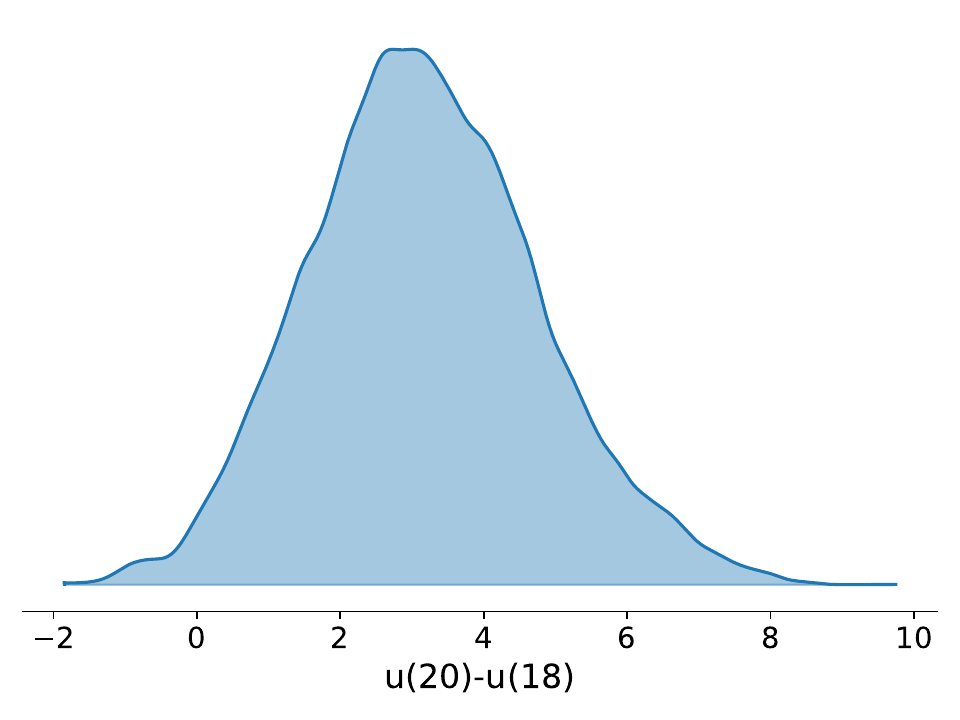}\\\includegraphics[width=0.4\linewidth]{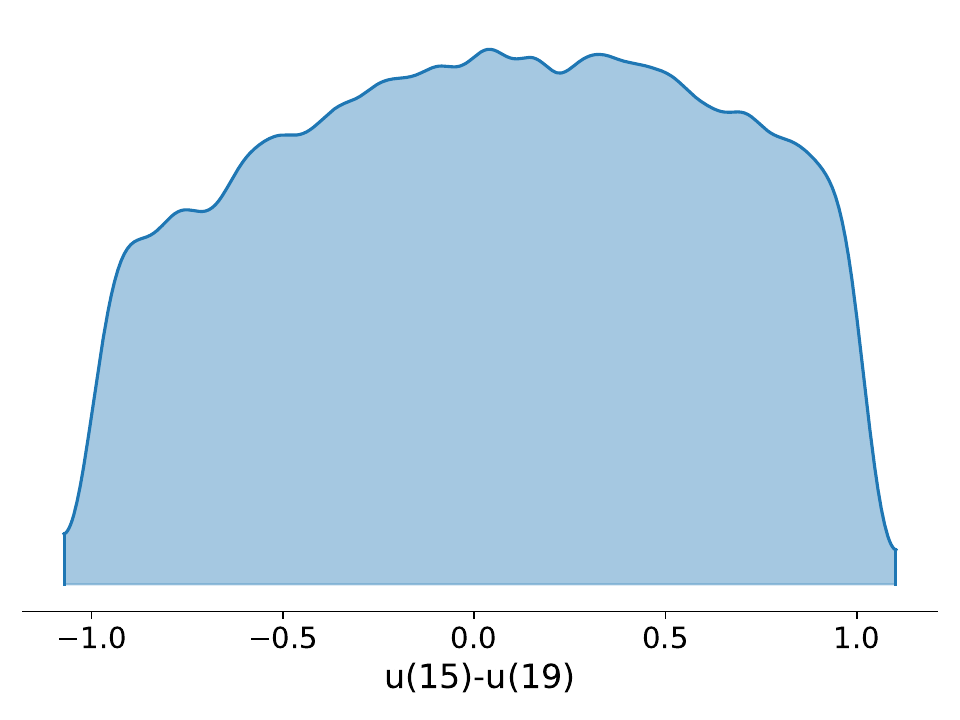}\includegraphics[width=0.4\linewidth]{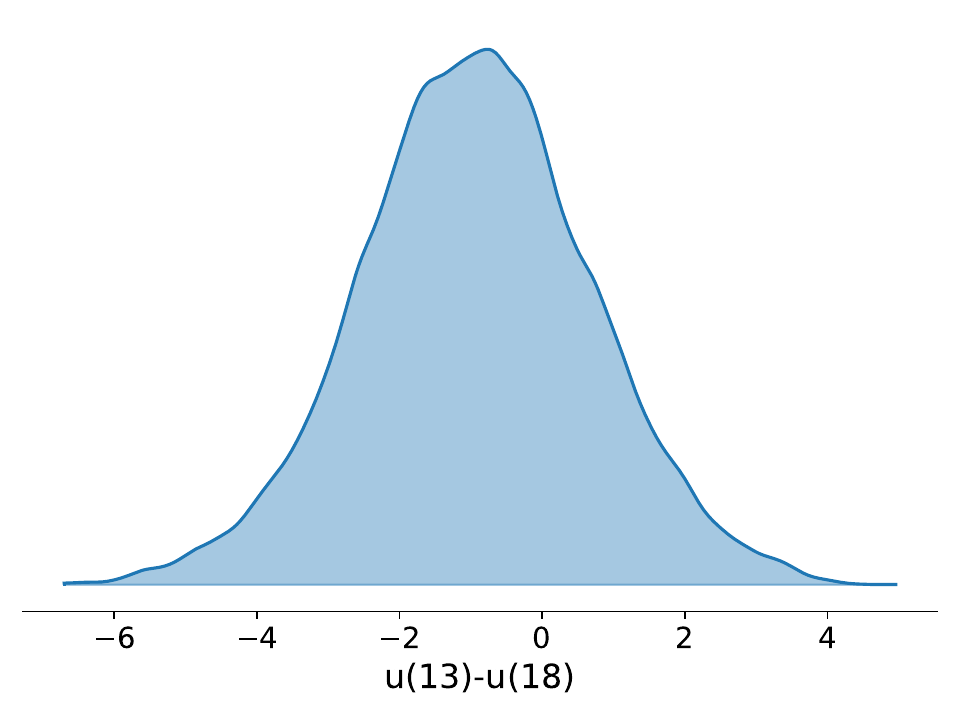}
\caption{20 pairwise preferences}\label{fig:th20JND}
\end{subfigure}\\
\begin{subfigure}[b]{.99\linewidth}
\centering
\includegraphics[width=0.4\linewidth]{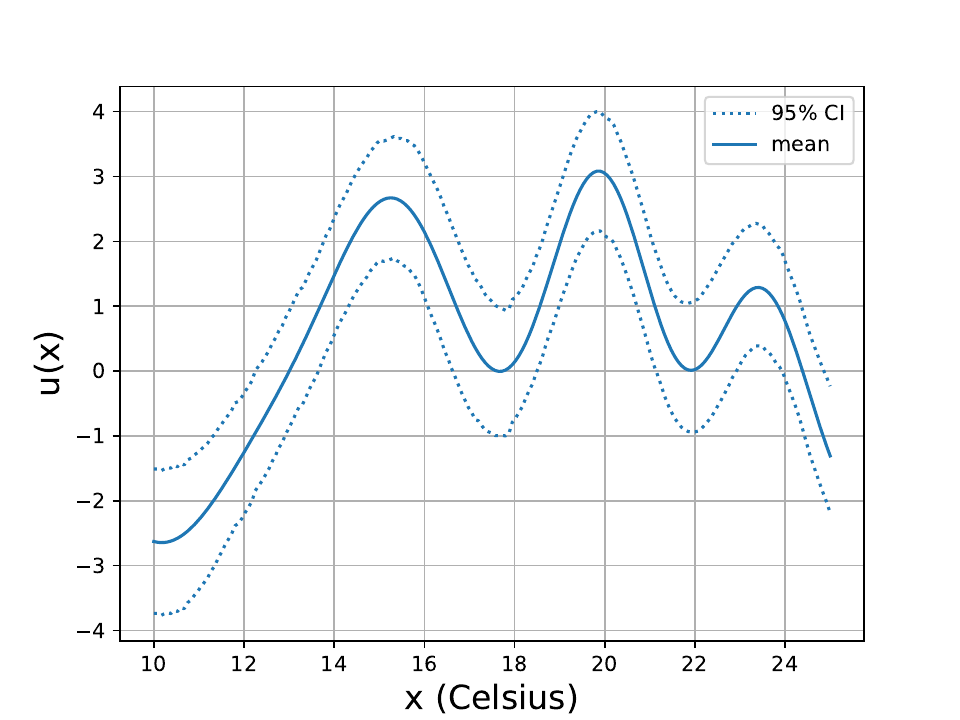}\includegraphics[width=0.4\linewidth]{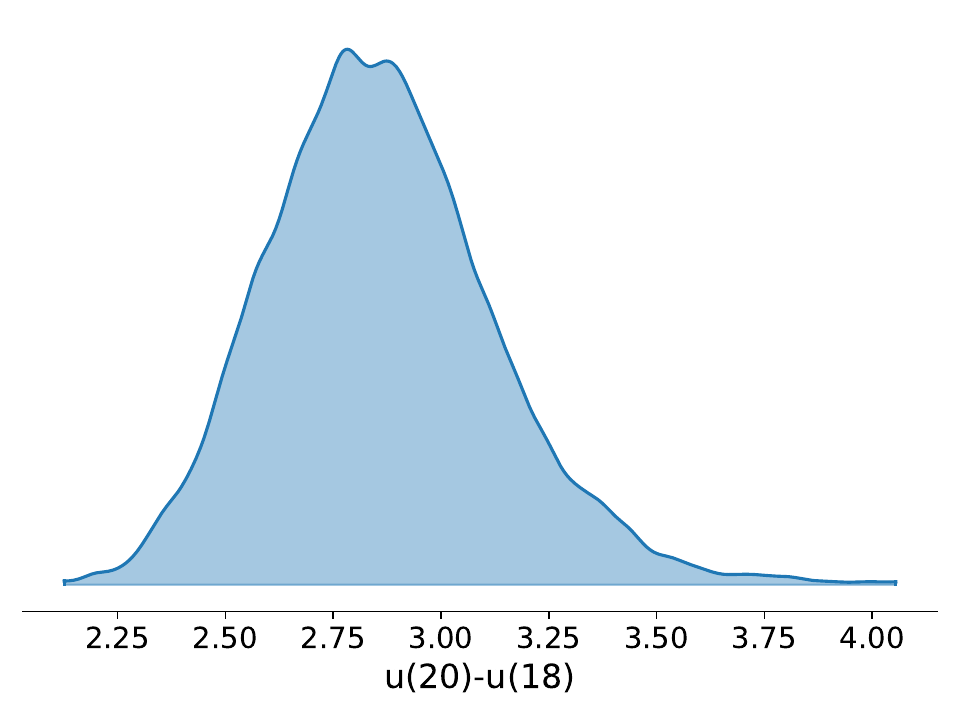}\\\includegraphics[width=0.4\linewidth]{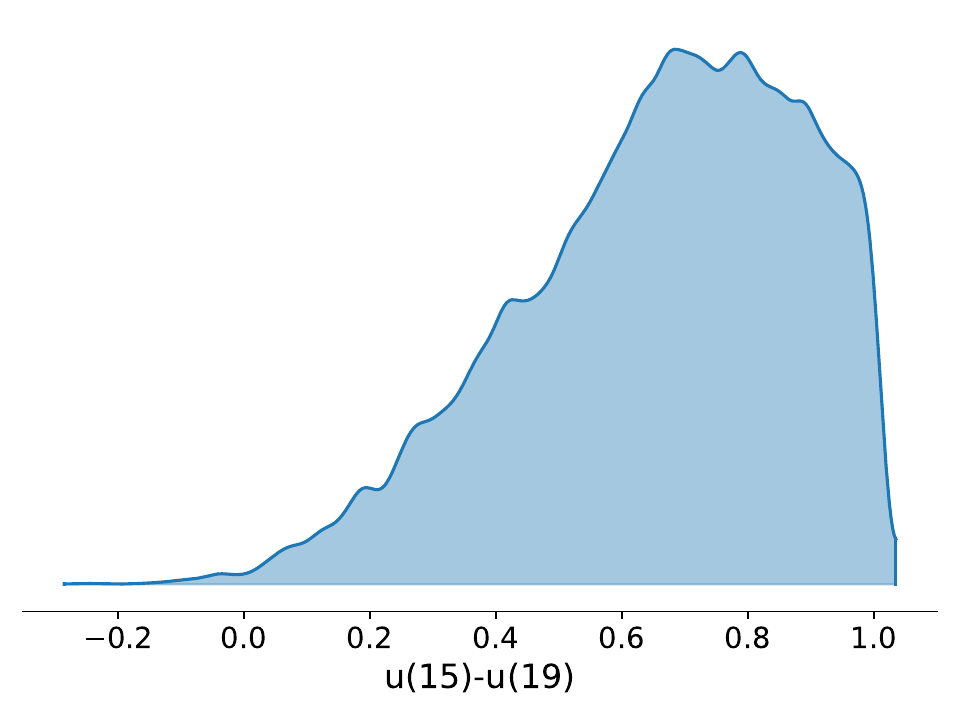}
\includegraphics[width=0.4\linewidth]{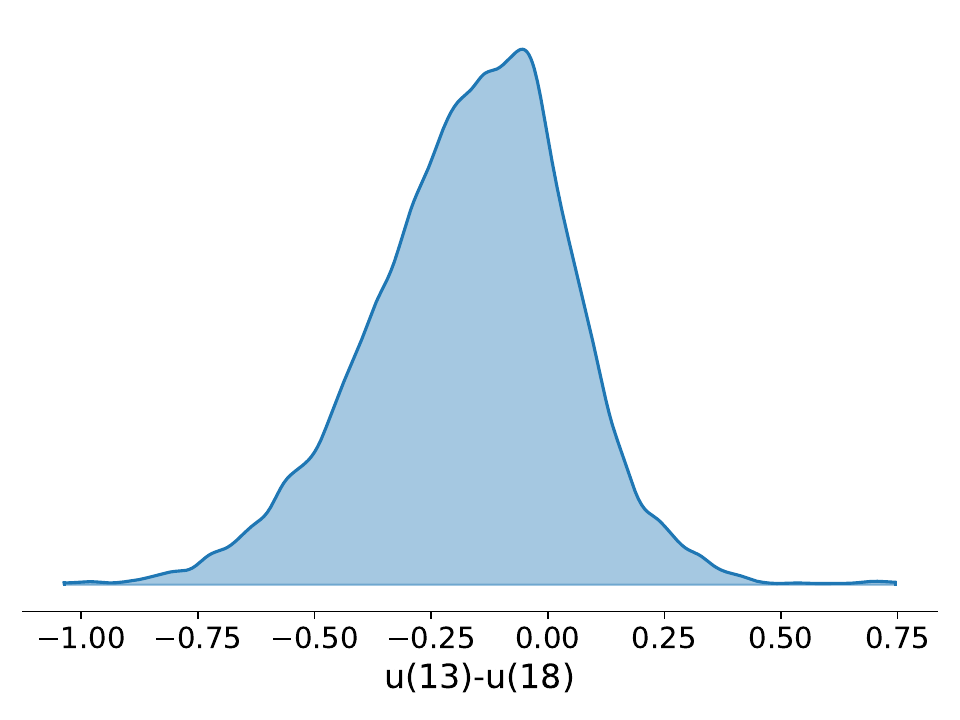}
\caption{40 pairwise preferences}\label{fig:th40JND}
\end{subfigure}\\
\caption{Cartesian plot: learned utility functions under Luce's model, the mean is the continuous line and the region between the dotted-lines is the 95\% credible region. The other three figures are the posterior marginals of $u(20)-u(18)$ (top), $u(15)-u(19)$ (left) and, respectively, $u(13)-u(18)$ (right).}
\label{fig:th2040JND}
\end{figure}
\end{example}

\clearpage 
\paragraph{Erroneous preferences model}
Luce introduced an interval of no noticeable difference to model  Alice's absence of strict preference between two indiscernible objects  $\bx , \by$. However, there are applications where Alice is forced to express her preference between   $\bx , \by$ even when the two objects are indiscernible.

\begin{figure}[h!]
\centering
\includegraphics[width=6cm]{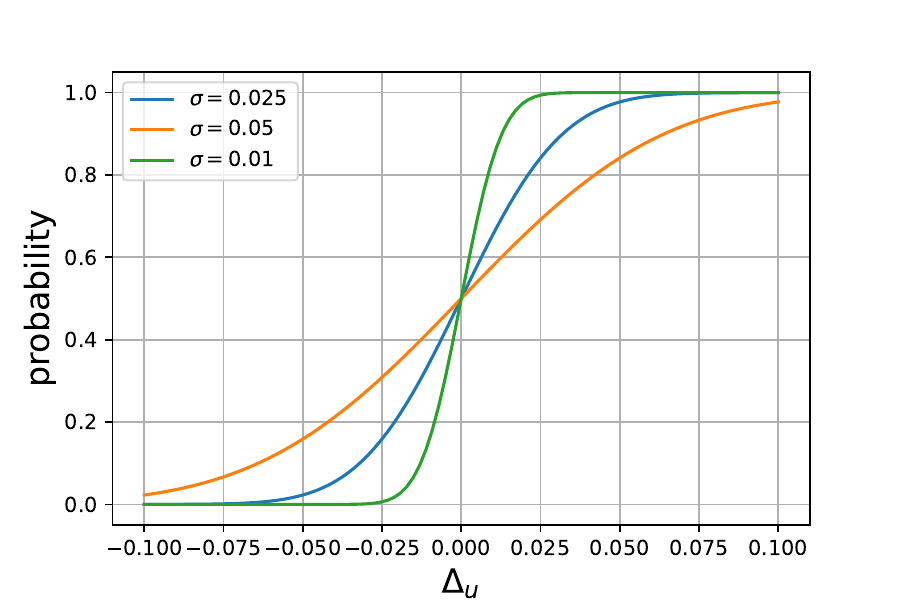}
\caption{$\Phi(\Delta_u/\sigma)$ for different values of $\sigma$.}
\label{fig:normcdf}
\end{figure}

This will result in wrong preference statements -- Alice may state   $\by \succ \bx$ when $u(\bx)>u(\by)$ -- overall leading to violations of asymmetry and negative transitivity. The probability of correctly stating $\bx \succ \by$  is a function of the difference $u(\bx)-u(\by)$, which can be  modelled by  the following likelihood:
\begin{equation}
    \label{eq:likelcdf0}
p(\bx \succ \by|u)=\Phi\left(\frac{u(\bx)-u(\by)}{\sigma}\right),
\end{equation}
where $\Phi(\cdot)$ is the Cumulative Distribution Function (CDF) of the 
standard Normal distribution. $\sigma>0$ is a scaling parameter, which plays a similar role to $\delta$ in Luce's model. As shown in Figure \ref{fig:normcdf}, for a given value of $\Delta_u=u(\bx)-u(\by)$, the probability of stating the correct preference decreases with $\sigma$.
In statistics, the  likelihood \eqref{eq:likelcdf0}  is known as \textit{probit}.

Hereafter we exploit a result derived in \citet{benavoli2021} to show that, assuming a GP prior on $u$ and the likelihood  \eqref{eq:likelcdf0}, the predictive posterior for $u$ is a Skew GP.  We recall that a skew Gaussian distribution \citep{azzalini2013skew} is an extension of the Gaussian distribution that includes additional parameters  to control asymmetry, allowing the distribution to be skewed rather than being perfectly symmetric. A Skew GP is a distribution over functions such that its marginals are  skew-Gaussian distributions. 
More details about the skew Gaussian distribution and the Skew GP are provided in Appendix \ref{app:skewgp}.

  \begin{SnugshadeB}
\begin{MethodB}[Probit for Erroneous Preferences]
\label{Model:Probit}
Consider a vector $X=[\bx_1,\bx_2,\dots,\bx_r]^\top$ of objects such that $\bx_i \in \mathbb{R}^c$ for all $i=1,\dots,r$, a training set of $m$ preferences  $\mathcal{D}_{m} = \{ \bx_{i_s}\succ \bx_{j_s}:~~ s = 1,\dots,m\}$ with $\bx_{i_s} \neq \bx_{j_s}$, $\bx_{i_s},\bx_{j_s} \in X$, 
for all $i_s,j_s \in\{1,2,\dots,r\}$. We denote by $u'(\cdot)=u(\cdot)/\sigma$. Under the assumption that the preference statements are conditionally independent given $u(X)$, 
from \eqref{eq:likelcdf0} we obtain that:
\begin{equation}
    \label{eq:likelcdf}
    p(\mathcal{D}_m|u'(X))=\prod_{s=1}^m \Phi\left(u'(\bx_{i_s})-u'(\bx_{j_s})\right)=\Phi_m(Wu'(X)),
\end{equation}
where $W$ has been defined  in Section \ref{sec:consistent} and $\Phi_m(\cdot)$ is the CDF of the standard $m$-dimensional multivariate normal distribution. 

Assuming $p(u') = \mathrm{GP}(u';0,k_{\boldsymbol{\theta}})$, the predictive posterior  $u'(X^*)$ for the test-points $X^*=[\bx_1,\dots,\bx_p]$ is a Skew GP of dimension $m$:
\begin{align}
\nonumber
p(u'(X^*)|\mathcal{D}_m,\boldsymbol{\theta})= \text{SkewGP}_{m}(&{\bf 0},K_{\boldsymbol{\theta}}(X^*,X^*), D_{K_{\boldsymbol{\theta}}(X^*,X^*)}^{-1}K_{\boldsymbol{\theta}}(X^*,X)W^T,\\
\label{eq:predPostSample}
&{\bf 0},W K_{\boldsymbol{\theta}}(X,X) W^T + I_m),
\end{align} 
where  $D_{K_{\boldsymbol{\theta}}(X^*,X^*)}$ is a diagonal matrix equal to the square-root of the diagonal of $K_{\boldsymbol{\theta}}(X^*,X^*)$. The distribution \eqref{eq:predPostSample} has  5 parameters, the first (zero) is a location parameter, the second  $K_{\boldsymbol{\theta}}(X^*,X^*)$ is the Kernel matrix, the last three parameters determine the skewness of the distribution.  Inference is carried out according to Algorithm \ref{alg:1}, where A1 is performed analytically to obtain \eqref{eq:predPostSample}; the Laplace's approximation is used in  A2  and MCMC (via lin-ess) in A3.
\end{MethodB}
\end{SnugshadeB}
It is worth pointing out that, even when Alice's preferences in $\mathcal{D}_m$ do not satisfy asymmetry and negative transitivity, the posterior is always well-defined in this case, because the model explicitly takes into account of the inconsistencies (errors) in Alice's preferences. 
It can be noted that the analytical derivation of the posterior, a Skew GP, allows us to perform step A1 in Algorithm \ref{alg:1} analytically. Moreover, step A3 can be performed efficiently and accurately, because sampling from a Skew GP can be reduced to sample independently from a multivariate Gaussian and, respectively, a truncated multivariate Gaussian \citep[Sec.\ 4]{benavoli2021}:
\begin{align}
\nonumber 
\text{Let } C_{*} &= K_{\boldsymbol{\theta}}(X^*,X)W^T ~\text{then} \\
\nonumber
u(X^*)&=\mathbf{r}_0+  C_{*} \left(W K_{\boldsymbol{\theta}}(X,X) W^T + I_m\right)^{-1}\mathbf{r}_{1},\\
\nonumber
\mathbf{r}_0 &\sim N\left({\bf 0},K_{\boldsymbol{\theta}}(X^*,X^*)-C_{*}\left(W K_{\boldsymbol{\theta}}(X,X) W^T + I_m\right)^{-1}C_{*}\right),\\
\nonumber
\mathbf{r}_1 &\sim \text{TN}_{\{u'(X)\geq {\bf 0}\}}\left({\bf 0},W K_{\boldsymbol{\theta}}(X,X) W^T + I_m\right).
\end{align}
We can  sample $\mathbf{r}_1$
using \textit{lin-ess}. As discussed in detail in  \citet{takeno2023towards}, the approximated posterior via sampling is more accurate than the Gaussian-based approximations (normally used in GP classification) computed via the Laplace's approximation, Expectation Propagation or Variational approximation.  In essence,  the
probit likelihood \eqref{eq:likelcdf} for preferences models the probability of a preference statement based on the
difference in latent utilities, which inherently leads to a skewed posterior distribution, making the Gaussian-based (i.e. symmetric) approximations of the posterior inaccurate.

After  rescaling $u'(\cdot)=\frac{u(\cdot)}{\sigma}$, the  only parameters in the model are the hyperparameters $\boldsymbol{\theta}$ in the kernel function. The scale  (variance) parameter $\sigma_k^2$ is  identifiable in this case, because the probability of correctly stating a preference $\bx_{i}\succ \bx_{j}$ is a function of the difference $u'(\bx_{i})-u'(\bx_{j})$. 
The  hyperparameters  $\boldsymbol{\theta}$ can be selected by maximising the marginal likelihood. 
The computation of
the marginal likelihood  requires the calculation of the CDF of the $m$-dimensional multivariate normal distribution. To approximate this integral, we can use the same approach described in Section \ref{sec:consistent}. However, since the likelihood is already a sigmoid-function (which is a smooth-indicator function\footnote{A sigmoid function can be thought as a  smooth-indicator function. In fact, an indicator function can only return values in the set $\{0,1\}$, while a sigmoid function is a smooth function that can return values in the interval $(0,1)$.}) we do not need to approximate it, as we did for the indicator function in Equation \eqref{eq:ML}. 

\begin{example}
\label{ex:6}
As before, we consider the set of temperatures  $\mathcal{X}=\{10,11,12,\dots,25\}$ and asks Alice about her preferences for $19$ pairs of temperatures. We generate Alice's preferences according to the model  in \eqref{eq:likelcdf0} with $u$ as in Figure \ref{fig:hometemp1} and $\sigma=0.05$.   This results in the dataset: 
\begin{align}
\nonumber
\mathcal{D}_{19}=\{&12\succ 10, 13\succ 25, 14\succ 13, 15\succ 19, 15\succ 22, 15\succ 23, 16\succ 11, \\ \nonumber
&19\succ 22, 19\succ 24, 20\succ 10, 20\succ 14, 20\succ 21, 20\succ 24, 20\succ 25,\\
\label{eq:erroneouspref}
&21\succ 13, 21\succ 19, 21\succ 25, 23\succ 25, 24\succ 25\}.
\end{align}
By comparing \eqref{eq:tempdataset} and \eqref{eq:erroneouspref}, it can be noted the following errors in Alice's preferences 
$15\succ 19$ and $21\succ 19$.
 In order to learn Alice's utility $u'$  from $\mathcal{D}_{19}$, we place a GP prior on $u'$ with a SE kernel and compute the predictive posterior as described in this section.

 Figure \ref{fig:th20err}-left shows the predictive posterior mean and 95\% credible interval for $u'$. The model is undecided about $u(20)-u(18)\stackrel{?}{\geq} 0$.  The uncertainty decreases if we consider 40 pairwise preferences as shown in Figure \ref{fig:th40err}: correctly the model states that $u(20)-u(18)\geq 0$ with probability $\approx 1$, that is we can infer that Alice prefers the temperature $20$ to $18$.

Figures \ref{fig:th40err_02} show that the model performs well also when preferences are generated using $\sigma=0.2$, which leads to  datasets $\mathcal{D}_m$ including many more errors.

\begin{figure}[h]
\centering  
\begin{subfigure}[b]{.99\linewidth}
\includegraphics[width=0.5\linewidth]{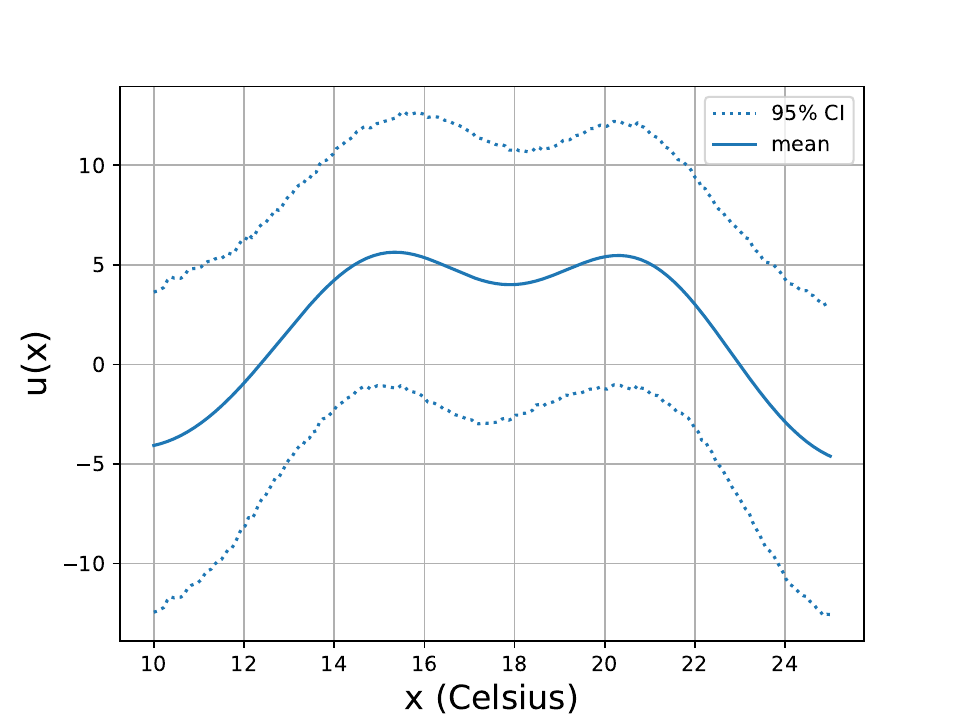}\includegraphics[width=0.5\linewidth]{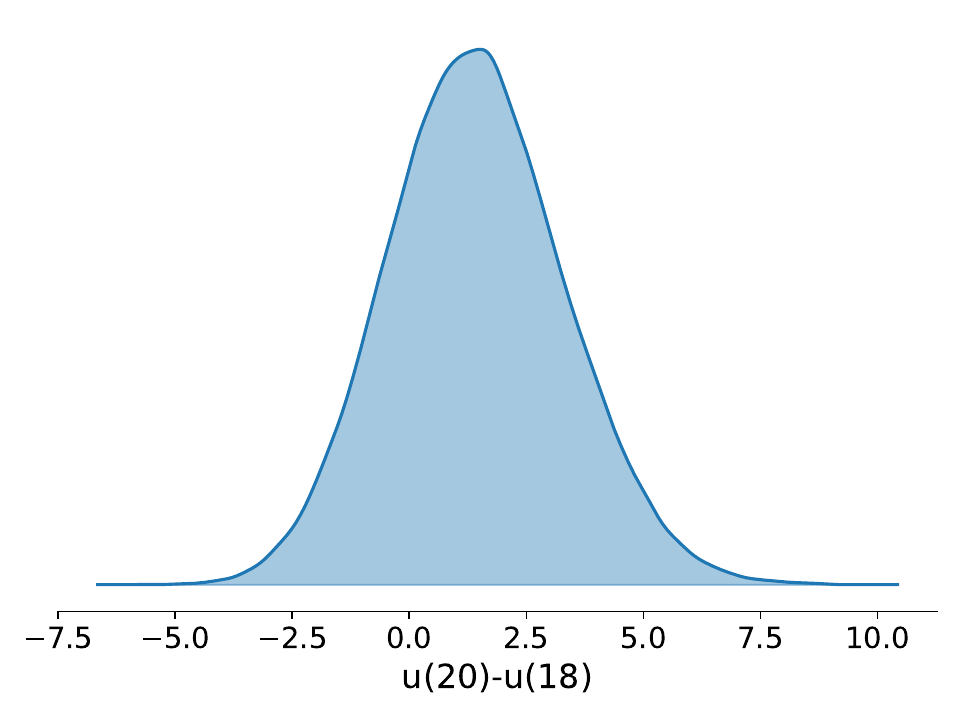}
\caption{20 pairwise preferences}\label{fig:th20err}
\end{subfigure}\\
\begin{subfigure}[b]{.99\linewidth}
\includegraphics[width=0.5\linewidth]{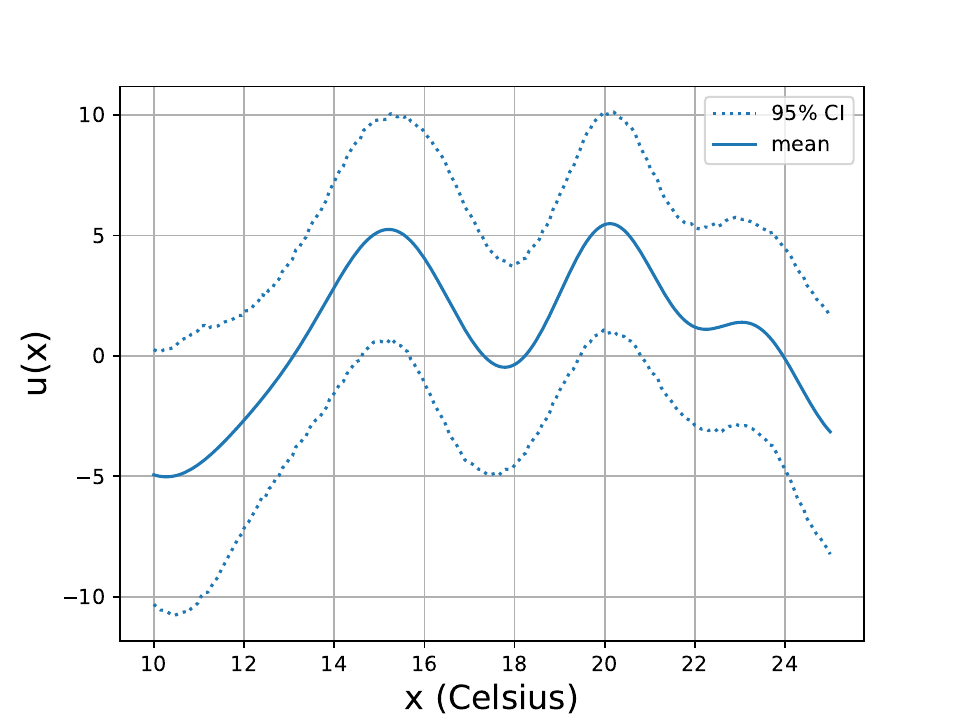}\includegraphics[width=0.5\linewidth]{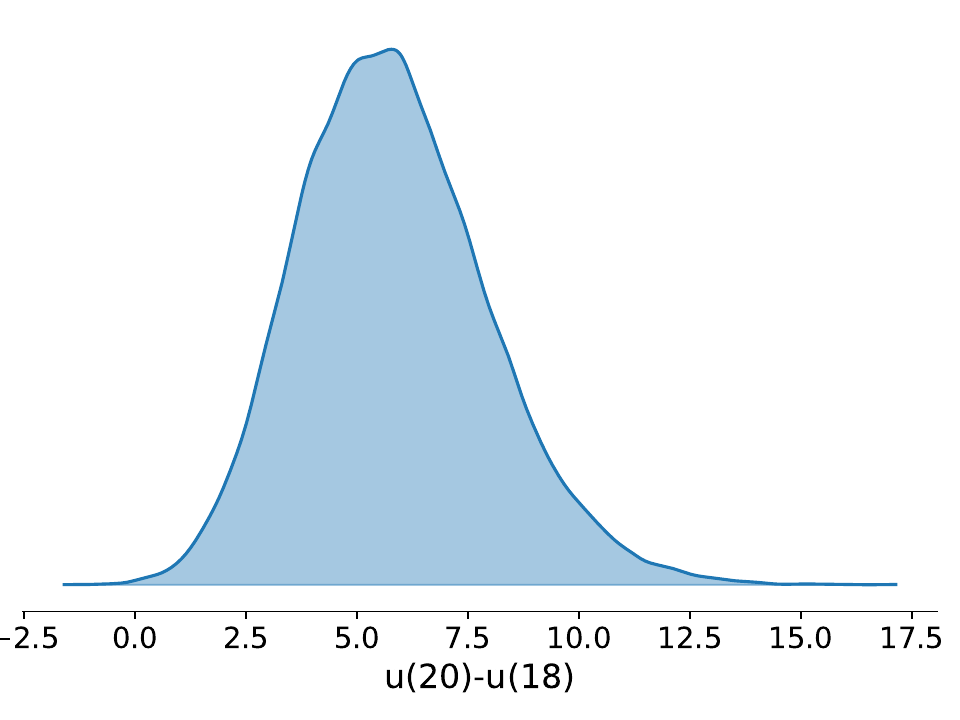} 
\caption{40 pairwise preferences}\label{fig:th40err}
\end{subfigure}\\
\caption{Left: learned utility functions under the probit model for erroneous preference, the mean is the continuous line and the region between the dotted-lines is the 95\% credible region. Right: posterior marginal of the difference $u(20)-u(18)$. The preferences were generated using $\sigma=0.05$.}
\label{fig:errpref005}
\end{figure}

\begin{figure}[h]
\centering
\begin{subfigure}[b]{.99\linewidth}
\includegraphics[width=0.5\linewidth]{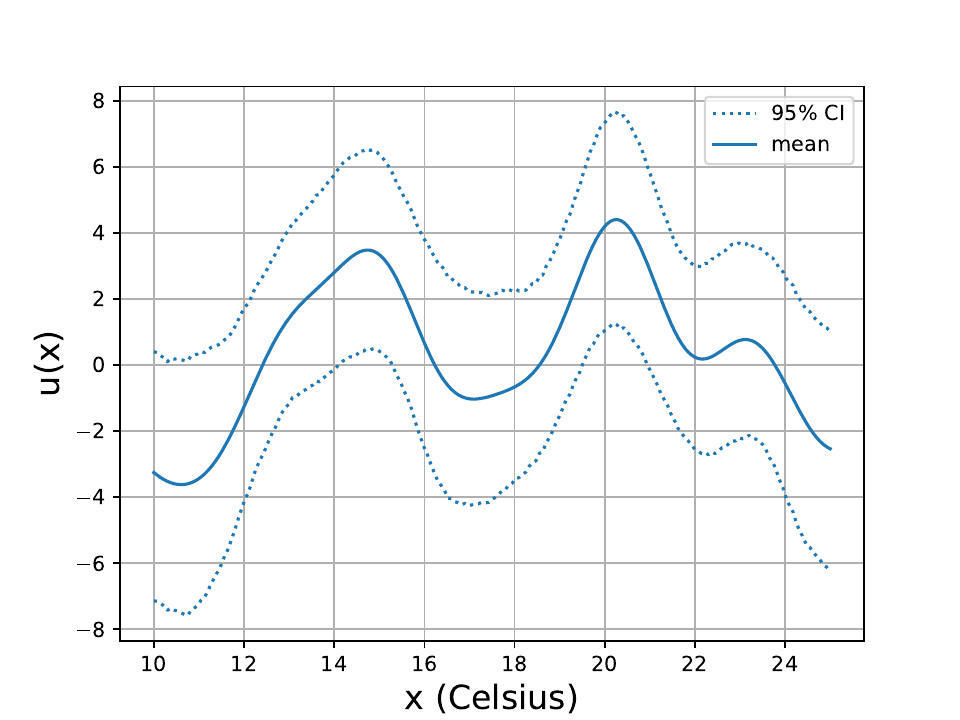}\includegraphics[width=0.5\linewidth]{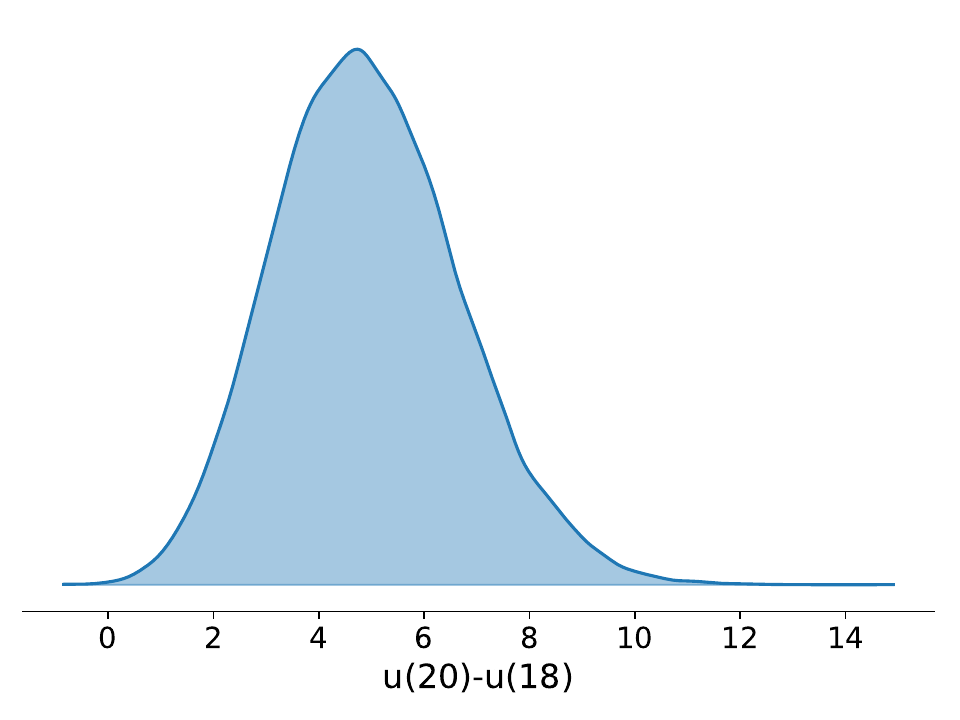}
\caption{40 pairwise preferences}\label{fig:th40err_02}
\end{subfigure}\\
\caption{Left: learned utility functions under the probit model for erroneous preference, the mean is the continuous line and the region between the dotted-lines is the 95\% credible region. Right: posterior marginal of the difference $u(20)-u(18)$. The preferences were generated with $\sigma=0.2$.}
\label{fig:errpref02}
\end{figure}
\end{example}

By comparing \eqref{eq:likeind0} and \eqref{eq:likelcdf}, it can be noticed that the indicator function in \eqref{eq:likeind0} has been replaced in \eqref{eq:likelcdf} by a sigmoid function. As a third alternative, in place of \eqref{eq:likelcdf} a logistic likelihood is also possible. However, this likelihood leads to a non-conjugate model and an analytical expression such as \eqref{eq:predPostSample} is not available.

\begin{remark}
 It is worth  emphasising the difference between Model \ref{Model:Probit} and the multinomial logit/probit models normally used in economics, see, e.g., \citet{mcfadden1974,mcfadden1978}.
These models commonly define a logit/probit likelihood   with a linear utility function, rather than assuming a GP prior and
using the likelihood in \eqref{eq:likelcdf}. The coefficients of the linear utility function are then estimated  by maximum likelihood. Note that this linear utility model will not fit  the data in Example \ref{ex:6}, because the underlying utility function is nonlinear. The advantage of Model \ref{Model:Probit} in Example \ref{ex:6} is that, by placing a GP prior with a squared-exponential kernel on the unknown utility, it can fit any nonlinear utility in a nonparametric way.  Moreover, Model \ref{Model:Probit} can also encompass many other models proposed in literature for a suitable choice of the kernel. For instance, by selecting a linear kernel \citep{rasmussen2006gaussian}, Model \ref{Model:Probit} would become  similar to the standard linear multinomial logit/probit model. With a spline kernel \citep{rasmussen2006gaussian}, it would become similar to the model proposed in \citet{kneib2007semiparametric}. The difference is that inference in Model \ref{Model:Probit} is carried out in the Bayesian framework instead of the frequentist framework (that is,   by maximum likelihood estimation). In Section \ref{sec:trasport}, we will provide an example of this flexibility and a numerical comparison with the standard linear multinomial probit model.
\end{remark}

\subsection{Convergence of the Probit model to the consistent preferences model}

In the case of erroneous preferences, the probit model introduced above has a special relationship with the consistent preferences model. The parameter $\sigma$ in \eqref{eq:likelcdf0} determines how ``erroneous'' the observed preferences are. For small values of $\sigma$ the probit and the consistent preferences model become very similar. This relationship can be formalised as follows.

\begin{proposition}
    Consider a vector $X=[\bx_1,\bx_2,\dots,\bx_r]^\top$ of objects such that $\bx_i \in \mathbb{R}^c$ for all $i=1,\dots,r$, a training set of $m$ preferences $
 \mathcal{D}_m = \{ \bx_{i_s} \succ \bx_{j_s}:~~ s = 1,\dots,m\}
 $ with $\bx_{i_s} \neq \bx_{j_s}$, $\bx_{i_s},\bx_{j_s} \in X$, 
$i_s,j_s \in\{1,2,\dots,r\}$ for each $s = 1,\dots,m$ and the likelihood of the probit model given by:
\begin{equation}
    \label{eq:likelcdfSigma}
    p(\mathcal{D}_m|u(X))=\prod_{s=1}^m \Phi\left(\dfrac{u(\bx_{i_s})-u(\bx_{j_s})}{\sigma}\right)=\Phi_m\left(W\dfrac{u(X)}{\sigma}\right),
\end{equation}
where we kept the non-scaled GP process $u$. Assuming the preferences are \textbf{consistent}, then the likelihood in \eqref{eq:likelcdfSigma} converges to the likelihood in \eqref{eq:likeind0box} as $\sigma \rightarrow 0$. 
\end{proposition}

\begin{proof}
For any $s \in \{1, \ldots, m\}$ we have
\begin{align*}
    \lim_{\sigma \rightarrow 0} \Phi\left(\dfrac{u(\bx_{i_s})-u(\bx_{j_s})}{\sigma}\right) &= \begin{cases}
    0 & \text{if } u(\bx_{i_s})-u(\bx_{j_s})<0 \\
    1 & \text{if } u(\bx_{i_s})-u(\bx_{j_s})>0
    \end{cases} \\
    &= I_{\{u(\bx_{i_s})-u(\bx_{j_s})>0\}}(u(X)),
\end{align*}
by the properties of the Gaussian CDF $\Phi$. Note that \\ $p(\mathcal{D}_m|u(X))=\prod_{s=1}^m \Phi\left(\dfrac{u(\bx_{i_s})-u(\bx_{j_s})}{\sigma}\right)$, thus we obtain the result.
\end{proof}

This result justifies the common practice of employing the probit-likelihood model even when the subject's preferences are known to be consistent.

\section{The Gaussian noise model}
\label{sec:gaussiannoise}
 In many situations, the observed  utility  $o(\bx),o(\by)$ of two objects $\bx,\by$ may differ from the true utility $u(\bx),u(\by)$ due to noise in the observation process.  This observation noise is commonly modelled by using  an additive Gaussian model \citep{thurstone2017law,ChuGhahramani_preference2005}, that is  $o(\bx)=u(\bx)+v_x$ and $o(\by)=u(\by)+v_y$, where $v_x,v_y$ are two independent and Gaussian distributed variables with zero mean and variance $\sigma^2$. Due to this noise, preferences can therefore violate asymmetry or negative transitivity or both. 

It is interesting to note that, for a preference statement, $\bx \succ \by$, the likelihood:
\begin{align}
\nonumber
p(\bx \succ \by|u(X))&=
\begin{aligned}
  \int I_{\{u(\bx)+v_x>u(\by)+v_y\}} (v_x,v_y)&N\left(v_x;0,\sigma^2\right) \cdot \\ &N\left(v_y;0,\sigma^2\right) dv_xdv_y \\
\end{aligned} \\ \label{eq:preferenceike1}
&=\Phi\left(\frac{u(\bx)-u(\by)}{\sqrt{2}\sigma}\right),
\end{align}
reduces to the likelihood model in \eqref{eq:likelcdf0}. The only distinction is the constant $\sqrt{2}$. However, the main difference between the erroneous-preference model and the Gaussian noise model becomes apparent when we compare more than two objects. Consider for instance the preferences $\bx \succ \by$ and $\bx \succ \bz$, the likelihood
\begin{align}
\nonumber
&p(\bx \succ \by, \bx \succ \bz|u(X)) \\ \nonumber
& \begin{aligned}
  =\int I_{\left\{\substack{u(\bx)+v_x>u(\by)+v_y,\\
  u(\bx)+v_x>u(\bz)+v_z}\right\}}(v_x,v_y,v_z) &N\left(v_x;0,\sigma^2\right)N\left(v_y;0,\sigma^2\right) \cdot \\
  &\cdot N\left(v_z;0,\sigma^2\right)dv_xdv_ydv_z
\end{aligned} \\ \label{eq:preferenceike2}
&\neq \Phi\left(\frac{u(\bx)-u(\by)}{\sqrt{2}\sigma}\right)\Phi\left(\frac{u(\bx)-u(\bz)}{\sqrt{2}\sigma}\right),
\end{align}
due to the common noise $v_x$ in the two preference statements. There is a straightforward way   to account for the common noise in the preference. We can simply use an augmented prior 
\begin{equation}
\label{eq:augm}
\begin{bmatrix}
u(X)\\
v(X)
\end{bmatrix} \sim N\left(
\begin{bmatrix}
{\bf 0}_m\\
{\bf 0}_m
\end{bmatrix}, \begin{bmatrix}
K_{\boldsymbol{\theta}}(X,X) & {\bf 0}_{m \times m}\\
{\bf 0}_{m \times m} & \sigma^2 I_m
\end{bmatrix}
\right)
\end{equation}
with $v(X)=[v_{x_1},\dots,v_{x_m}]^\top$ and consider the indicator likelihood.

  \begin{SnugshadeB}
\begin{MethodB}[Preferences with Gaussian Noise Error]
\label{model:Gaussian}
Consider a vector $X=[\bx_1,\bx_2,\dots,\bx_r]^\top$ of objects such that $\bx_i \in \mathbb{R}^c$ for all $i=1,\dots,r$ and a training set of $m$ preferences $
 \mathcal{D}_m = \{ \bx_{i_s} \succ \bx_{j_s}:~~ s = 1,\dots,m\}
 $ with $\bx_{i_s} \neq \bx_{j_s}$, $\bx_{i_s},\bx_{j_s} \in X$, 
$i_s,j_s \in\{1,2,\dots,r\}$ for each $s = 1,\dots,m$. Assuming the preferences are \textbf{noisy} due to an additive Gaussian noise. Then, the probability of observing the data $ \mathcal{D}_m $ given the utility function vector $u(X)$ and the noise vector $v(X)$ is equal to the likelihood function:
\begin{equation}
\label{eq:likeindnoisecomb}
\begin{aligned}
p(\mathcal{D}_m|u(X),v(X))&=\prod_{s=1}^m I_{\{u(\bx_{i_s})+v_{i_s}-u(\bx_{j_s})-v_{j_s}>0\}}(u(X),v(X))\\
    &=I_{\left\{\widetilde{W}\begin{bmatrix} u(X)\\v(X)\end{bmatrix} >0\right\}}(u(X),v(X)),
\end{aligned}
\end{equation}
where $\widetilde{W}$ is a suitable matrix which allows to rewrite the inequalities in the indicator in matrix-form.
Assuming a GP prior $p(u) = \mathrm{GP}(u;0,k_{\boldsymbol{\theta}})$, the predictive posterior for $u(X^*)$ at the test points $X^*=[\bx_1,\dots,\bx_p]$ is
\begin{align}
    \label{eq:posteriorexactpredgaussian}
&p(u(X^*)|\mathcal{D}_m,\boldsymbol{\theta})= \\ \nonumber
    & \begin{aligned}
=\int &p(u(X^*)|u(X))\\
&TN_{\left\{\widetilde{W}\begin{bmatrix} u(X)\\v(X)\end{bmatrix} >0\right\}}\left(\begin{bmatrix}
u(X)\\
v(X)
\end{bmatrix};\begin{bmatrix}
{\bf 0}_m\\
{\bf 0}_m
\end{bmatrix},\begin{bmatrix}
K_{\boldsymbol{\theta}}(X,X) & {\bf 0}_{m \times m}\\
{\bf 0}_{m \times m} & \sigma^2 I_m
\end{bmatrix}\right)du(X)dv(X)
    \end{aligned}
\end{align}
where $p(u(X^*)|u(X))$ is defined in Proposition \ref{eq:prepost}. Inference is carried out according to Algorithm \ref{alg:1}, where  the Laplace's approximation is used in  A2  and MCMC (lin-ess) in A3.
\end{MethodB}
\end{SnugshadeB}
Note that, the above model is structurally similar to Model \ref{Model:1}. Therefore, we can use the same approach to compute the predictive distribution \eqref{eq:posteriorexactpredgaussian} via lin-ess and to learn the hyperparameters $\boldsymbol{\theta}$, including $\sigma^2$.

Finally, observe that, if the pairs in the preference dataset are unique (the same object is only compared once), this model is equivalent to Model \ref{Model:Probit}.

\begin{example}
\label{ex:7}
We consider the set of temperatures $\mathcal{X}=\{10,11,12,\dots,25\}$ and, for each temperature, we generated a  realisation of observed temperature
$o(x_i)=u(x_i)+v(x_i)$, with $u$ as in Figure \ref{fig:hometemp1} and $v(x_i)$ sampled from independent Gaussian distributions with $\sigma=0.1$. We then generated $80$ random pairwise preferences (without repetitions)  as follows: $x_i \succ x_j$ if $o(x_i)>o(x_j)$. This could represent a situation where drafts in the house perturb Alice's utility.

We compared two models.  Figure \ref{fig:errprefGauss}-left displays the posterior utility (mean and credible interval) obtained from Model \ref{model:Gaussian}, which accounts for the common noise in the preference statements. Figure \ref{fig:errprefGauss}-right shows the posterior utility derived from Model \ref{Model:Probit}, which ignores the common noise. We can see that Model \ref{Model:Probit} wrongly fits the noise and produces an incorrect estimate of the function (compare it with the true utility in Figure \ref{fig:hometemp1}). Moreover, the uncertainty is clearly underestimated. This demonstrates that modelling the common noise correctly is very important in practice.

\begin{figure}[h]
\centering
\includegraphics[width=0.5\linewidth]{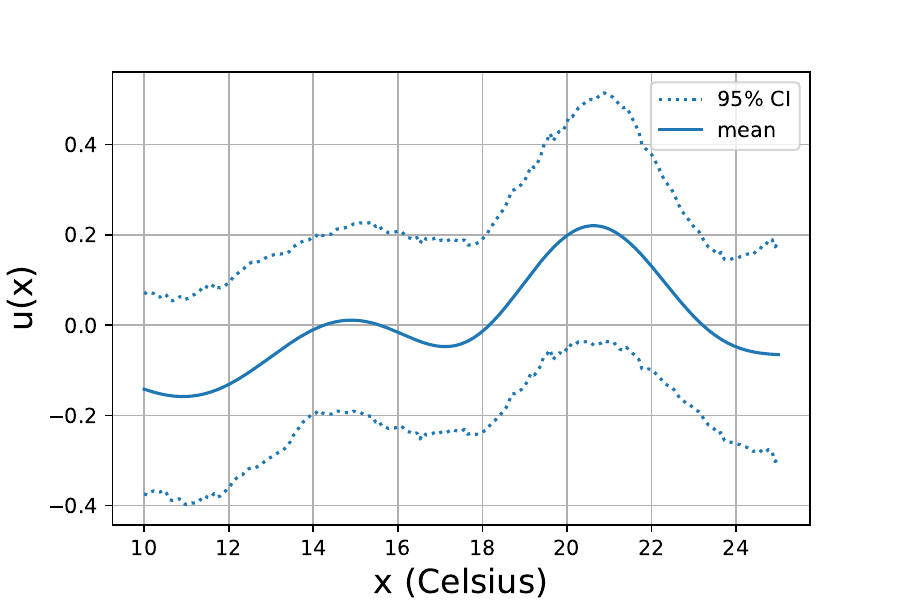}\includegraphics[width=0.5\linewidth]{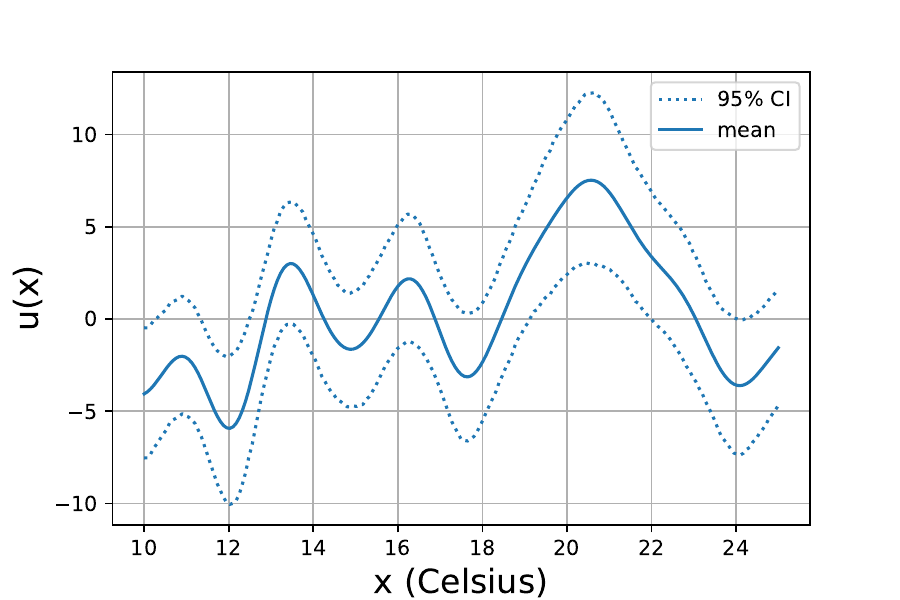}
\caption{Left: learned utility functions under the Gaussian-noise model for preferences, the mean is the continuous line and the region between the dotted-lines is the 95\% credible region. Right: learned utility functions under the Probit model for erroneous preferences.}
\label{fig:errprefGauss}
\end{figure}

\end{example}
In the likelihood \ref{eq:preferenceike1}, we  assumed a constant variance for the noise (homoscedastic model). A heteroscedastic model for preference learning has been proposed in \citet{sinaga2024heteroscedastic}.
Finally, it is worth mentioning that noisy preferences subject to Gumbel noise lead to a logistic likelihood. We will discuss this in Section \ref{sec:PL}.

\section{Multiple implicit utility functions}
\label{sec:multipleu}

In some applications, there may be a variety of attributes to consider while comparing objects.  Therefore, 
Alice's preference $\bx \succ \by$ may be determined by \textit{multiple utility functions} $u_1,\dots,u_d$ taking into account different characteristics of the objects under consideration:
$$
u_1(\bx),\dots,u_d(\bx) ~\text{ versus }~ u_1(\by),\dots,u_d(\by).
$$

We assume that both the functions $u_i$ and the dimension $d$ are unknown and consider two cases.

\begin{enumerate}
\itemsep0.1em
    \item When expressing her judgements, Alice is able to aggregate the multiple utility functions into a single dimension. For instance, she can implicitly consider the additive aggregation: 
        $$
u(\bx)=u_1(\bx)+\dots+u_d(\bx).
$$
By assuming that the aggregated function $u$ is the unknown utility, we are back to the single utility  model. 
\item Alice judges an object to be better than another object  if for instance it is better with respect to all attributes (Pareto efficiency), leading to the relations:
\begin{align}
\label{eq:paretosucc}
\bx \succ \by \text{ iff }& \min_{i}(u_i(\bx)-u_i(\by))>0,\\   
\label{eq:paretoconfl}
\bx \frown \by \text{ iff }& \max_{i}(u_i(\bx)-u_i(\by))>0 ~\&~ \min_{i}(u_i(\bx)-u_i(\by))<0, 
\end{align}
where $\bx \frown \by$ means that there is a  \textit{conflict}, that is neither $\bx$ nor $\by$ is judged to be better with respect to all attributes.
\end{enumerate}
The relation $\succ $ in \eqref{eq:paretosucc} is asymmetric but is not negatively transitive. The relation $\frown$ is not an indifference relation, because is not transitive. 

We can extend Model \ref{Model:Probit} to handle multiple utilities.
Specifically, we replace the strict statements \eqref{eq:paretosucc}--\eqref{eq:paretoconfl} with the probit likelihood to account for errors due to Alice's inability to distinguish two objects that have similar utility $u_i$ for $i=1,\dots,d$.
This leads to the following likelihood:
\begin{align}
\label{eq:paretolike}
  p(\bx \succ \by|u_1,\dots,u_d)&=\prod_{i=1}^d \Phi\left(\frac{u_i(\bx)-u_i(\by)}{\sigma}\right),\\
  \label{eq:paretolikeconflict}
   p(\bx \frown \by|u_1,\dots,u_d)&=1-\prod_{i=1}^d \Phi\left(\frac{u_i(\bx)-u_i(\by)}{\sigma}\right)-\prod_{i=1}^d \Phi\left(\frac{u_i(\by)-u_i(\bx)}{\sigma}\right).
\end{align}
Note that, \eqref{eq:paretolike} and \eqref{eq:paretolikeconflict} are probabilistic relaxations of  \eqref{eq:paretosucc} and, respectively, \eqref{eq:paretoconfl}. 

We will defer the discussions on multiple utilities for preferences over objects until Section \ref{sec:choice}, where we will introduce a more general model based on \textit{choice functions}.

\section{Learning preferences via a two-argument function}
\label{sec:twoargument}
A binary relation $R$ on $\mathcal{X} \times \mathcal{X}$ can be represented through a two-argument function
$q : \mathcal{X} \times \mathcal{X} \rightarrow \mathbb{R}$. If $(\bx,\by)\in R$, denoted as $\bx R\by$, then $q(\bx,\by)>0$. Since in general $q(\bx,\by)\neq q(\by,\bx)$ we can equivalently write $q(\bx,\by)$ as a function of the vector $[\bx,\by]$, $q([\bx,\by])$.
The function $q$ has been interpreted as a ``strength of preference''  \citep{shafer1974nontransitive,fishburn1988nonlinear}, with values of $q([\bx,\by])$ close to zero indicating a difficult
decision (the decision maker cannot distinguish $\bx,\by$).

This is a natural generalisation of representation results for consistent preferences, in which case
one can set $q([\bx,\by]) = u(\bx)-u(\by)$ for a utility function $u$.
In this case, from the asymmetry property of preferences, we can derive that $q$ must be a  skew-symmetric function $q([\bx,\by]) =-q([\by,\bx])$. 
Moreover, $q$ must satisfy negative transitivity: if  $q([\bx,\by])>0$ then for any other $\bz \in \mathcal{X}$ either $q([\bx,\bz])>0$ or $q([\bz,\by])>0$ or both. This follows by $q([\bx,\by]) = u(\bx)-u(\by) = (u(\bx)-u(\bz))-(u(\by)-u(\bz))$.

In the case $q([\bx,\by]) = u(\bx)-u(\by)$, we can observe that a GP prior on $u$ induces a GP prior on $q$, because Gaussianity is preserved by affine transformations.
It is easy to show that if $u \sim GP(0,k_{\boldsymbol{\theta}}(\bx,\bx'))$, then $q \sim GP(0,k_{pref}([\bx,\by],[\bx',\by']))$ where
\begin{equation}
    \label{eq:prefkernel}
k_{pref}([\bx,\by],[\bx',\by'])=    k_{\boldsymbol{\theta}}(\bx,\bx')+k_{\boldsymbol{\theta}}(\by,\by')-k_{\boldsymbol{\theta}}(\bx,\by')-k_{\boldsymbol{\theta}}(\by,\bx'),
\end{equation}
is the so-called \textit{preference induced kernel} \citep{houlsby2011bayesian}.
It is worth noticing that the prior model $q \sim GP(0,k_{pref}([\bx,\by],[\bx',\by']))$ puts zero mass on functions  $q$ violating asymmetry and negative transitivity. In fact, assume that 
$$
X=\begin{bmatrix}
   \bx & \by\\
   \by & \bx
  \end{bmatrix},
$$
then  $K_{pref}(X,X)$ is equal to:
\begin{align*}
&E\begin{bmatrix}
q([\bx,\by])q([\bx,\by])& 
q([\bx,\by])q([\by,\bx])\\
q([\by,\bx])q([\bx,\by]) & q([\by,\bx])q([\by,\bx]) 
\end{bmatrix} \\
&=\begin{bmatrix}
1& 
-1\\
-1 & 1
\end{bmatrix}(k_{\boldsymbol{\theta}}(\bx,\bx)+k_{\boldsymbol{\theta}}(\by,\by)-k_{\boldsymbol{\theta}}(\bx,\by)-k_{\boldsymbol{\theta}}(\by,\bx)).
\end{align*}
The covariance matrix has rank one and correlation coefficient equal to $-1$, which implies that $q([\bx,\by])=-q([\by,\bx])$. 
Therefore, $k_{pref}$ puts zero mass on functions  $q$ violating asymmetry. Similarly, consider the following data-matrix
$$
X=\begin{bmatrix}
   \bx & \by\\
   \by & \bz\\
   \bx & \bz
  \end{bmatrix},
$$
this would correspond to the preference of a subject who prefers $\bx \succ \by$, $\by \succ \bz$ and $\bx \succ \bz$. The covariance matrix $K_{pref}(X,X)$ is equal to:
$$
 \resizebox{0.91\hsize}{!}{$
\left[\begin{smallmatrix}
 k(\bx,\bx)+k(\by,\by)-k(\bx,\by)-k(\bx,\by) & k(\bx,\by)+k(\by,\bz)-k(\bx,\bz)-k(\by,\by)& k(\bx,\bx)+k(\by,\bz)-k(\bx,\bz)-k(\bx,\by)\\
k(\bx,\by)+k(\by,\bz)-k(\bx,\bz)-k(\by,\by) & k(\by,\by)+k(\bz,\bz)-k(\by,\bz)-k(\by,\bz)& k(\by,\bx)+k(\bz,\bz)-k(\by,\bz)-k(\bx,\bz)\\
   k(\bx,\bx)+k(\by,\bz)-k(\bx,\bz)-k(\bx,\by) &  k(\by,\bx)+k(\bz,\bz)-k(\by,\bz)-k(\bx,\bz)& k(\bx,\bx)+k(\bz,\bz)-k(\bx,\bz)-k(\bz,\bx)\\
\end{smallmatrix}\right]$},
$$
where we omitted the subscript $\boldsymbol{\theta}$ in $k_{\boldsymbol{\theta}}$ to simplify the notation.
It is immediate to verify that the above matrix has rank 2 (the first column is equal to the third column minus the second column), which shows that $\bx \succ \bz$ is implied by $\bx \succ \by$ and $\by \succ \bz$.

By using $q \sim GP(0,k_{pref}([\bx,\by],[\bx',\by']))$, we can equivalently reformulate Model \ref{Model:Probit} as a GP classification problem with 
$$
X=\begin{bmatrix}
   \bx_{i_1} & \bx_{j_1}\\
   \bx_{i_2} & \bx_{j_2}\\
   \vdots & \vdots\\
   \bx_{i_m} & \bx_{j_m}\\
  \end{bmatrix},~~~ Y=\begin{bmatrix}
   1\\
   1\\
   \vdots \\
   1\\
  \end{bmatrix},
$$
where $X$ is the preference matrix, where each row states a preference for the left object over the right object. $Y$ is the class-label vector, which is always equal to one because we arranged the elements in each row of $X$ such that the left element is preferred the right one.

  \begin{SnugshadeB}
\begin{MethodB}[Probit Model for Erroneous Preferences as a Classification Problem]
\label{model:Probitclassification}
Consider the matrices:
\begin{equation}
\label{eq:encodingForClassifier}
    X=\begin{bmatrix}
   \bx_{i_1} & \bx_{j_1}\\
   \bx_{i_2} & \bx_{j_2}\\
   \vdots & \vdots\\
   \bx_{i_m} & \bx_{j_m}\\
  \end{bmatrix},~~~ Y=\begin{bmatrix}
   1\\
   1\\
   \vdots \\
   1\\
  \end{bmatrix},    
\end{equation}
with $\bx_{i_s} \neq \bx_{j_s}$, $\bx_{i_s},\bx_{j_s} \in [\bx_1,\dots,\bx_r]$, 
for all $i_s,j_s \in\{1,2,\dots,r\}$
and $s\in\{1,2,\dots,m\}$. We model the preferences with the function $q$, so that $q([\bx,\by])>0$ if $\bx \succ \by$.

 Under the assumption that the preference statements in $X$ are conditionally independent given $q(X)$, we obtain that:
\begin{equation}
    \label{eq:likelcdf1}
    p(Y|q(X))=\prod_{s=1}^m \Phi\left(q([\bx_{i_s},\bx_{j_s}])\right)=\Phi_m(q(X)).
\end{equation}

Assuming $q \sim GP(0,k_{pref}([\bx,\by],[\bx',\by']))$, the predictive posterior  $q(X^*)$ for $X^*=[\bx_{ia},\bx_{ib}]$ (that is, the prediction for the strength of the preference  $\bx_{ia} \succ \bx_{ib}$)  is a Skew GP of dimension $m$ \citep[Sec.\ 3.2]{benavoli2021}:
\begin{align}
\nonumber
p(q(X^*)|\mathcal{D}_m,\boldsymbol{\theta})= \text{SkewGP}_{m}(&{\bf 0},K_{pref}(X^*,X^*), D_{K_{pref}(X^*,X^*)}^{-1}K_{pref}(X^*,X),\\
\label{eq:predPostSample1}
&{\bf 0},K_{pref}(X,X)  + I_m),
\end{align} 
where  $D_{K_{pref}(X^*,X^*)}$ is a diagonal matrix equal to the square-root of the diagonal of $K_{pref}(X^*,X^*)$. Inference is carried out according to Algorithm \ref{alg:1}, where  the Laplace's approximation is used in  A2  and MCMC (lin-ess) in A3.
\end{MethodB}
\end{SnugshadeB}
Model \ref{Model:Probit} and Model \ref{model:Probitclassification} are equivalent. The only advantage of Model \ref{model:Probitclassification}  is that, by framing preference learning as a classification problem, we can directly use methods and software developed in GP classification. For instance, by using variational approximation   \citep{hensman2015scalable,bui2017unifying}, one can easily scale this model to large datasets.

Model~\ref{model:Probitclassification} frames the preference learning problem as a classification task, but it remains a proper preference model due to the use of the kernel  $k_{pref}$. In contrast, training a standard classification model (such as a GP classifier with an SE kernel or a  neural network) typically leads to predictions that violate transitivity. The example below demonstrates numerically that such violations are common in practice.

\begin{example}
\label{ex:violationTransitivity}
We revisit the example of Alice’s home comfort, this time assuming that her utility is a function of both temperature and humidity. The goal is to demonstrate that the models proposed in this section can straightforwardly be applied in a multidimensional setting, while also providing a more challenging example for the comparison with standard classifiers we aim to carry out. Figure \ref{fig:comfort2d} shows a heat map of the true utility function.  

\begin{figure}[h]
\centering
\includegraphics[width=7cm]{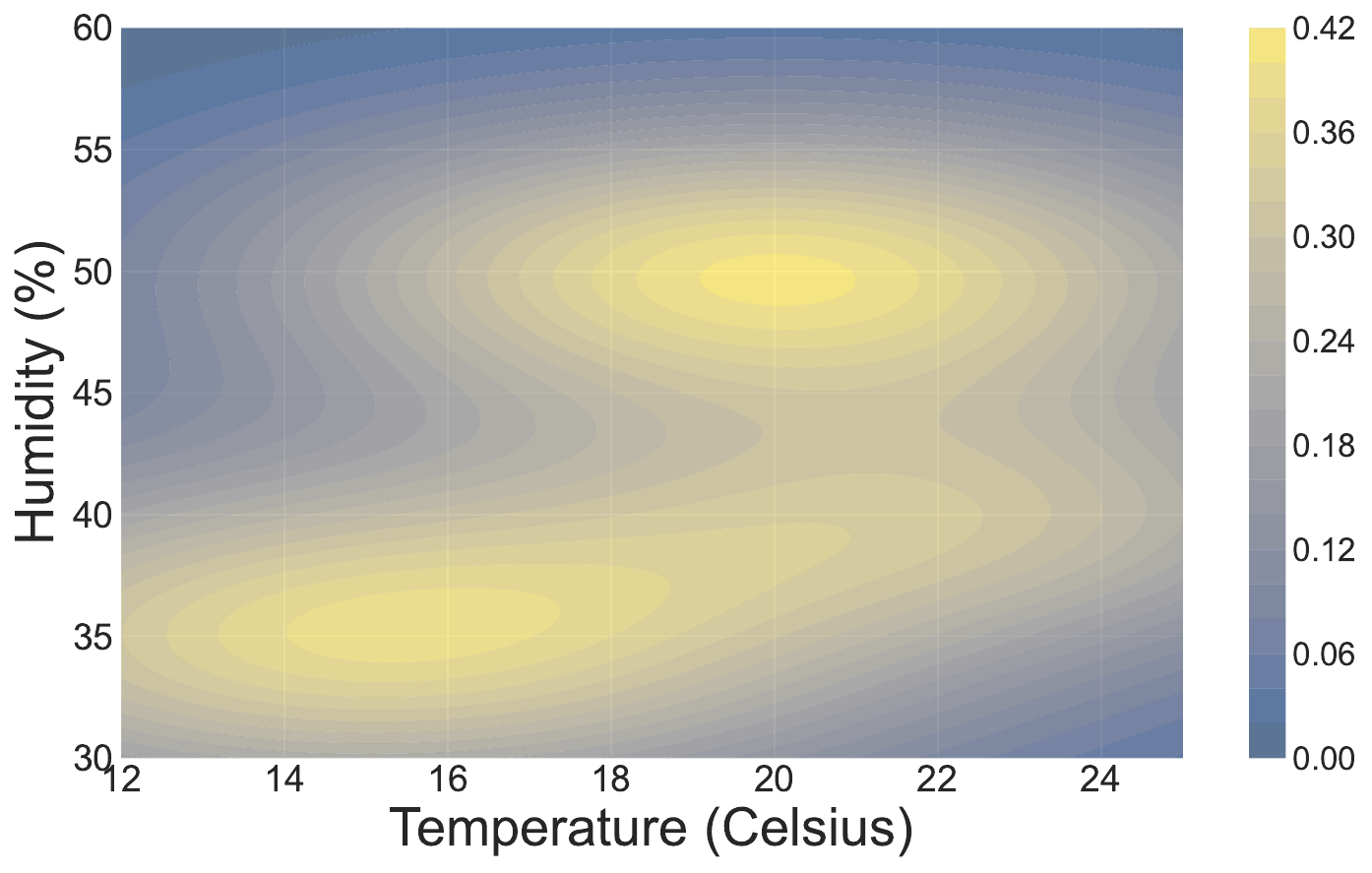}
\caption{Alice's utility as a function of temperature and humidity. There are 3 local maxima at $[15,35]$, $[20,50]$ and $[23,40]$.}
\label{fig:comfort2d}
\end{figure}

The objects are two-dimensional vectors of temperature and humidity, e.g. $\bx_1 = [20,35] \in \mathbb{R}^2$. From the set $\mathcal{X} = \{12, 13, 14, \ldots, 24,25\} \times \{30, 35, \ldots, 55, 60\}$, we ask Alice to evaluate $130$ pairs of vectors, resulting in the dataset 
\begin{equation}
    \mathcal{D}_{130} = \left\{ \begin{bmatrix}
        12 \\ 30
    \end{bmatrix} \succ \begin{bmatrix}
        18 \\ 60
    \end{bmatrix},
    \begin{bmatrix}
        15 \\ 35
    \end{bmatrix} \succ \begin{bmatrix}
        22 \\ 40
    \end{bmatrix},
    \ldots,
    \begin{bmatrix}
        21 \\ 50
    \end{bmatrix} \succ \begin{bmatrix}
        24 \\ 55
    \end{bmatrix}\right\}.
\end{equation}
As in the previous examples, we aim to infer Alice's utility from those preference observations and to predict her preferences on pairs which were not observed. 

Our goal is to compare Model~\ref{model:Probitclassification} 
with two standard classifiers:  a GP-based and  a neural network classifier.  
For the two classifiers, we encode the preferences in $\mathcal{D}_{130}$ as in~\eqref{eq:encodingForClassifier}. However, to enforce asymmetry,  we further augment the data by taking all pairs in the reverse order and assigning to that input the negative class. Therefore, the overall training set is:
$$
X=\begin{bmatrix}
   \bx_{i_1} & \bx_{j_1}\\
   \bx_{i_2} & \bx_{j_2}\\
   \vdots & \vdots\\
   \bx_{i_m} & \bx_{j_m}\\
      \bx_{j_1} & \bx_{i_1}\\
   \bx_{j_2} & \bx_{i_2}\\
   \vdots & \vdots\\
   \bx_{j_m} & \bx_{i_m}
  \end{bmatrix},~~~ Y=\begin{bmatrix}
   1\\
   1\\
   \vdots \\
   1\\
    0\\
   0\\
   \vdots \\
   0\\
  \end{bmatrix}.
$$
Then, the probability of the predicted preference for a new pair of objects $\bx_{i_s}, \bx_{j_s}$ is computed as    $0.5 p(y=1|[\bx_{i_s}, \bx_{j_s}])+0.5 p(y=0|[\bx_{j_s}, \bx_{j_s}]))$, see for instance \citep[Section 3.3]{chau2022learning}.\footnote{As discussed in \citep[Section 3.3]{chau2022learning}, the theoretical justification of  this asymmetry-enforcing approach is    questionable,
and the additional computational cost due to doubling
the data size may be problematic.}
The GP classifier uses a standard SE kernel and it is trained with variational inference \citep{Gardner_2018_gpytorch}. The neural network classifier is a full Multi-Layer Perceptron (MLP) with two layers of 5 neurons each and rectifier (RELU) activation function. We train the MLP by optimizing the log-loss function.

We compare the three models in terms of accuracy on a test set of $60$ pairs. We repeat the experiment $50$ times generating random pairs for each repetition. The average accuracy over all repetitions is reported in the first line of Table~\ref{tab:resComfort}.

{
\rowcolors{2}{gray!25}{white}
\begin{table}
\centering
\small
\begin{tabular}{|c|c|c|c|}
\rowcolor{gray!50}
 \hline
 & Model~\ref{model:Probitclassification} & GP classifier SE Kernel & MLP \\
 \hline
Accuracy & 0.95 & 0.86 & 0.85  \\
Transitive correctly predicted & 1.0 & 0.94 & 0.94 \\
  \hline
\end{tabular}
\caption{Prediction accuracy for the Alice's temperature-humidity comfort dataset. Results averaged over $50$ repetitions.}
\label{tab:resComfort}
\end{table}}

To assess that the difference in accuracy between Model~\ref{model:Probitclassification} and the other classifiers   is statistically and practically significant, we use the (nonparametric) Bayesian signed-rank test \citep{benavoli14}, which considers two classifiers practically equivalent if their difference in accuracy is less than $0.01$ ($1\%$). The interval $[-0.01, 0.01]$ thus defines a region of practical equivalence (rope) for classifiers. The test returns the posterior probability of the vector\footnote{The meaning of the three terms is: (i) Classifier 1 has higher accuracy than classifier 2 (more than 1\% difference); (ii) Classifier 1 and 2 are practically equivalent (difference in accuracy less than 1\%); (iii) Classifier 2  has higher accuracy than  classifier 1 (more than 1\% difference).} $[p(Cl_1 > Cl_2), p(Cl_1 \approx  Cl_2), p(Cl_1 < Cl_2)]$ and, therefore, this posterior can be visualised in the probability simplex, as shown in Figure ~\ref{fig:ropes}. Model~\ref{model:Probitclassification} is always better than a standard GP classifier (with probability $1$). The GP classifier  is never worse than the MLP classifier and it is better with probability $0.35$. 
Therefore,  Model~\ref{model:Probitclassification} outperforms both the two standard classifiers as expected.

\begin{figure}[h]
\centering
\includegraphics[width=0.495\textwidth]{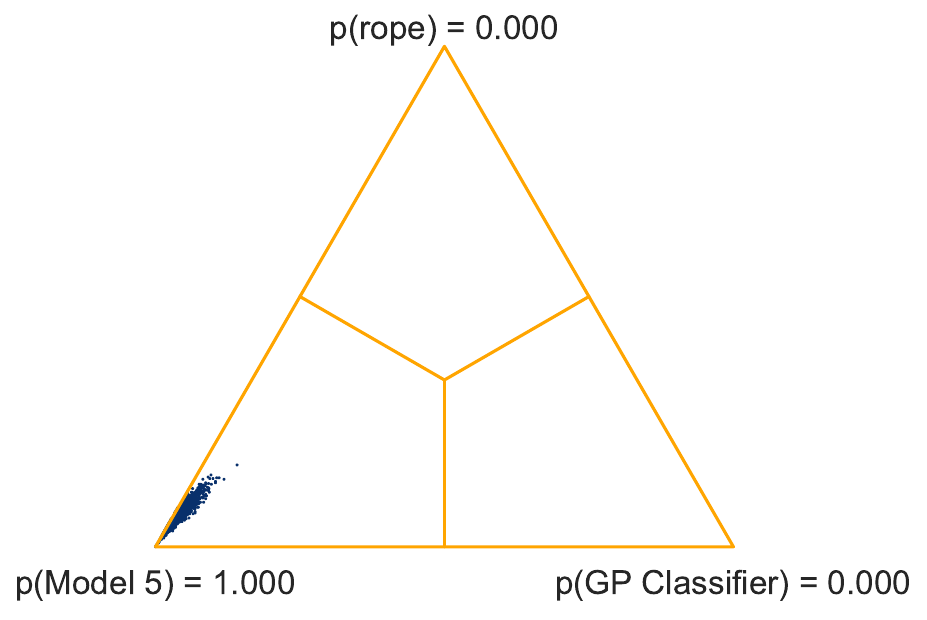}
\includegraphics[width=0.495\textwidth]{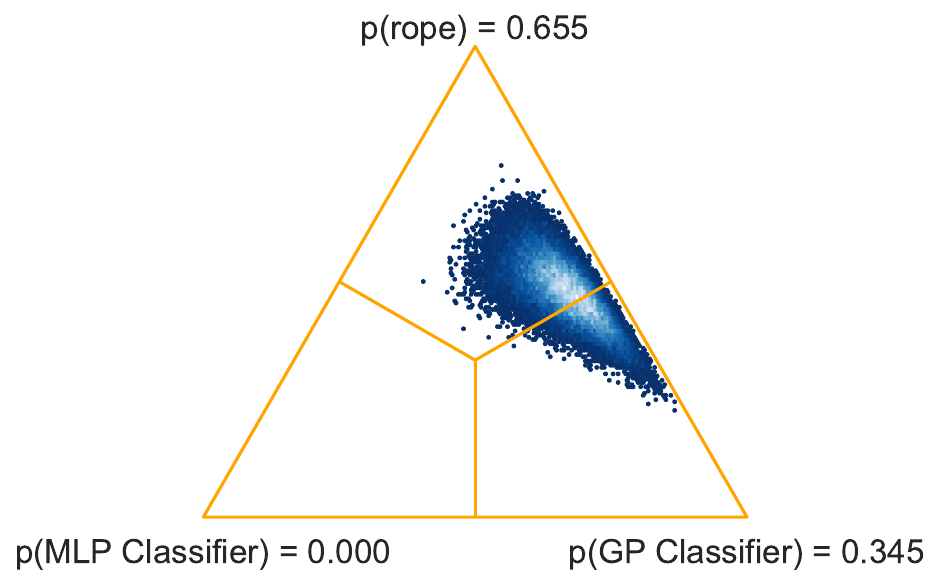}
\caption{Bayesian Wilcoxon signed-rank test with rope $[-0.01, 0.01]$.}
\label{fig:ropes}
\end{figure}

We further evaluate if the models satisfy transitivity by considering triplets of preferences that imply transitivity, e.g. $$\begin{bmatrix}
        21 \\ 50
    \end{bmatrix} \succ \begin{bmatrix}
        18 \\ 35
    \end{bmatrix} \text{ and } \begin{bmatrix}
        18 \\ 35
    \end{bmatrix} \succ \begin{bmatrix}
        24 \\ 55
    \end{bmatrix} \Rightarrow \begin{bmatrix}
        21 \\ 50
    \end{bmatrix} \succ \begin{bmatrix}
        24 \\ 55
    \end{bmatrix}.$$
    
We keep the first two elements of each triplet in the training set and evaluate whether the final preference is correctly predicted in the test set. If a model accurately encodes transitivity, we expect all preferences in the test set to be correctly predicted. The second row of Table~\ref{tab:resComfort} reports the average proportion of correctly predicted preferences over the $50$ repetitions. As expected, Model~\ref{model:Probitclassification} correctly predicts all preferences that result from a transitive triplet. The other two models achieve a high proportion of correctly evaluated preferences, however they do not guarantee that transitivity is respected.

\end{example}

\subsection{Intrinsically nontransitive preferences}
In Section \ref{sec:problems}, we discussed three possible reasons why a subject's preferences may fail to satisfy asymmetry and/or negative transitivity: (i) limit of discernibility; (ii) noise; (iii) incompleteness due to multiple utilities.
In all these cases, the underlying model is rational, in the sense that there are underlying utility functions, which model the subject's preferences. By exploiting the formalism introduced in the previous section, we instead now introduce an underlying \textit{complete} binary relation which is asymmetric, but nontransitive.

If negative transitivity does not hold, preferences cannot be represented by utility functions, $q([\bx,\by]) \neq u(\bx)-u(\by)$. However, it is possible to represent the  nontransitive binary relations  through skew-symmetric functions (that is, binary relations that only satisfy asymmetry). These binary relations are commonly referred to as \textit{nontransitive preferences}.
Consider a skew-symmetric function $q : \mathcal{X} \times \mathcal{X} \rightarrow \mathbb{R}$, that is such that $q([\bx,\by])=-q([\by,\bx])$. We say that a nontransitive preference on $\mathcal{X}$ is represented by a two-argument function $q$ if, for all $\bx,\by \in \mathcal{X}$ , $q([\bx,\by]) > 0$ holds if and only if $\bx R \by$. 

 \cite{pahikkala2010learning} has shown that is possible to define a GP prior on skew-symmetric functions $q$, which does not enforce transitivity, by considering the following kernel
\begin{equation}
    \label{eq:ntprefkernel}
k_{ntpref}([\bx,\by],[\bx',\by'])=    k_{\boldsymbol{\theta}}(\bx,\bx')k_{\boldsymbol{\theta}}(\by,\by')-k_{\boldsymbol{\theta}}(\bx,\by')k_{\boldsymbol{\theta}}(\by,\bx'),
\end{equation}
which is commonly referred to as \textit{nontransitive preference kernel}.
Indeed, consider again the data matrix
$$
X=\begin{bmatrix}
   \bx & \by\\
   \by & \bx
  \end{bmatrix},
$$
it can be verified that 
$$
K_{ntpref}(X,X)=\begin{bmatrix}
1& 
-1\\
-1 & 1
\end{bmatrix}(k_{\boldsymbol{\theta}}(\bx,\bx)k_{\boldsymbol{\theta}}(\by,\by)-k_{\boldsymbol{\theta}}(\bx,\by)k_{\boldsymbol{\theta}}(\by,\bx)),
$$
which has rank one, showing that $q([\bx,\by])=-q([\by,\bx])$. However, $k_{ntpref}$ does not enforce transitivity (contrarily to $k_{pref}$).
By simply replacing  $q \sim GP(0,k_{pref}([\bx,\by],[\bx',\by']))$ with $q \sim GP(0,k_{ntpref}([\bx,\by],[\bx',\by']))$ in Model \ref{model:Probitclassification}, we can derive a probit model for nontransitive preferences as  GP classification.
A preference-learning framework which employs the  GP prior $q \sim GP(0,k_{ntpref}([\bx,\by],[\bx',\by']))$ was proposed by \cite{chau2022learning}. \citet{Hu_etal_2022_PrefShapley} used the kernel $k_{ntpref}$ to develop an explainability algorithm for preferences with a logistic likelihood.  

We do not know if a subject's preferences are transitive so, why do we not always resort to a model that relies on fewer assumptions (only asymmetry for instance)? The rationale behind this is that it is  more insightful (and more accurate) to model the factors contributing to a subject's apparent irrationality, such as noise, just-noticeable-differences, noise, and multiple utilities, rather than assuming the subject is simply irrational.

\section{Computational complexity}

As we have emphasised in the last section, object-preference learning is equivalent to binary classification, and, therefore, the complexity of inference for the models we presented in this section is similar to that of GP classification (with the exception of the multiple utility model, which we will discuss in Section \ref{sec:choice}.) Through sparse variational approximation, the complexity can be reduced to $\mathcal{O}(m)$, where $m$ is the number of preferences. Compared to GP classification, $m$ can be large. In fact, if we have $r$ objects, there are potentially  $m=r(r-1)/2$ unique pairwise preferences. For large $m$, the computational burden can be reduced, for instance, by applying stochastic variational inference \citep{hensman2015scalable} and by performing subsampling of the  $m$ preferences during training.

\chapter{Learning from label preferences}
\label{sec:labelpref}
We now turn to another kind of preference: preference over labels. In label preference, each covariate vector ${\bf x}$ is associated with a predefined set of labels $\mathcal{C}=\{c_1,\dots,c_d\}$ and a subject is asked to express preference relations over the label set. The labels are the objects of the preference. We introduce the notation $c_i \succ_{\bf x} c_j$ to indicate that, for the covariate ${\bf x}$, the label $c_i$ is preferred to $c_j$.  We use the term label here because, as we will show, we associate a utility function with each label. In other words, we will use one utility, function of the covariates $\bf x$, for each label $c_i$. The covariates ${\bf x}$  serve as contextual variables rather than as the objects over which preferences are expressed (the preferences are  over the labels).  We consider two scenarios where a subject can express their preferences over the labels: by providing pairwise comparisons between the labels or by giving a complete ordering (ranking) of the labels.
We will consider  classical models that have been proposed to capture these types of preferences and show how we can easily extend them with GPs. These extensions allow to introduce nonlinear functions of the covariates ${\bf x}$ as the utilities that determine the subject's preferences over the labels.

\begin{example}
We  present a simple example to illustrate label preferences. Imagine we ask Alice to express her preferences over three types of desserts: brownie, fruit cake and ice cream, throughout the year. The covariate $x$ in this case is a number between 1 and 365 (the day of the year). For some values of $x$, we have observations in the form of orderings over the three types of desserts, such as:
$$
\begin{array}{l}
x_1=10~~~~\textrm{brownie} \succ \textrm{fruitcake}  \succ \textrm{icecream} \\
x_2=90~~~~\textrm{fruitcake}  \succ \textrm{brownie} \succ \textrm{icecream} \\
x_3=100~~~~\textrm{fruitcake}  \succ \textrm{brownie}  \\
x_4=150~~~~\textrm{fruitcake}  \succ \textrm{icecream}  \\
x_5=180~~~~\textrm{icecream}  \succ \textrm{fruitcake} \succ \textrm{brownie}\\
\end{array}
$$
It can be noted that the above preference statements include partial orderings between the desserts (that is, complete orderings of only some items). Since we deal with strict preferences, we assume that there are no ties.
Note that, these preferences can also be expressed using pairwise comparisons, for instance $\text{brownie} \succ_{x_1} \text{fruitcake}$, $\text{fruitcake} \succ_{x_1} \text{icecream}$. In this problem, our goal is to predict Alice's preference/ordering for the different types of dessert on the day $x^*$.
\end{example}

It is important to note that preferences over labels play a crucial role in machine learning applications \citep{liu2009learning} such as search-engine and, in general, information retrieval . Specifically, the methods employed for this purpose are known as \textit{learning to rank}. There are  two main approaches \citep{liu2009learning}: listwise and pairwise. The listwise methods consider the ordering of a set of items (documents in information retrieval), while the pairwise methods are based solely on pairwise comparisons. In this section, we will present a listwise (Model \ref{model:PL}) and two pairwise (Model \ref{model:thurston} and \ref{model:pairwiselabel}) models, which are directly related to classical probabilistic preference modelling approaches. 
Models \ref{model:thurston}–\ref{model:pairwiselabel} generalise widely used models -- namely, the Thurstone, Plackett-Luce, and Bradley-Terry models -- originally developed in psychology,  social sciences, and sports rankings. The Thurstone and Bradley-Terry models handle pairwise comparisons between labels, with the probability of one item being preferred over another determined by the difference in utilities of the two labels. The Plackett-Luce model extends this framework to full or partial orderings by assigning probabilities to item permutations of the labels based on their relative utilities. Traditionally, these models are fitted via maximum likelihood estimation, and the utilities either depend only on the item labels or vary linearly with the covariates. In contrast, we discuss a nonlinear Bayesian extension of these models using GPs. It is also worth noting that, just as object preference can be cast as a binary classification problem, label preference can be framed as a multiclass classification problem -- a connection that will be made explicit in Section \ref{sec:PL}.
\section{Thurstone's model}
\label{sec:thurstone}
One natural way to represent label preferences is to define a utility function $u_i : \mathcal{X} \rightarrow \mathbb{R}$ for each label $c_i$, $i = 1,\dots,d$. Here, $u_i({\bf x})$ is the utility assigned to label $c_i$ by the object ${\bf x}$. To obtain a label preference for ${\bf x}$, the labels are ordered according to these utility functions, that is
$$
c_i \succ_{{\bf x}} c_j ~\text{ iff }~ u_i({\bf x})>u_j({\bf x}).
$$

 For each label, it is assumed that the observed utility is perturbed by Gaussian noise \citep{thurstone2017law}:
\begin{equation}
    \label{eq:noise_thur}
    \widetilde{u}_i({\bf x}) = u_i({\bf x}) + v_i({\bf x}),
\end{equation}
where $v_i({\bf x}) \sim N(0,\sigma^2)$ is the noise associated to the label $i$ in the instance $k$, and preferences are determined as follows $
c_i \succ_{{\bf x}} c_j$ if $\widetilde{u}_i({\bf x})>\widetilde{u}_j({\bf x})$.
We can then easily extend Model \ref{model:Gaussian} to deal with label preferences. To simplify the notation, we will denote the label $c_i$ as $i$. Therefore, we will write $
c_i \succ_{{\bf x}} c_j$ simply as $
i \succ_{{\bf x}} j$. The training dataset will consist of orderings of the labels. For instance, in the case of three labels ($d=3$) and for the objects ${\bf x}_1,\dots,{\bf x}_r$, the observations may be:
\begin{equation}
\label{eq:orderingnotex}
\begin{array}{c}
1 \succ_{{\bf x}_1} 2 \succ_{{\bf x}_1}3,~~1 \succ_{{\bf x}_2} 3 \succ_{{\bf x}_2}2,~~ 2 \succ_{{\bf x}_3} 1 \succ_{{\bf x}_3}3,~~ 1 \succ_{{\bf x}_4} 2 \succ_{{\bf x}_4}3.
\end{array}
\end{equation}
Each one of these $k=4$ observation-ordering is a permutation of the labels. We recall that the number of possible permutations $\Pi_d$ of $d$ objects is $d!$. We can denote each one of these permutations  with a vector $\boldsymbol{\pi}_{k}=[\pi_{k1},\dots,\pi_{kd}]^\top \in \Pi_d$. For example, for $d=3$, $\boldsymbol{\pi}_{k}$ may be equal to $[2,1,3]^\top$. In this case, we can equivalently write the observations in \eqref{eq:orderingnotex} more compactly as
\begin{equation}
\label{eq:orderingnotex1}
\begin{array}{c}
\{\pi_{k1} \succ_{{\bf x}^{(k)}} \pi_{k2}\succ_{{\bf x}^{(k)}}\pi_{k3}\mid k=1,\dots,4\},
\end{array}
\end{equation}
where $\boldsymbol{\pi}_{1}=[1,2,3]$,  $\boldsymbol{\pi}_{2}=[1,3,2]$,  $\boldsymbol{\pi}_{3}=[2,1,3]$, $\boldsymbol{\pi}_{4}=[1,2,3]$ and $[\boldsymbol{x}^{(1)},\boldsymbol{x}^{(2)},\boldsymbol{x}^{(3)},\boldsymbol{x}^{(4)}]=[\boldsymbol{x}_1,\dots,\boldsymbol{x}_4]$.\footnote{Note that, due to the Gaussian noise, we may have contradictory orderings in the dataset, such as $1 \succ_{{\bf x}_1} 2 \succ_{{\bf x}_1}3$ and $2 \succ_{{\bf x}_1} 1 \succ_{{\bf x}_1}3$.}
 \begin{SnugshadeB}
\begin{MethodB}[Thurstonian Model for Label Preferences]
\label{model:thurston}
Consider the training dataset:
\begin{equation}
\label{eq:datasetlabelThu}
\begin{aligned}
\mathcal{D}_m=\{&{\pi_{k1}} \succ_{{\bf x}^{(k)}} {\pi_{k2}} \succ_{{\bf x}^{(k)}} {\pi_{k3}} \succ_{{\bf x}^{(k)}} \dots  \succ_{{\bf x}^{(k)}} {\pi_{kd}}: \\
&~\boldsymbol{\pi}_{k} \in \Pi_d, ~\bx^{(k)} \in X,~ k=1,\dots,m\},
\end{aligned}
\end{equation}
 where $\boldsymbol{\pi}_{k}=[\pi_{k1},\dots,\pi_{kd}]^\top \in \Pi_d$ is a permutation of the indices $\{1,2,\dots,d\}$, called ordering , $X=\{\bx_1,\bx_2,\dots,\bx_r\}$ with $\bx_i \in \mathcal{X}$ and $k$ indexes the object ${\bf x}^{(k)} \in X$ considered in the $k$-th comparison between the labels. Consider the vectors
\begin{align*}
&{\bf u}({\bf x}_i)=\begin{bmatrix}
u_1({\bf x}_i) \\ 
u_2({\bf x}_i)\\ 
\vdots \\
u_d({\bf x}_i)\\ 
\end{bmatrix},~~~{\bf u}(X)=\begin{bmatrix}
{\bf u}({\bf x}_1) \\ 
{\bf u}({\bf x}_2) 
\\ 
\vdots \\
{\bf u}({\bf x}_d) \\ 
\end{bmatrix}, \\
&{\bf v}({\bf x}_i)=\begin{bmatrix}
v_1({\bf x}_i) \\ 
v_2({\bf x}_i)\\ 
\vdots \\
v_d({\bf x}_i)\\ 
\end{bmatrix},~~~{\bf v}(X)=\begin{bmatrix}
{\bf v}({\bf x}_1) \\ 
{\bf v}({\bf x}_2) 
\\ 
\vdots \\
{\bf v}({\bf x}_d) \\ 
\end{bmatrix}.
\end{align*}
Under the assumption that the preference statements are conditionally independent given ${{\bf u}}(X),{{\bf v}}(X)$, we obtain the likelihood:

\begin{align}
\label{eq:likelcdfmultinoise}
    &p(\mathcal{D}_m|{\bf u}(X),{\bf v}(X)) = \\ \nonumber    &\prod_{k=1}^m\prod_{i=1}^{d-1} I_{\scaleto{\left\{u_{{\pi_{ki}}}({\bf x}^{(k)})+v_{{\pi_{ki}}}({\bf x}^{(k)})-u_{{\pi_{k(i+1)}}}({\bf x}^{(k)})-v_{{\pi_{k(i+1)}}}({\bf x}^{(k)})>0\right\}}{18pt}}({\bf u}(X),{\bf v}(X)),
\end{align}
where $u_{{\pi_{ki}}}$ (resp.\ $u_{{\pi_{k(i+1)}}}$) denotes the utility function associated to the label  ${{\pi_{ki}}}$ (resp.\ ${{\pi_{k(i+1)}}}$), similar notation is  used for $v_{{\pi_{ki}}},v_{{\pi_{k(i+1)}}}$. We can write the constraints expressed by the $m$ indicator functions in matrix form as 
$$
\begin{bmatrix}
\widetilde{W}& \widetilde{W}
\end{bmatrix}
\begin{bmatrix}
{\bf u}(X)\\
{\bf v}(X)
\end{bmatrix}>0,
$$
where $\widetilde{W}$ is a $m \times (rd)$ data-dependent matrix. As in GP multiclass classification \citep{williams1998bayesian}, we assign independent GP priors to each utility function resulting into:
\begin{equation}
\label{eq:multiprior}
u_i \sim \mathrm{GP}(u;0,k_i), ~~~~~
{\bf u}=\begin{bmatrix}
u_1 \\ 
u_2\\ 
\vdots \\
u_d\\ 
\end{bmatrix} \sim GP\left({\bf 0},\mqty[\dmat[0]{k_1,k_2,\ddots,k_d}]\right).    
\end{equation}
The posterior can be derived using the algorithm described in Section \ref{sec:gaussiannoise}. We assume we use the same type of kernel $k_i=k_j$ for each utility function, although the hyperparameters can be different and estimated using the same approach used for object preference.
\end{MethodB}
\end{SnugshadeB}
The above model can also be applied in cases where Alice only provides a partial ordering, i.e., an ordering of a subset of the labels. Indeed, the likelihood \eqref{eq:likelcdfmultinoise} is still valid, but the inner product will include less
terms.

\begin{example}
\label{ex:dessThurst}
Imagine Alice's true utilities for 
 brownie, fruit cake and ice cream, throughout the year, are depicted in Figure \ref{fig:dessert1}. 
\begin{figure}[h]
\centering
\includegraphics[width=7cm]{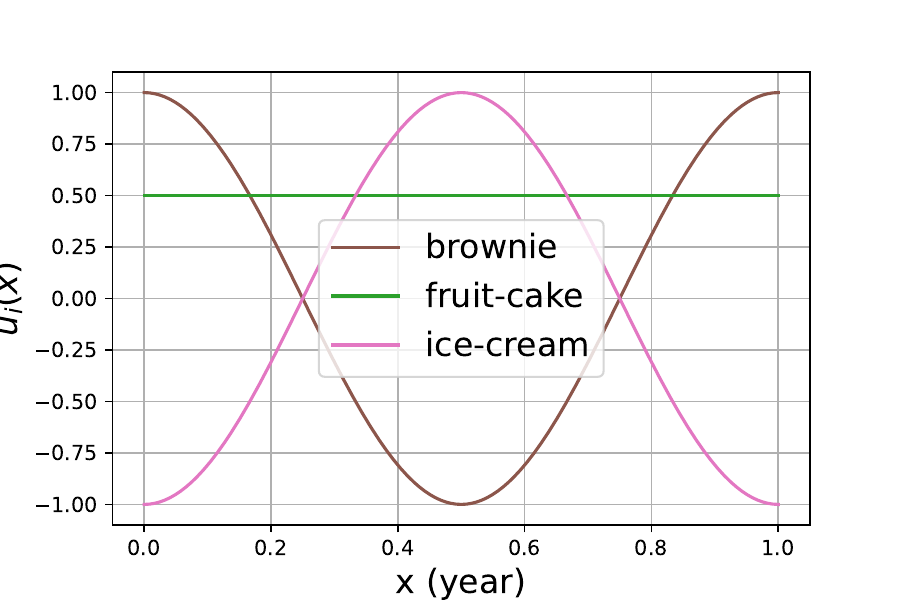}
\caption{Alice's true utilities for 
 brownie, fruit cake and ice cream.}
\label{fig:dessert1}
\end{figure}
In order to infer her utility function,  we ask Alice her preferences  regarding the three types of desserts on 50 different days, which are randomly selected throughout the course of the year.
For instance, in day
$x=75/365\approx 0.2$, her preferences are 
$\textrm{fruit-cake}  \succ \textrm{brownie} \succ \textrm{ice-cream}$ (see Figure \ref{fig:dessert1}). 
 Therefore, the dataset, $\mathcal{D}_{50}$, includes  a collection of fifty triplet-preferences for the three desserts, each one corresponding to each of the fifty days.
We aim to learn Alice's utility functions for the three desserts from  $\mathcal{D}_{50}$ and predict her preferences for any other day of the year.
Figure \ref{fig:dessert1est}-top displays the posterior-mean of the  utilities obtained from Model \ref{model:thurston}. It can be noticed that the ordering between the three mean functions is in agreement with that of the original utilities (in  Figure \ref{fig:dessert1}) for each value of $x$.
Figure \ref{fig:dessert1est}-bottom shows the posterior mean and  95\% credible intervals for each of the three utilities as computed by Model \ref{model:thurston}.
\begin{figure}[h]
\centering
\includegraphics[width=7cm]{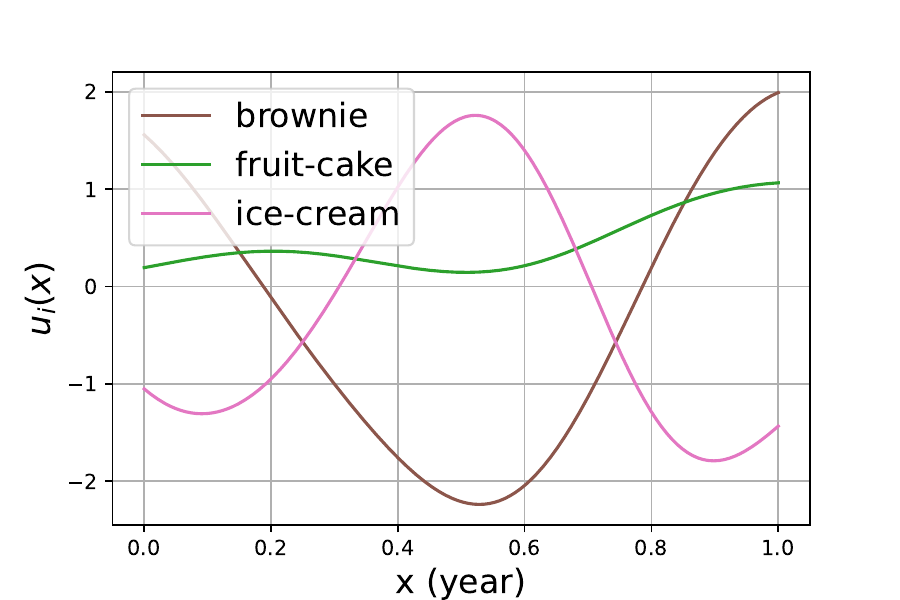}\\
\includegraphics[width=5cm]{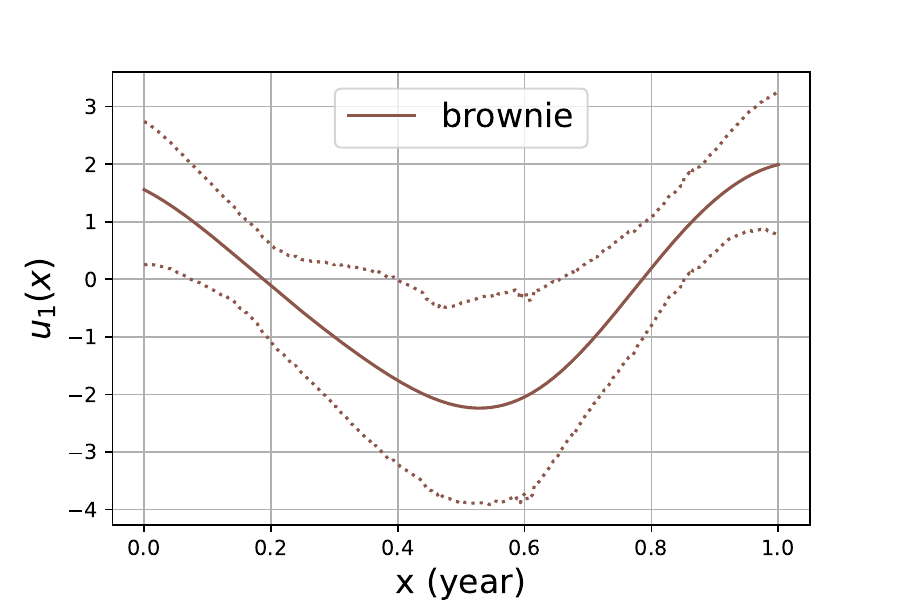}
\includegraphics[width=5cm]{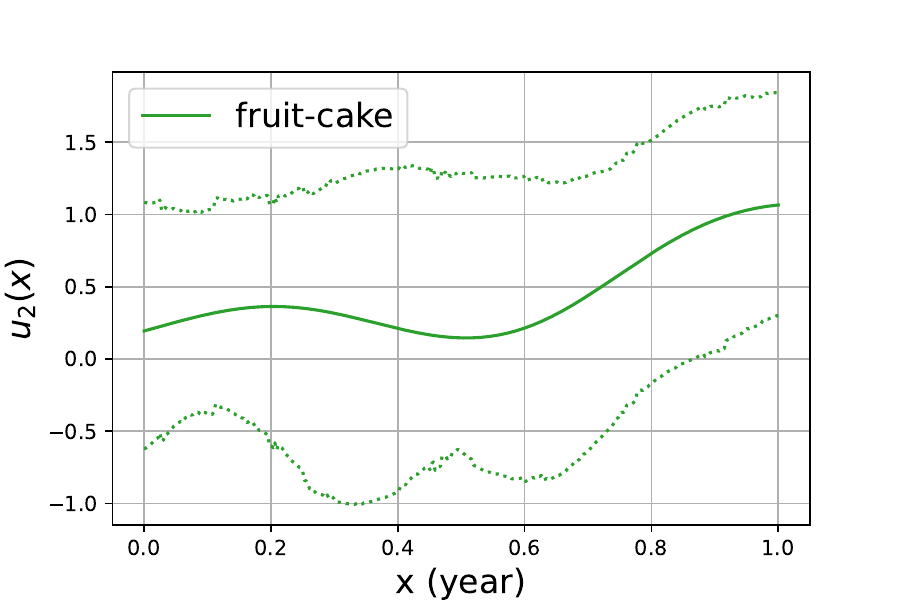}
\includegraphics[width=5cm]{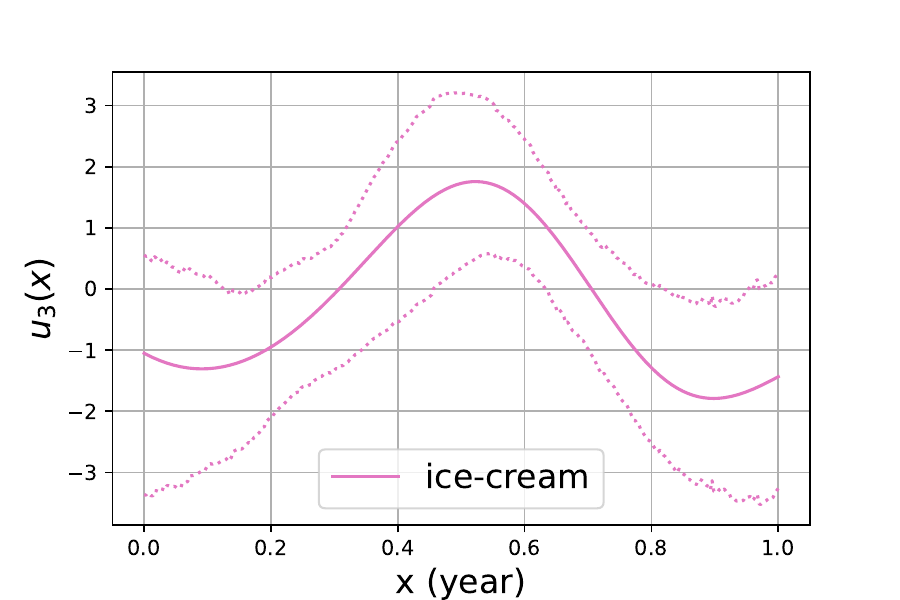}
\caption{Top: mean of the predicted posterior utilities for  the three desserts. Bottom: Predictive mean and 95\% credible interval for each one of the three desserts.}
\label{fig:dessert1est}
\end{figure}
If we wish to predict Alice's preferences for the three desserts in day $x^*=90/365\approx 0.25$, then we can do this by computing the posterior distributions of the differences $u_{fruitcake}-u_{brownie}$ and $u_{brownie}-u_{icecream}$ by sampling from the posterior. The result is shown in Figure \ref{fig:3dessersdist}. For instance, the support of the distribution for  $u_{fruitcake}-u_{brownie}>0$ does not include zero, this allows us to conclude that Alice prefers fruit-cake over brownie in day $90$ with probability $\approx 1$. We are instead undecided about the comparison brownie versus ice-cream.

\begin{figure}
    \centering
\includegraphics[width=7cm]{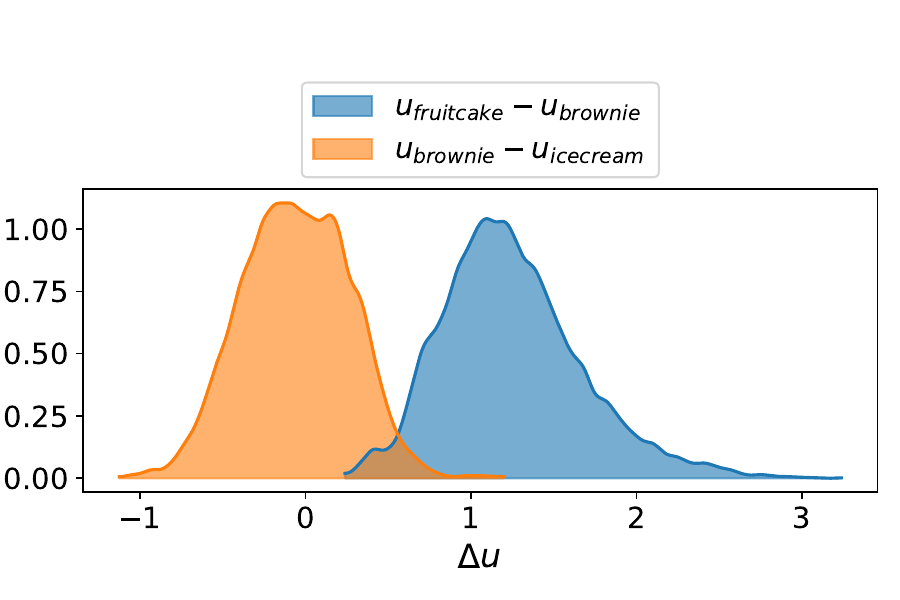}
    \caption{Posterior distributions of the differences $u_{fruitcake}-u_{brownie}$ and $u_{brownie}-u_{icecream}$.}
    \label{fig:3dessersdist}
\end{figure}
\end{example}

\section{Plackett-Luce model}
\label{sec:PL}
Assume that, for each instance ${\bf x}^{(k)}$, we interpret a ranking of  the $d$ labels as a sequence of $d-1$ independent choices, that is first Alice chooses the
first item, then chooses the second among the remaining alternatives, and so on. The Plackett-Luce (PL) \citep{luce2012individual,plackett1975analysis} model provides a distribution over orderings. 
Given a dataset of orderings
$$
{\pi_{k1}} \succ_{{\bf x}^{(k)}} {\pi_{k2}} \succ_{{\bf x}^{(k)}} {\pi_{k3}} \succ_{{\bf x}^{(k)}} \dots  \succ_{{\bf x}^{(k)}} {\pi_{kd}},
$$
for $k=1,\dots,m$, where $\boldsymbol{\pi}_{k}=[\pi_{k1},\dots,\pi_{kd}]^\top \in \Pi_d$  is a permutation of the indices $\{1,2,\dots,d\}$, the PL distribution is  described in terms of the associated ordering $\boldsymbol{\pi}_{k}$ as
\begin{equation}
    \label{eq:lucelike}
p(\boldsymbol{\pi}_{k}|{\bf a})=\prod_{i=1}^{d-1} \frac{a_{\pi_{ki}}}{\sum_{j=i}^d a_{\pi_{kj}}},
\end{equation}
for $k=1,\dots,m$, where ${\bf a} \in \mathbb{R}^d$ with $a_i>0$ for $i=1,\dots,d$ are so-called score variables.
This model satisfies the \textit{Luce's axiom of Choice} \citep{luce2012individual}, which states that the odds of choosing an item over another do not depend on the set of items from which the choice is made. Indeed, for any choice $\boldsymbol{\pi}_{k}$ and for any two alternatives $s$ and $t$ in the choice-set ($s,t\geq i$), the ratio of the probabilities is
$$
\frac{\frac{a_{\pi_{ks}}}{\sum_{j=i}^d a_{\pi_{kj}}}}{\frac{a_{\pi_{kt}}}{\sum_{j=i}^d a_{\pi_{kj}}}}=\frac{a_{\pi_{ks}}}{a_{\pi_{kt}}},
$$
which does not depend on any alternatives other than $s$ and $t$. A standard approach is to define the score  $a_i$ as function of the utility $u_i$, in particular it is common to consider the relation  $a_i=e^{u_i}$. In this case, the likelihood \eqref{eq:lucelike} becomes equal to:
\begin{equation}
    \label{eq:lucelike1}
p(\boldsymbol{\pi}_{k}|{\bf u}({\bf x}^{(k)}))=\prod_{i=1}^{d-1} \frac{e^{u_i({\bf x}^{(k)})}}{\sum_{j=i}^d e^{u_j({\bf x}^{(k)})}},
\end{equation}
which is known as \textit{exploded logit} likelihood model. It can be noticed that each term in the product in \eqref{eq:lucelike1} corresponds to the softmax function used in multi-class classification.

In Model \ref{model:PL}, we will use GPs to learn the utility functions underlying the PL-model.~\\
~\\

 \begin{SnugshadeB}
\begin{MethodB}[Plackett-Luce  Model for Label Ordering Data]
\label{model:PL}
Consider a training dataset including the ordering of $d$-labels for each one of the $m$ covariates:
$$
\begin{aligned}
\mathcal{D}_m=\{&{\pi_{k1}} \succ_{{\bf x}^{(k)}} {\pi_{k2}} \succ_{{\bf x}^{(k)}} {\pi_{k3}} \succ_{{\bf x}^{(k)}} \dots  \succ_{{\bf x}^{(k)}} {\pi_{kd}}: \\
&\boldsymbol{\pi}_{k} \in \Pi_d, \bx^{(k)} \in X,~ k=1,\dots,m\}.
\end{aligned}
$$

Under the assumption that the $m$ orderings statements are conditionally independent given
 $\mathbf{u}(X)$, we obtain the likelihood:
\begin{equation}
\label{eq:likePLprod}
    p(\mathcal{D}_m|{\bf u}(X))=\prod_{k=1}^m \prod_{i=1}^{d-1} \frac{e^{u_{\pi_{ki}}({\bf x}^{(k)})}}{\sum_{j=i}^d e^{u_{\pi_{kj}}({\bf x}^{(k)})}}.
\end{equation}
 We assign independent GP priors to each utility function resulting into:
\begin{equation}
\label{eq:multiprior2}
u_i \sim \mathrm{GP}(u;0,k_i), ~~~~~
{\bf u}=\begin{bmatrix}
u_1 \\ 
u_2\\ 
\vdots \\
u_d\\ 
\end{bmatrix} \sim GP\left({\bf 0},\mqty[\dmat[0]{k_1,k_2,\ddots,k_d}]\right).    
\end{equation}
For simplicity, we assume the same kernel $k_i=k_j$ for each utility function, although the hyperparameters can be different.  Given the above likelihood and prior, inference is carried out as described in Algorithm \ref{alg:1}, where steps A1--A3 are performed using variational inference as approximation method for the posterior.
\end{MethodB}
\end{SnugshadeB}
This model was proposed by \cite{nguyen2021top}, who also account for ties using a threshold $\delta$ that represents the limit of discernibility. Therefore, their model integrates Model \ref{model:PL} with Model \ref{Model:JND}.

Note that,  Model \ref{model:PL}  can also be applied in  cases where Alice only provides an ordering of a subset of the labels. Indeed, the likelihood \eqref{eq:likePLprod} is still valid, but the inner product will include less terms. Furthermore, we can see that if Alice only states her most preferred label, the likelihood \eqref{eq:likePLprod} and Model \ref{model:PL} simplify to the standard multiclass classification GP model. We can exploit the structure of the likelihood to reduce the number of variational parameters in the covariance matrix of the Gaussian variational distribution, similarly to what is done in \citet{opper2009variational}. However, in this case, the variational covariance matrix cannot be diagonal, we need some additional parameters to model the correlation between the various utilities functions.

There is a connection between  Thurstone's model and Plackett-Luce's model. A key result by \citet{yellott1977relationship} states that if the
 variables $\widetilde{u}_i$ in \eqref{eq:noise_thur} are independent, and their distributions are identical except for their means, then 
Thurstone's model gives rise to a PL model if and only
if the  $\widetilde{u}_i$ are distributed according to a \textit{Gumbel distribution}.

\begin{proposition}[\cite{yellott1977relationship}]
\label{prop:gumbel}
Assume that the noise variables, $v_i$, in \eqref{eq:noise_thur} are independent and distributed according to a \textit{Gumbel distribution}, whose cumulative distribution is $p(v_i)=e^{-e^{-v_i}}$, then
\begin{equation}
\label{eq:yellott1}
p(\widetilde{u}_i = \max(\widetilde{u}_1,\dots,\widetilde{u}_d))=\frac{a_i}{\sum_{j=1}^d a_j},  ~\text{ with }~  a_j=e^{u_j}.
\end{equation}
\end{proposition}
The proof and details are reported in Appendix \ref{app:gumbel}. This relation is known in machine learning as the ``Gumbel-max trick''.

\begin{figure}
\centering
\includegraphics[width=7cm]{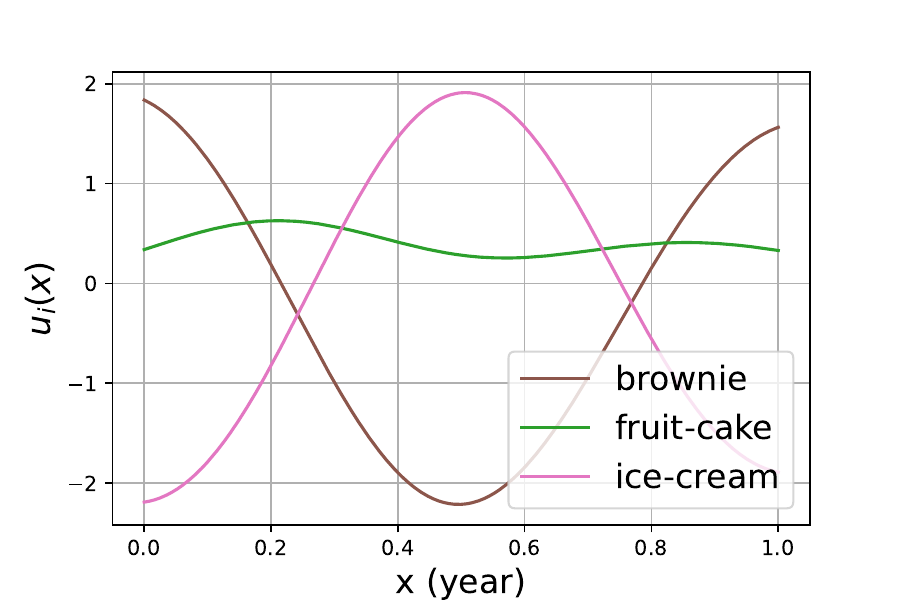}\\
\includegraphics[width=5cm]{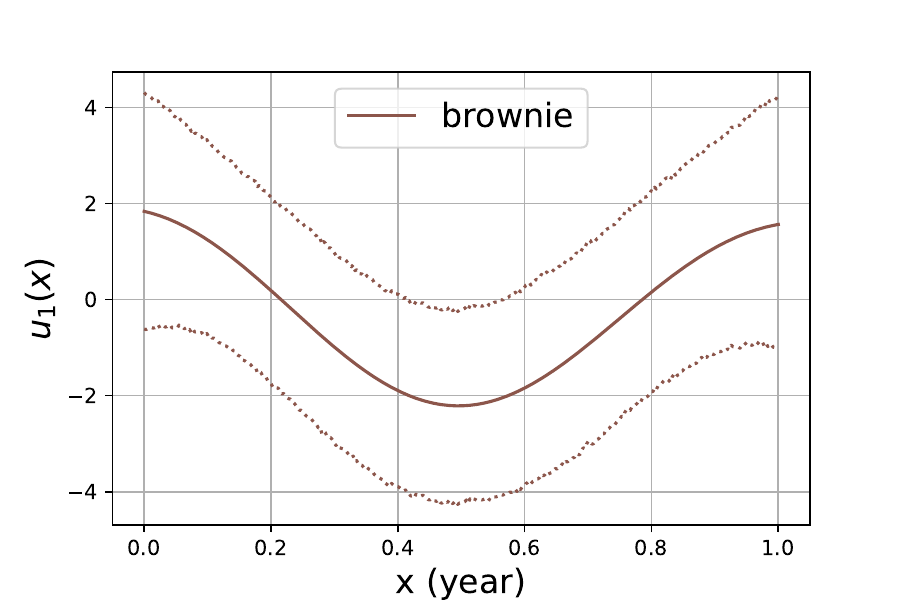}
\includegraphics[width=5cm]{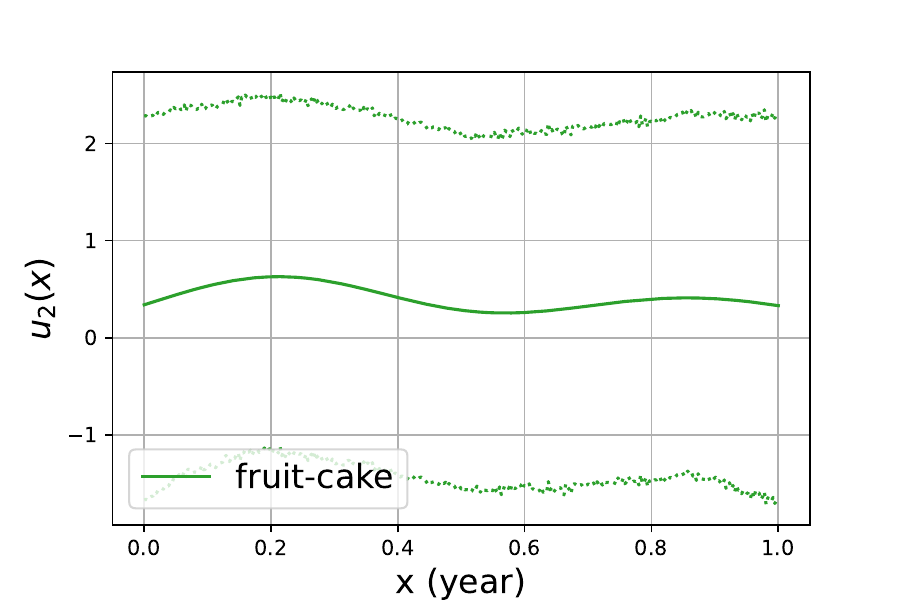}
\includegraphics[width=5cm]{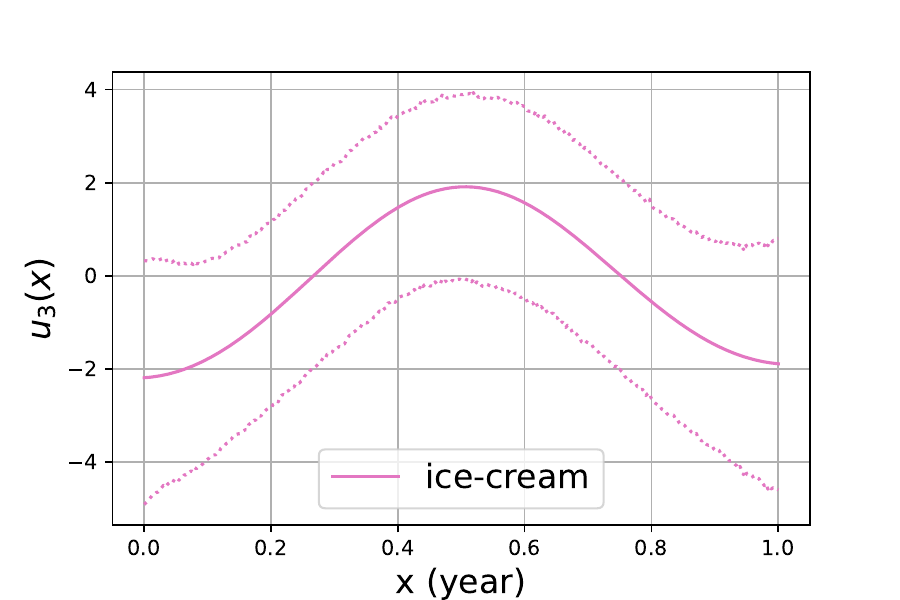}
\caption{Top: mean of the predicted posterior utilities for  the three desserts. Bottom: Predictive mean and 95\% credible interval for each one of the three desserts.}
\label{fig:dessert1PL}
\end{figure}

\begin{example}
\label{ex:dessPL}
We consider the same scenario as in Example \ref{ex:dessThurst}, where Alice orders her favourite dessert among brownie, fruit cake and ice cream on 50 different days, according to the true utilities shown in Figure \ref{fig:dessert1}.  This results in the same dataset as in Example \ref{ex:dessThurst}, but we interpret the triplets as orderings. Our goal is to infer Alice's utility functions for each dessert from the dataset $\mathcal{D}_{50}$ and predict how she would order them on any other day of the year.
Figure \ref{fig:dessert1PL}-top shows the
posterior-mean of the utilities obtained from
Model \ref{model:PL}. Also in this case, it can be noticed that the the ranking
between the three mean functions is in agreement with that of the original utilities.
Figure \ref{fig:dessert1PL}-bottom shows the mean and 95\% credible intervals for each of the three utilities separately.
Since the number of possible orderings is equal to the factorial of the total number of choices (three desserts), to calculate the predicted probability of each possible ordering on a given day $x$, we can draw samples of the utilities $\mathbf{u}(x)$ from the posterior distribution and then use them to approximate ranking distribution. 
Figure \ref{fig:dessertRank} shows the ranking distribution at day $x=0.25$ for this simple example.
\begin{figure}
\centering
\includegraphics[width=7cm]{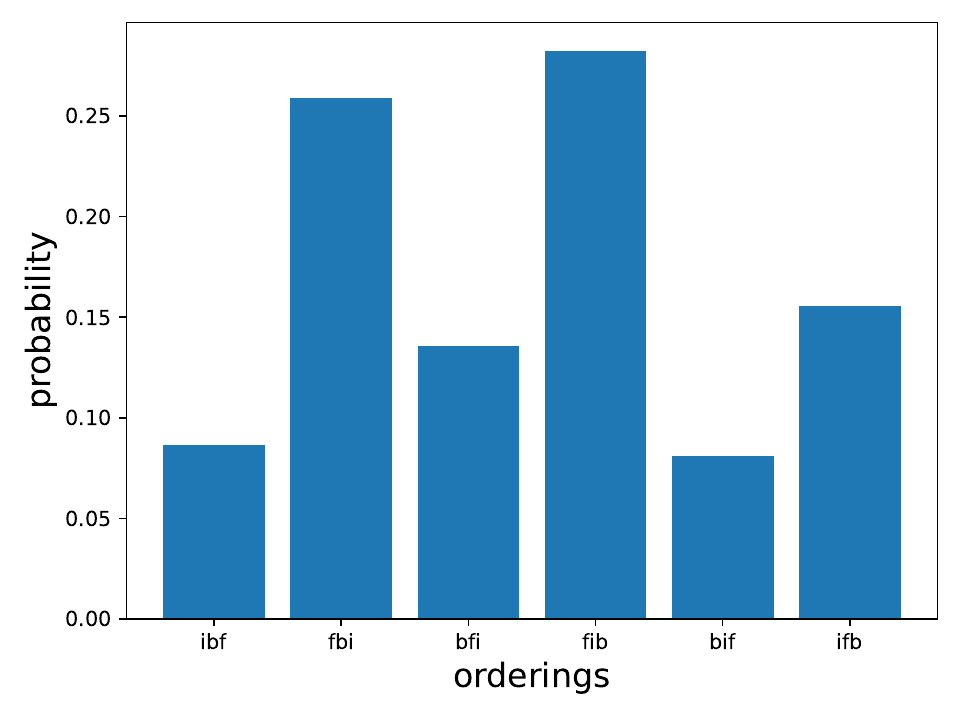}
\caption{Predicted ranking distribution at day $x=0.25$. The string `ibf' stays for  \textrm{icecream} $\succ$ \textrm{brownie}  $\succ$ \textrm{fruitcake}. The most probable ordering is `fib' \textrm{fruitcake}  $\succ$ \textrm{icecream} $\succ$ \textrm{brownie}.}
\label{fig:dessertRank}
\end{figure}
\end{example}

\section{Paired comparison data}
\label{sec:labelPW}
Instead of asking Alice to order the $d$ labels we could ask her to
choose which of each pair of labels is preferred. Thus instead of giving
the ordering ($\textrm{brownie} \succ_{\bf x} \textrm{fruitcake}  \succ_{\bf x} \textrm{icecream}$), she would say that, for object ${\bf x}$, ``brownie is preferred to
fruitcake'', ``fruitcake is preferred to icecream'' and ``brownie is preferred to icecream''.
For $d$ labels there are  $m= {{d}\choose{2}}$ of such pairwise comparisons.
In the following model, we assume that every time label $i$ is compared with other labels, a new utility observation $\widetilde{u}_i({\bf x})$ is produced, resulting in a new realisation of the noise in \eqref{eq:noise_thur} (even for the same object ${\bf x}$). In this case, under Gaussian noise in \eqref{eq:noise_thur}, the resulting preference model is just an extension of Model \ref{Model:Probit} presented in Section \ref{sec:discern}. 
In  the model below, we will use the notation $i_s \succ_{{\bf x}^{(s)}} j_s$ to mean that, in the $s$-th pairwise comparison, the label $i_s$ was preferred to the label $j_s$ for the object ${\bf x}^{(s)}$, where $i_s,j_s \in \mathcal{C}$ and ${\bf x}^{(s)}  \in \{\bx_1,\bx_2,\dots,\bx_r\}$.

 \begin{SnugshadeB}
\begin{MethodB}[Paired comparison model with GP prior]
\label{model:pairwiselabel}

Consider the training dataset 
\begin{equation}
    \label{eq:datasetlabel}
    \mathcal{D}_m = \left\{ i_s \succ_{{\bf x}^{(s)}} j_s:~~ s = 1,\dots,m\right\},
\end{equation}
where, for each $s$, $i_s \neq j_s$, $i_s,j_s \in \mathcal{C}$ and ${\bf x}^{(s)}  \in \{\bx_1,\bx_2,\dots,\bx_r\}$ with $\bx_i \in \mathcal{X}$. Consider the vector $X=[\bx_1,\bx_2,\dots,\bx_r]^\top$ and 
$$
{\bf u}({\bf x}_i)=\begin{bmatrix}
u_1({\bf x}_i) \\ 
u_2({\bf x}_i)\\ 
\vdots \\
u_d({\bf x}_i)\\ 
\end{bmatrix},~~~{\bf u}(X)=\begin{bmatrix}
{\bf u}({\bf x}_1) \\ 
{\bf u}({\bf x}_2) 
\\ 
\vdots \\
{\bf u}({\bf x}_d) \\ 
\end{bmatrix}.
$$
Under the assumption that the preference statements are conditionally independent given ${\bf u}(X)$, we obtain the likelihood:
\begin{equation}
    \label{eq:likelcdfmulti}
    p(\mathcal{D}_m|{\bf u}(X))=\prod_{s=1}^m \Phi\left(u_{i_s}({\bf x}^{(s)})-u_{j_s}({\bf x}^{(s)})\right)=\Phi\left(\widetilde{W}{\bf u}(X)\right),
\end{equation}
where $u_{i_s}$ (resp.\ $u_{j_s}$) denotes the utility function associated to the label  $i_s$ (resp.\ $j_s$) and the last term in \eqref{eq:likelcdfmulti} gives the expression of the likelihood using the matrix form for a suitable $m \times (rd)$ matrix  $\widetilde{W}$. We assign independent GP priors to each utility function resulting into:
\begin{equation}
\label{eq:multiprior300}
u_i \sim \mathrm{GP}(u;0,k_i), ~~~~~
{\bf u}=\begin{bmatrix}
u_1 \\ 
u_2\\ 
\vdots \\
u_d\\ 
\end{bmatrix} \sim GP\left({\bf 0},\mqty[\dmat[0]{k_1,k_2,\ddots,k_d}]\right).    
\end{equation}
The posterior is similar to that in \eqref{eq:predPostSample} but with $\widetilde{W}$ instead of $W$, and the diagonal kernel in \eqref{eq:multiprior300}.
\end{MethodB}
\end{SnugshadeB}
This model was first derived by \cite{ChuGhahramani_preference2005}. Note that, under Gumbel noise in \eqref{eq:noise_thur}, the likelihood  \eqref{eq:likelcdfmulti} is equal to
\begin{equation}
    \label{eq:likelcdfmulti1}
    p(\mathcal{D}_m|{\bf u}(X))=\prod_{s=1}^m
    \frac{\exp({u_{i_s}({\bf x}^{(s)})})}{\exp({u_{i_s}({\bf x}^{(s)})})+\exp({u_{j_s}({\bf x}^{(s)})})},
\end{equation}
which corresponds to a combination of Bradley and Terry model \citep{bradley1952rank} and Babington Smith model \citep{smith1950discussion} proposed by \citet{mallows1957non}.

In order to understand the difference between the above pairwise comparison model and the previous two models consider the preferences: $\textrm{brownie}\succ_{\bf x} \textrm{fruitcake}$ and
$ \textrm{fruitcake}\succ_{\bf x} \textrm{icecream}$. Under the independence assumption underlying the likelihood \eqref{eq:likelcdfmulti} and Gaussian noise, these preferences imply the utility relations:
\begin{enumerate}
\itemsep0.1em
\item first preference:  $
u_{\textrm{brownie}}+v_{\textrm{brownie}}>u_{\textrm{fruitcake}}+v_{\textrm{fruitcake}}$;
\item second preference: $
u_{\textrm{fruitcake}}+v'_{\textrm{fruitcake}}>u_{\textrm{icecream}}+v_{\textrm{icecream}}$;
\end{enumerate}
where $v_{\textrm{fruitcake}},v'_{\textrm{fruitcake}}$ are two different realisations of the noise. Therefore, we cannot invoke the transitive property to conclude that $\textrm{brownie}\succ_{\bf x} \textrm{icecream}$ (and, in general, Model \ref{model:pairwiselabel} will not predict that). Therefore, Model \ref{model:pairwiselabel} should only be applied when the pairwise comparisons are independent, otherwise the posterior will be incorrect, as similarly shown in Example \ref{ex:7}.

\section{Correlated GP prior}
In the models presented in this section, we used an independent GP prior for each utility function $u_i$ for $i=1,2,\dots,d$, that is, a diagonal prior covariance function. However, since these utility functions are compared in the likelihood, the posterior covariance is no longer diagonal: the model learns dependence between the utility functions.

Alternatively, a prior correlation  may also be assumed, therefore considering a non-diagonal prior covariance function. For instance, we could use a \textit{coregionalised kernel}: this is a sum of
products between a kernel function for the input ${\bf x}$, and a kernel function that encodes the interactions
among the utilities, see for instance \citep{micchelli2004kernels,alvarez2012kernels}.
This kernel can be used to model known prior correlation or trade-off between two or more utility functions.

\section{Computational complexity in label preferences}
Compared to object preference, the computational complexity of label preference is generally higher, as it depends on the number of labels $d$. Specifically, when comparing the likelihoods in Model \ref{model:thurston} and Model \ref{model:PL}, the likelihood can contain up to $md$ terms. In Model \ref{model:pairwiselabel}, it contains  $m$ terms, but $m$ may be very large, since, for instance, a single ordering of $d$ elements is transformed into $(d - 1)!$ pairwise comparisons
 to apply Model \ref{model:pairwiselabel}. Considering the prior, we assume a GP for each utility function. To ensure scalability, one can for instance apply stochastic variational inference \citep{hensman2015scalable} and perform subsampling along both the $m$ and $d$ dimensions. We will use this approach in Sections ~\ref{sec:tennis} and \ref{sec:movie} to apply Model \ref{model:pairwiselabel} for predicting tennis match outcomes (where $m$ is the number of matches and $d$ is the number of tennis players) and for implementing a movie recommendation system (where $m$ is the number of users and $d$ is the number of movies).

\chapter{Learning from choice data}
\label{sec:choice}
In the previous sections, we have assumed that Alice expresses her judgements either as preferences or as an ordering. 
 However, we often only observe her choices. For instance, given a set of products on an internet commerce site, she may decide to click on/buy some of them. Mathematically, Alice's choices can be formalised through the concept of choice functions, as discussed in Section \ref{sec:choicefunintro}.
\begin{example}
\label{ex:cupcake0}
Let us explore an example where we seek to model Alice's preferences among various types of cupcake. We assume that the differences between these cupcake types is determined by their recipe.  In the context of choice modelling,  our primary goal is to observe Alice's choices among the various cupcake options, without delving into the specifics of how her preferences are determined. To be more specific, we present Alice with a set of different cupcakes, each one prepared with a different recipe that defines its characteristics, and ask her to pick the ones that appeal to her the most. For example, given the choice set
		\mbox{\normalsize {\normalsize $A=$}$\bigg\{$} 
		\includegraphics[width=0.8cm,trim={10cm 10cm 10cm 0},clip]{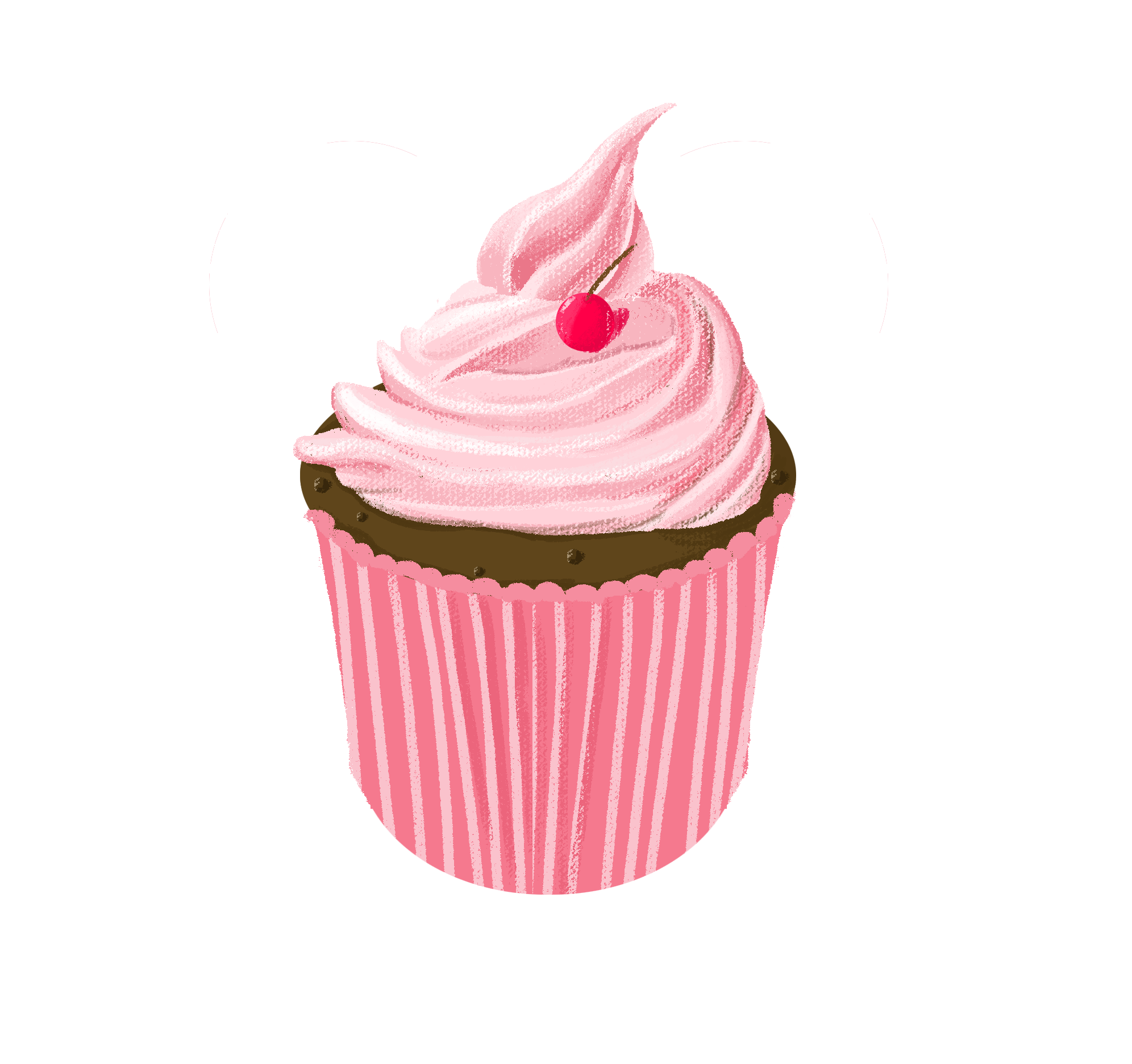},  
		\includegraphics[width=0.8cm,trim={1cm 1.5cm 0 0},clip]{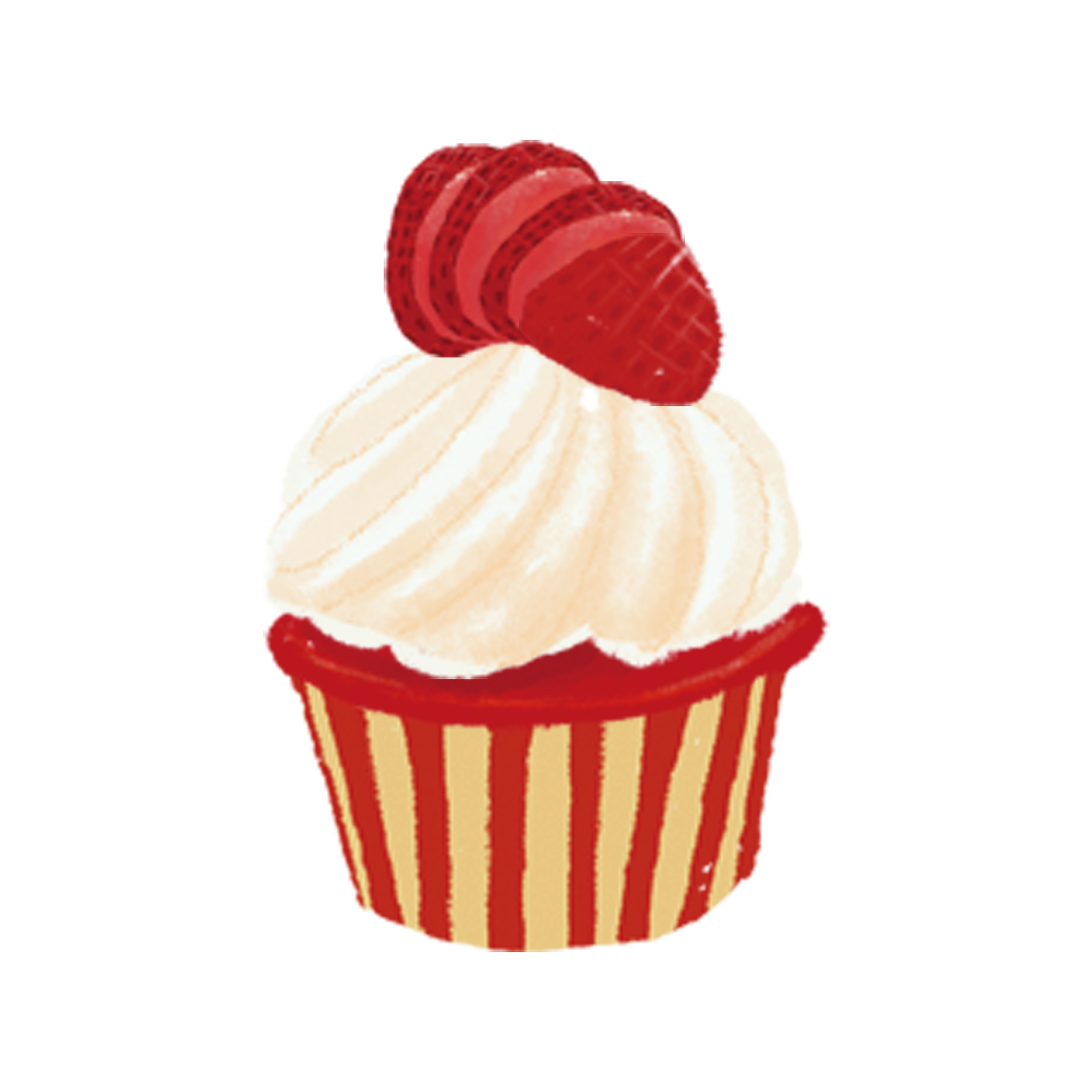},  
		\includegraphics[width=0.8cm,trim={10cm 4.5cm 0 0},clip]{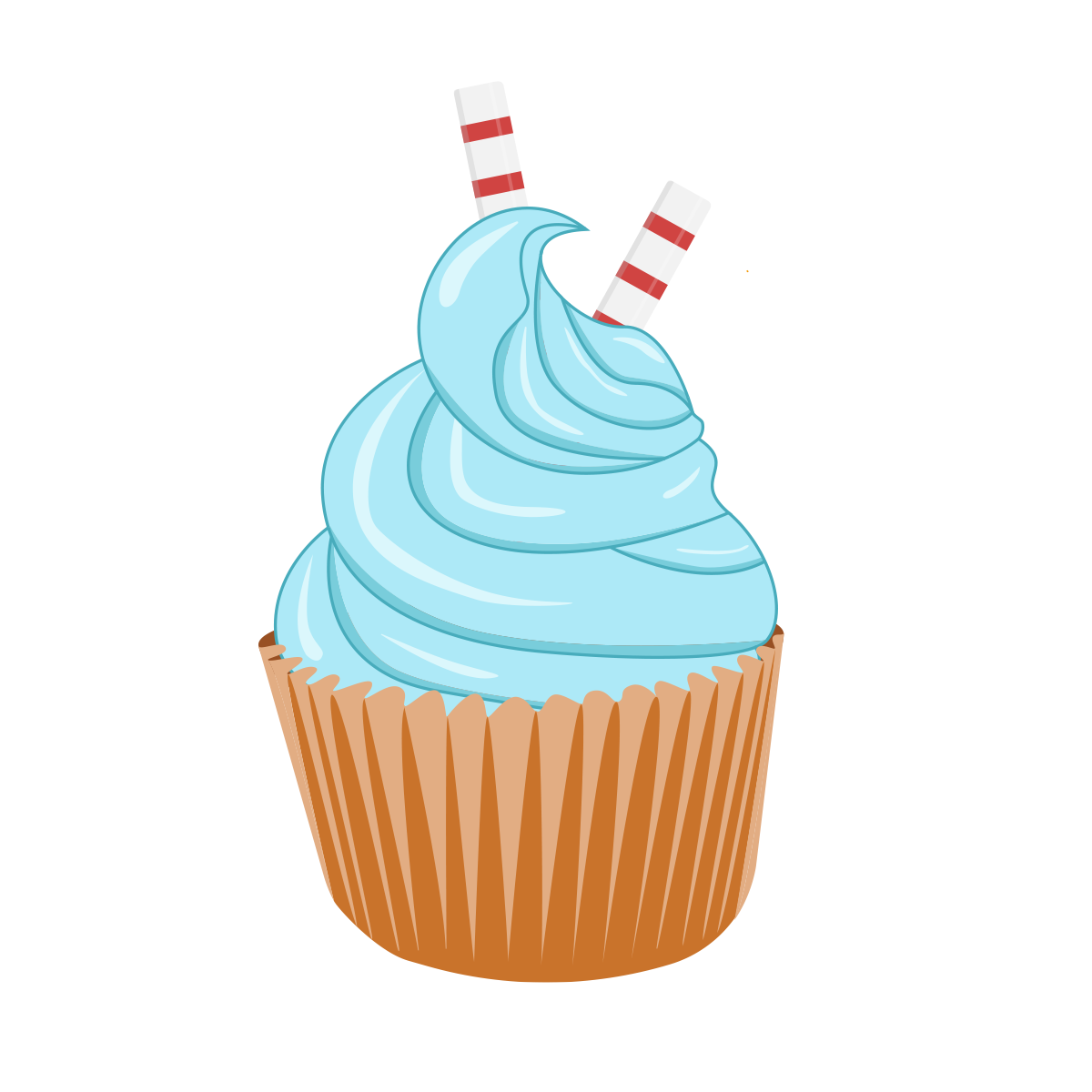},\includegraphics[width=0.6cm,trim={10cm 6cm 10cm 0},clip]{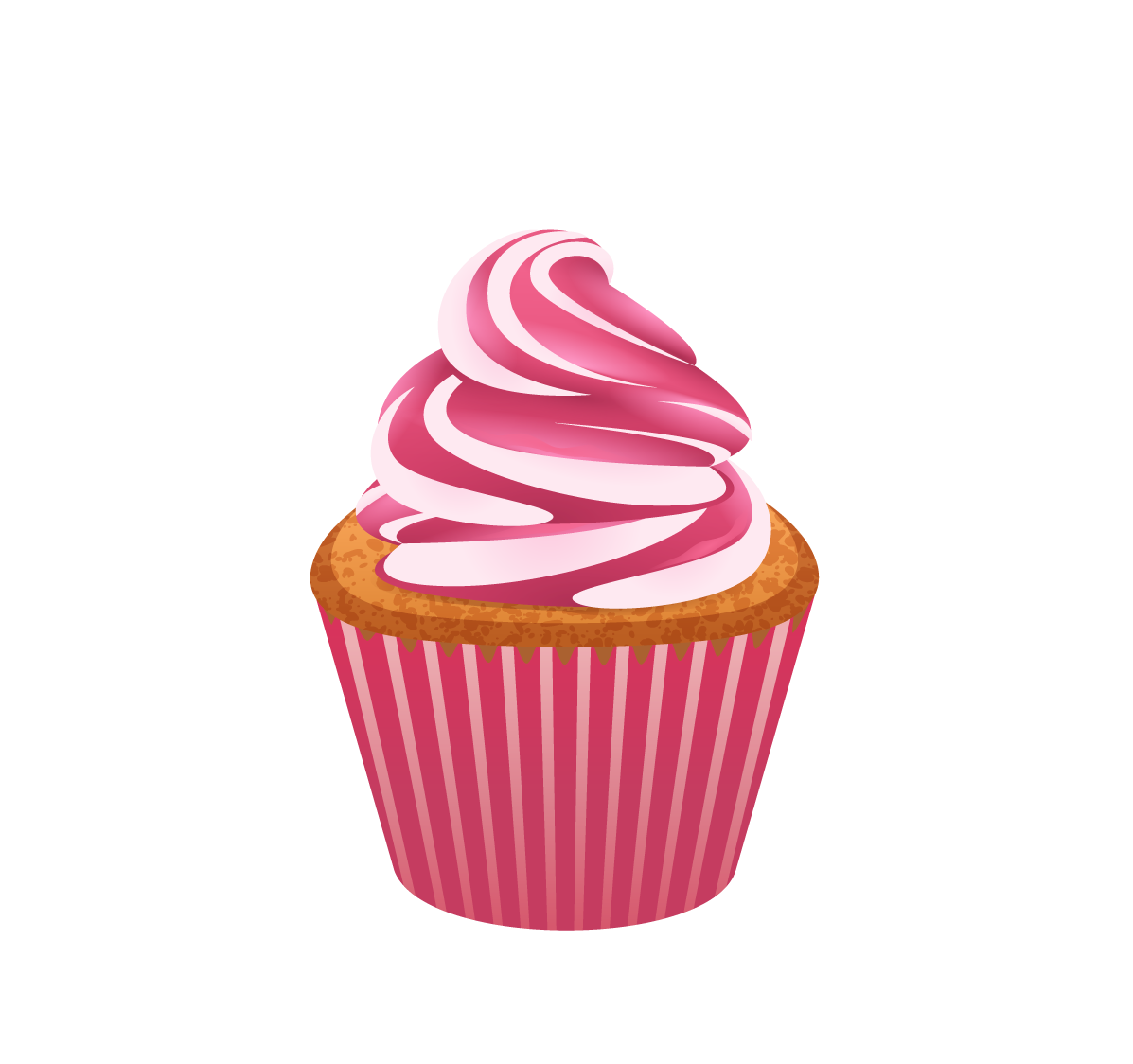}		\mbox{\normalsize $\bigg\}$}, she chooses

		\mbox{\normalsize {\normalsize$C(A)=$}$\bigg\{$}  \includegraphics[width=0.8cm,trim={1cm 1.5cm 0 0},clip]{figs/cupcake2.png},  
\includegraphics[width=0.8cm,trim={10cm 4.5cm 0 0},clip]{figs/cupcake5.png}
\mbox{\normalsize$\bigg\}$} and she rejects 	\mbox{\normalsize {\normalsize$R(A)=$}$\bigg\{$} \includegraphics[width=0.6cm,trim={10cm 10cm 10cm 10cm},clip]{figs/cupcake1.png},\includegraphics[width=0.6cm,trim={10cm 6cm 10cm 2cm},clip]{figs/cupcake4.png}\mbox{\normalsize$\bigg\}$}.
What is the meaning of these choices? For simplicity, let us consider a single ingredient for the cupcakes, denoted as $x \in [0,1]$. For instance, $x$ may be the normalised amount of butter. We present Alice with sets of cupcakes that differ only in the amount of butter used. She is asked to taste them and select her favourites. In doing so, Alice is effectively choosing the recipe, that is, the butter quantity she prefers. Therefore, we can represent each cupcake with its butter quantity $\Big\{$\includegraphics[width=0.6cm,trim={10cm 11cm 10cm 10cm},clip]{figs/cupcake1.png}\mbox{\normalsize$\Big\}$}=$\{0.1\}$. Assume she chooses as the following
$$
\begin{aligned}
A_1&=\{0,0.25,0.5\}, &C(A_1)&=\{0.25,0.5\},\\
A_2&=\{0.4,0.5,0.6\},& C(A_2)&=\{0.4,0.5,0.6\},\\
A_3&=\{0,0.1,0.2\},& C(A_3)&=\{0.2\}.\\
\end{aligned}
$$
It can be noted that there are cases where she chooses more than one cupcake. This may happen because she is considering different characteristics of the object, such as taste and softness. This may also happen because she is uncertain about her choices, for instance because she is selecting cupcakes to share with a group of friends (in economics this problem is referred to multiple-self decision making).
From a machine learning perspective, our goal is to learn a model that allows us to predict Alice's choices for the different set of cupcakes $A^*=\{x_a^*,x_b^*,x_c^*\}$.
\end{example}
In the following sections, we describe how to learn Alice’s choice  function $C$ from observed choice data $(A_k,C(A_k)): \text{ for } k=1,\dots,m$, and how to use it to predict her selections when presented with new sets of objects. To do so, we exploit the results from Section \ref{sec:choicefunintro}, which show that rational or pseudo-rational choice functions can be represented by a vector of utility functions.

\section{Rational and Pseudo-rational models for choice data}
\label{sec:chocieslearning}
We will start with Pareto rational choice functions, defined in Section \ref{sec:choicefunintro}. For each $A$, we  interpret $C(A)$ as the \textit{undominated set} in the \textit{strong Pareto sense}, with $R(A)$ being the set of dominated options. We consider a latent vector of utility functions ${\bf u}({\bf o})=[u_1({\bf o}),\dots,u_{d}({\bf o})]^\top$,  which embeds the objects ${\bf o}$ into a  space $\mathbb{R}^{d}$. We say that an object ${\bf o} \in A$ is not dominated by an object ${\bf v}$ if 
\begin{equation}
    \min_{i \in \{1, \ldots, d\}} (u_i({\bf o})-u_i({\bf v})) > 0, 
\end{equation}
that is all utilities $u_i$ take a larger value at ${\bf o}$ than at ${\bf v}$. 

\begin{example}
In our cupcake example, the utilities $u_1,u_2,\dots,u_d$ may represent different attributes of the cupcakes, such as taste and softness, that Alice implicitly considers when making her choices.
\end{example}

The choice set can then be represented through a Pareto set of strongly undominated options:
\begin{align}
 \label{eq:likcondpareto1}
 \neg & \left( \min_{i \in\{1,\dots,d\}} (u_i({\bf o})-u_i({\bf v}))\leq 0, ~~\forall {\bf o} \in C(A)\right), ~~\forall {\bf v} \in R(A),\\
  \label{eq:likcondpareto2}
 &\min_{i \in\{1,\dots,d\}} (u_i({\bf o})-u_i({\bf v}))\leq 0, ~\forall {\bf o}, {\bf v} \in C(A), ~~ {\bf o} \neq {\bf v}.
\end{align}

The first condition, in \eqref{eq:likcondpareto1}, means that, for each option ${\bf v} \in R(A)$, it is not true ($\neg$ stands for logical negation) that all options in $C(A)$ are no-better than ${\bf v}$. In particular, there is at least an option in $C(A)$ which is better than ${\bf v}$. Condition~\eqref{eq:likcondpareto2} means that, for each option in $C(A)$, there is no better option in $C(A)$. This requires that the latent functions values of the options should be consistent with the choice function implied relations. 

To account for possible violations of rationality, similarly to what is done in Model \ref{Model:Probit}, we replace the hard-constraints \eqref{eq:likcondpareto1}--\eqref{eq:likcondpareto2} with soft-constraints defining a likelihood model. In particular,  consider the dataset of choices:
$$
\mathcal{D}_m=\{(A_k,C(A_k)): \text{ for } k=1,\dots,m\},$$
given the latent vector function ${\bf u}$, the likelihood is
 \begin{equation}
  p(\mathcal{D}_m|{\bf u}(X))=\prod_{k=1}^m p(C(A_k),A_k|{\bf u}(X)),
 \end{equation}
with
\begin{equation}
  \label{eq:likelihoodexpanse00}
   \begin{aligned}
  p(&C(A_k),A_k|{\bf u}(X)) \\
  &= \prod\limits_{\{{\bf o},{\bf v}\} \in C_\sharp(A_k)}\Bigg( 1-\prod_{i=1}^d \Phi\left(\frac{u_i({\bf o})-u_i({\bf v})}{\sigma}\right) 
   -\prod_{i=1}^d \Phi\left(\frac{u_i({\bf v})-u_i({\bf o})}{\sigma}\right)\Bigg)\\
         &\prod_{{\bf v} \in R(A_k)}\Bigg(1- \prod_{{\bf o} \in C(A_k)} \left(1- \prod_{i=1}^d \Phi\left(\frac{u_i({\bf o})-u_i({\bf v})}{\sigma}\right)\right)\Bigg),\\
   \end{aligned}    
 \end{equation}
  where the notation $\{{\bf o},{\bf v}\} \in C_\sharp(A_k)$ means that the pair $\{{\bf o},{\bf v}\}$ is an element of $C_\sharp(A_k)$, which  denotes the set   of all possible 2-combination (without repetition) of the elements of the set $C(A_k)$.
We assumed that the choice-statements $k=1,\dots,m$ are conditionally independent given the utilities ${\bf u}(X)$, which is in line with our model based on Pareto rationalisation.    The product in the first and second row in \eqref{eq:likelihoodexpanse00} is a probabilistic relaxation of \eqref{eq:likcondpareto2}.
   Note the negation $1-\prod_{i=1}^d \Phi\left(\frac{u_i({\bf o})-u_i({\bf v})}{\sigma}\right)-\prod_{i=1}^d \Phi\left(\frac{u_i({\bf v})-u_i({\bf o})}{\sigma}\right)$, which states that ${\bf o}$ does not dominate ${\bf v}$ and vice versa.
  The product in the last row in \eqref{eq:likelihoodexpanse00} is a probabilistic relaxation of \eqref{eq:likcondpareto1}. 

\begin{example}[Inspired by example~1 in \citet{benavoli2023a}]
    To understand  the likelihood \eqref{eq:likelihoodexpanse00}, let us consider four objects ${\bf o}_1,{\bf o}_2,{\bf o}_3,{\bf o}_4$ and  the  utilities:
$$
{\bf u}({\bf o}_1)=\begin{bmatrix}
0.2\\
0
\end{bmatrix}, {\bf u}({\bf o}_2)=\begin{bmatrix}
0.1\\
0.2
\end{bmatrix}, {\bf u}({\bf o}_3)=\begin{bmatrix}
-1\\
-1
\end{bmatrix}, {\bf u}({\bf o}_4)=\begin{bmatrix}
-0.5\\
-0.5
\end{bmatrix}.
$$
The objects $\{{\bf o}_1,{\bf o}_2\}$ are hard to tell apart since they have very similar utilities. For this reason,  Alice may make mistakes when comparing them. For instance, she makes the following choices:
$$
C(\{{\bf o}_1,{\bf o}_2,{\bf o}_3\})=\{{\bf o}_1,{\bf o}_2\}, ~~C(\{{\bf o}_1,{\bf o}_2,{\bf o}_4\})=\{{\bf o}_1\}.
$$
Those choices are not Pareto rationalisable according to the order induced by the utility functions above. In particular, this implies that conditions \eqref{eq:likcondpareto1}--\eqref{eq:likcondpareto2} are not simultaneously satisfied. On the other hand, the likelihood~\eqref{eq:likelihoodexpanse00} is not zero on this dataset, because it accounts for errors in Alice's choices. 
Assuming  $\sigma=1$, the likelihood for the two choices is
$$
\begin{aligned}{}
p(\{{\bf o}_1,{\bf o}_2\},\{{\bf o}_1,{\bf o}_2,{\bf o}_3\}|{\bf u}(X))&\approx 0.48,\\
p(\{{\bf o}_1\},\{{\bf o}_1,{\bf o}_2,{\bf o}_4\}|{\bf u}(X))&\approx 0.12.\\
\end{aligned}
$$
This means that, for the above latent utilities, the probability that Alice jointly makes these choices is
$0.48 \cdot 0.12=0.057$.
 Note that, the probability of error increases with the parameter $\sigma$. Indeed, the probability  $p(\{{\bf o}_1,{\bf o}_2\},\{{\bf o}_1,{\bf o}_2,{\bf o}_3\}|{\bf u}(X))p(\{{\bf o}_1\},\{{\bf o}_1,{\bf o}_2,{\bf o}_4\}|{\bf u}(X))$ tends to zero  for $\sigma \rightarrow 0$ since, in this case, the likelihood \eqref{eq:likelihoodexpanse00}  reduces to \eqref{eq:likcondpareto1}--\eqref{eq:likcondpareto2} and Alice's choice is Pareto irrational.
\end{example}

We will now move to pseudo-rationalisable choice functions. In this case, we say that an object ${\bf o} \in A$ is not dominated by an object ${\bf v}$ if:
\begin{equation}
    \max_{i \in \{1, \ldots, d\}} (u_i({\bf o})-u_i({\bf v})) > 0, 
    \label{eq:notDominatedPseudo}
\end{equation}
that is, there is a utility where object ${\bf o}$ is better than ${\bf v}$. The choice set is then represented through the following conditions:
\begin{align}
 \label{eq:likcondmaxU1}
 & \left( \forall i \in \{1,\dots,d\} ~~\neg\left(\min_{{\bf o} \in C(A)} (u_i({\bf v})-u_i({\bf o}))\geq 0\right)\right), ~~\forall v \in R(A),\\
  \label{eq:likcondmaxU2}
 &\min_{i \in\{1,\dots,d\}} (u_i({\bf o})-u_i({\bf v}))\leq 0, ~\forall {\bf o},{\bf v}\in C(A), ~~ {\bf o} \neq {\bf v},
\end{align}
where condition~\eqref{eq:likcondmaxU1} means that for all ${\bf v} \in R(A)$, it is not true that the value of any latent function in ${\bf v}$ is higher than their value in any ${\bf o} \in C(A)$. Condition~\eqref{eq:likcondmaxU2} instead imposes that each object in $C(A)$ is not dominated according to \eqref{eq:notDominatedPseudo}. By using the same technique used above to relax the constraints we obtain the likelihood $ p(\mathcal{D}_m|{\bf u}(X))=\prod_{k=1}^m p(C(A_k),A_k|{\bf u}(X))$ with:
\begin{equation}
  \label{eq:likelihoodpseudorat}
   \begin{aligned} 
   p(&C(A_k),A_k|{\bf u}(X))= \\
  &\prod\limits_{\{{\bf o},{\bf v}\} \in C_\sharp(A_k)}\Bigg( 1-\prod_{i=1}^d \Phi\left(\frac{u_i({\bf o})-u_i({\bf v})}{\sigma}\right)-\prod_{i=1}^d \Phi\left(\frac{u_i({\bf v})-u_i({\bf o})}{\sigma}\right)\Bigg)\\
         &\prod_{{\bf v} \in R(A_k)}\Bigg(\prod_{i=1}^d  \left(1-\prod_{{\bf o} \in C(A_k)} \Phi\left(\frac{u_i({\bf v})-u_i({\bf o})}{\sigma}\right)\right)\Bigg).\\
   \end{aligned}    
 \end{equation}
Note that the first part of \eqref{eq:likelihoodpseudorat} remains the same as in the rational case, however the second part now enforces the dominance defined in~\eqref{eq:notDominatedPseudo}.

\begin{example}
       To understand  the above likelihood, let us consider again four objects ${\bf o}_1,{\bf o}_2,{\bf o}_3,{\bf o}_4$ and  the  utilities:
$$
{\bf u}({\bf o}_1)=\begin{bmatrix}
0.2\\
-1.5
\end{bmatrix}, {\bf u}({\bf o}_2)=\begin{bmatrix}
-1.5\\
0.2
\end{bmatrix}, {\bf u}({\bf o}_3)=\begin{bmatrix}
-1\\
-1
\end{bmatrix}, {\bf u}({\bf o}_4)=\begin{bmatrix}
-2\\
-2
\end{bmatrix}.
$$
Assume Alice makes the following choices:
$$
C(\{{\bf o}_1,{\bf o}_2,{\bf o}_3\})=\{{\bf o}_1,{\bf o}_2\}, ~~C(\{{\bf o}_1,{\bf o}_2,{\bf o}_4\})=\{{\bf o}_1,{\bf o}_2\}.
$$
Those choices are not Pareto rationalisable, but they are Pareto pseudo-rationalisable according to the order induced by the utility functions above. In particular, this implies that conditions \eqref{eq:likcondpareto1}-- \eqref{eq:likcondpareto2} are not simultaneously satisfied, while this is the case for  \eqref{eq:likcondmaxU1}--\eqref{eq:likcondmaxU2}.
Indeed, for $\sigma=1$, the
likelihood~\eqref{eq:likelihoodpseudorat}
$$
\begin{aligned}
p(\{{\bf o}_1,{\bf o}_2\},\{{\bf o}_1,{\bf o}_2,{\bf o}_3\}|{\bf u}(X))&\approx 0.77,\\
p(\{{\bf o}_1,{\bf o}_2\},\{{\bf o}_1,{\bf o}_2,{\bf o}_4\}|{\bf u}(X))&\approx 0.91.\\
\end{aligned}
$$
and their product is $0.77\cdot 0.91 \approx 0.71$, while the likelihood~\eqref{eq:likelihoodexpanse00} is equal to
$$
\begin{aligned}{}
p(\{{\bf o}_1,{\bf o}_2\},\{{\bf o}_1,{\bf o}_2,{\bf o}_3\}|{\bf u}(X))&\approx 0.43,\\
p(\{{\bf o}_1,{\bf o}_2\},\{{\bf o}_1,{\bf o}_2,{\bf o}_4\}|{\bf u}(X))&\approx 0.82.\\
\end{aligned}
$$
and their product is $0.43 \cdot 0.82=0.35$. For $\sigma \rightarrow 0$, the two products converge to $1$ and, respectively $0$. This holds because Alice's choices are  pseudo-rationalisable but not Pareto rationalisable.
\end{example}

 \begin{SnugshadeB}
\begin{MethodB}[Rational and Pseudo-rational Models for Choices]
\label{model:choice}
Consider the training dataset:
\begin{equation}
    \label{eq:datasetchoice}
\mathcal{D}_m=\{(A_k,C(A_k)): \text{ for } k=1,\dots,m\},
\end{equation}
where, for each $k$, $\emptyset \neq C(A_k) \subseteq A_k$ and $A_k \subset \{\bx_1,\bx_2,\dots,\bx_r\}$ with $\bx_i \in \mathcal{X}$. Consider the vector $X=[\bx_1,\bx_2,\dots,\bx_r]^\top$ and 
$$
{\bf u}({\bf x}_i)=\begin{bmatrix}
u_1({\bf x}_i) \\ 
u_2({\bf x}_i)\\ 
\vdots \\
u_d({\bf x}_i)\\ 
\end{bmatrix},~~~{\bf u}(X)=\begin{bmatrix}
{\bf u}({\bf x}_1) \\ 
{\bf u}({\bf x}_2) 
\\ 
\vdots \\
{\bf u}({\bf x}_r) \\ 
\end{bmatrix}.
$$
Under the assumption that the choice statements are conditionally independent given ${\bf u}(X)$, we obtain the likelihood:
\begin{equation}
\begin{aligned}
p(\mathcal{D}_m|{\bf u}(X))&=\prod_{k=1}^m p(C(A_k),A_k|{\bf u}(X)) ~~\text{ with }  \\p(C(A_k),A_k|{\bf u}(X))&=\left\{
  \begin{array}{ll}
\text{\eqref{eq:likelihoodexpanse00}} & \text{for rational choices}\\
\text{\eqref{eq:likelihoodpseudorat}} & \text{for pseudo-rational choices}
  \end{array}
  \right.
  \end{aligned}
\end{equation}
 We assign independent GP priors to each utility function resulting into:
\begin{equation}
\label{eq:multiprior3}
u_i \sim \mathrm{GP}(u;0,k_i), ~~~~~
{\bf u}=\begin{bmatrix}
u_1 \\ 
u_2\\ 
\vdots \\
u_d\\ 
\end{bmatrix} \sim GP\left({\bf 0},\mqty[\dmat[0]{k_1,k_2,\ddots,k_d}]\right).    
\end{equation}
Given the above likelihoods and prior, inference is carried out as described in Algorithm \ref{alg:1}, where steps A1--A3 are performed using variational inference as approximation method for the posterior.
\end{MethodB}
\end{SnugshadeB}
Note that, Model \ref{model:choice} comprises two distinct models depending on the selected likelihood. The derivation of the likelihood \eqref{eq:likelihoodpseudorat} is based on less assumptions regarding the choice functions, making it more general. However, it is less informative in applications where the subject's choices are based on Pareto rationalisation. In this case, it is better to use \eqref{eq:likelihoodexpanse00}.
Learning rationalisable choice functions  via a Pareto embedding was originally proposed by \cite{pfannschmidt2020learning}, but using a hinge-loss and a neural network based model. The GP model was derived in \citet{benavoli2023a}.\footnote{We can exploit the structure of the likelihood to reduce the number of variational parameters in the covariance matrix of the Gaussian variational distribution, similarly to what done in \citet{opper2009variational}. Also in this case, the variational covariance matrix cannot be diagonal, we need some additional parameters to model the correlation between the various utilities functions.}
Recently, \cite{pfannschmidt2022learning} have extended their work to learn context-based choices. This framework deals with situations in which a subject's choices are context-dependent,  which means that the preference in favour of a certain choice alternative may depend on what other options are also available in the choice-set. In general, these models  reject the Chernoff-axiom \citep{tversky1993context}.

In Model \ref{model:choice}, we have assumed that the number of latent utilities is fixed and known. Model \ref{model:choice} works well if the number of latent functions in the model is greater than or equal of the number of implicit criteria that Alice uses to make her choices. If this upper bound is unknown, a way to learn the number of utility functions is through the Leave-One-Out cross-validation score, which can be computed from the posterior distribution \citep{benavoli2023a}.

\begin{example}
\label{ex:cupcakechoice}
We consider a scenario as in Example \ref{ex:cupcake0}, where Alice chooses her favourite cupcakes from given choice-sets (where the cupcakes differ for the amount of butter). We assume her choices are Pareto-rational and determined by the two latent  utilities shown in Figure \ref{fig:cupcakes}. 
\begin{figure}[h]
\centering
\includegraphics[width=6cm]{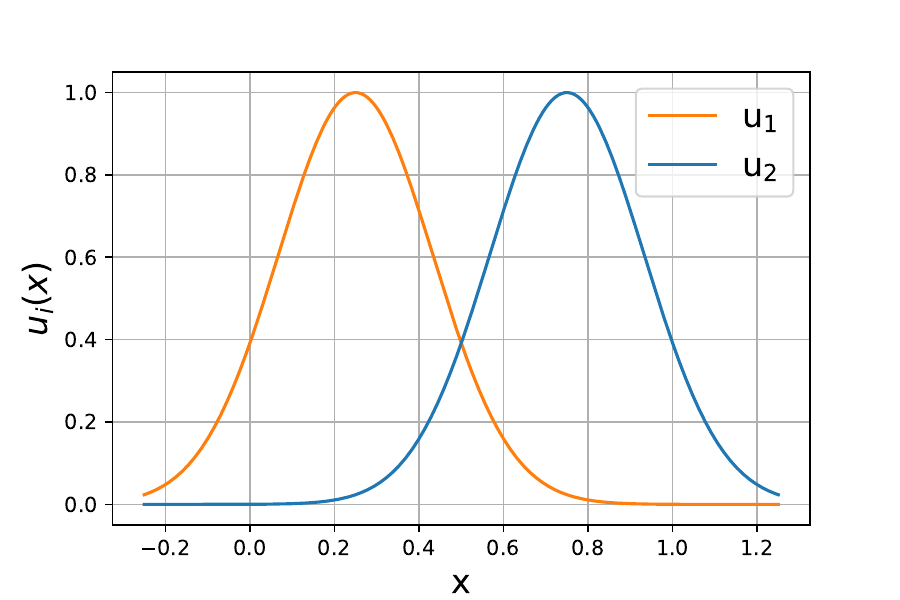}
\caption{True latent utilities function that underlay Alice's Pareto-rational choices.}
\label{fig:cupcakes}
\end{figure}
We use these utilities ${\bf u}=[u_1,u_2]^\top$ to define a choice function via Pareto dominance. For instance, consider the set of options $A=\{x_1,x_2,x_3\}$, assume that ${\bf u}(x_1)=[1,0.05]$, ${\bf u}(x_2)=[0.5,0]$, $
{\bf u}(x_3)=[0.05,1]$,
we have that $C(A)=\{x_1,x_3\}$ and $R(A)=A \backslash C(A_k)=\{x_2\}$. In fact, one can notice that  $[1,0.05]$ dominates  $[0.5,0]$ on both the utilities, and $[1,0.05]$  and $[0.05,1]$ are incomparable.  We sample $60$ inputs $x_i$ at random in $[-0.25,1.25]$ and, using the above approach, we generated a dataset 
of $m=200$ random subsets $\{A_k\}_{k=1}^m$ of the $60$ points each one of size $|A_k|=3$  and compute the corresponding choice pairs $(C(A_k),A_k)$ based on  ${\bf u}$.

Our goal is to infer Alice's utility functions from the dataset $\mathcal{D}_{200}$ and predict her choices.

Figure \ref{fig:choicecupcakeest}-left shows the
posterior-mean of the utilities and 95\% credible intervals for the learned two utility functions. It can be noticed that the learned utilities are in agreement with the true ones in Figure \ref{fig:cupcakes}. Figure \ref{fig:choicecupcakeest}-right shows 5 samples of the two utilities from the posterior. These samples are correlated -- they always cross near the middle. This happens because the variational posterior captures the relative difference between the two utilities that determines the Pareto-based choices (modelled through the likelihood \eqref{eq:likelihoodexpanse00}).

\begin{figure}[h]
\centering
\includegraphics[width=0.495\textwidth]{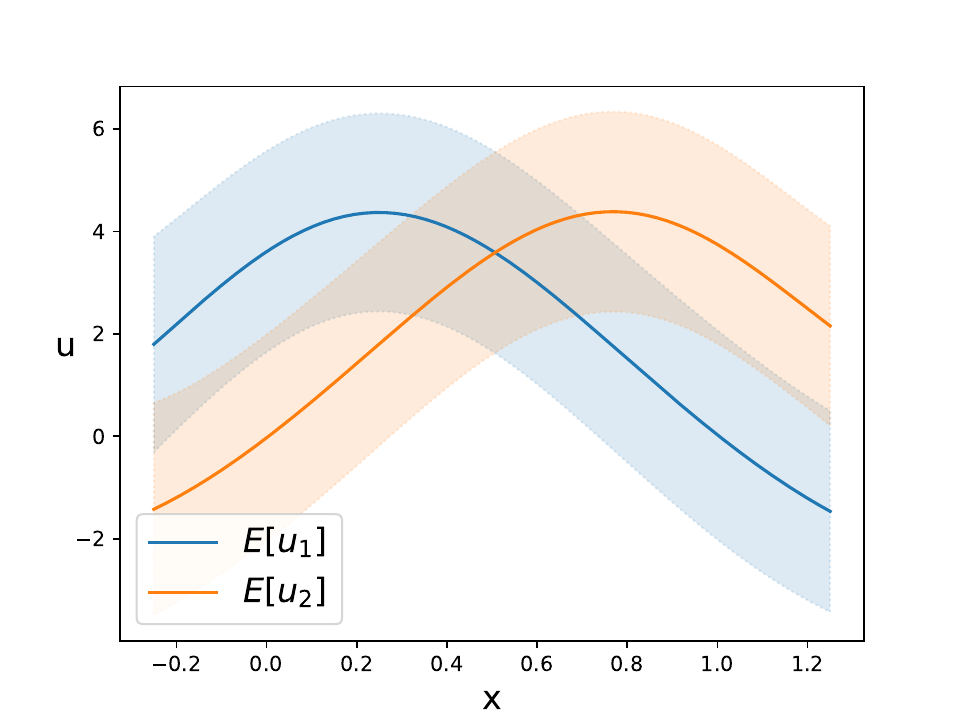}
\includegraphics[width=0.495\textwidth]{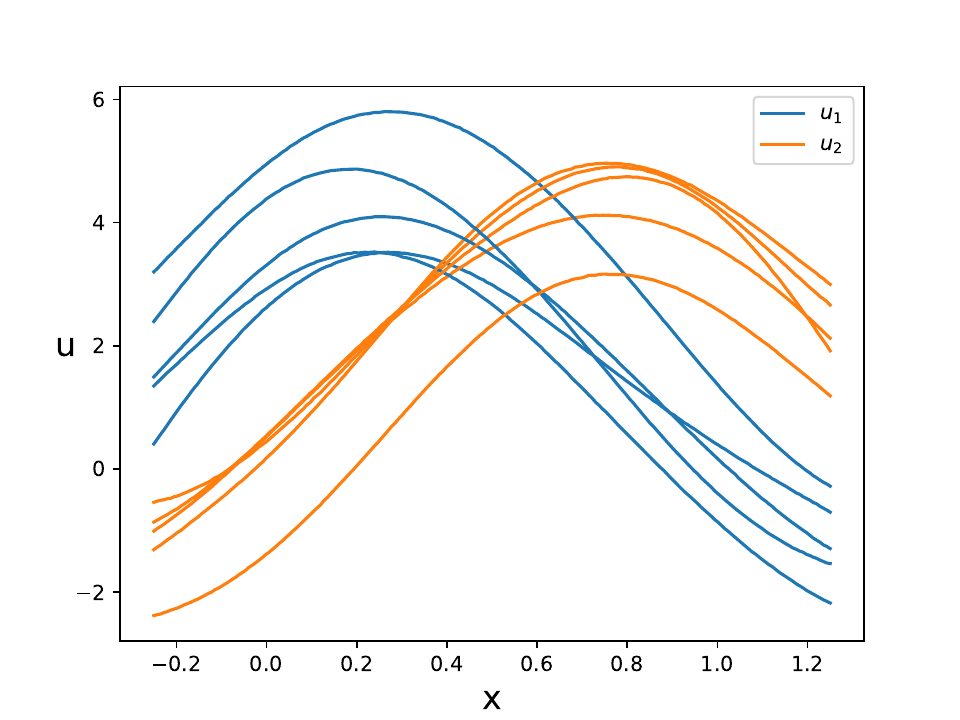}
\caption{Learned utilities from rational choice data using Model \ref{model:choice} with rational likelihood. Left: mean and credible region of the predicted posterior utilities. Right: a sample from the posterior.}
\label{fig:choicecupcakeest}
\end{figure}

We have repeated this experiment generating pseudo-rational choices and then used the rational-likelihood and, respectively, pseudo-rational-likelihood based model to learn back the utilities. The posterior in the two cases are shown in Figure  
 \ref{fig:choicecupcakeest1} and, respectively, \ref{fig:choicecupcakeest2}. In Figure \ref{fig:choicecupcakeest1}, it can be noticed that the relative difference between the two expected utilities and  the posterior samples is wrong. They do not cross in the middle of the interval $[-0.25,1.25]$. This holds because the data are not Pareto rational and the learned model is therefore ``biased''. Figure \ref{fig:choicecupcakeest2} shows instead that the pseudo-rational model is able to correctly learn the utilities.

\begin{figure}[h]
\centering
\includegraphics[width=0.495\textwidth]{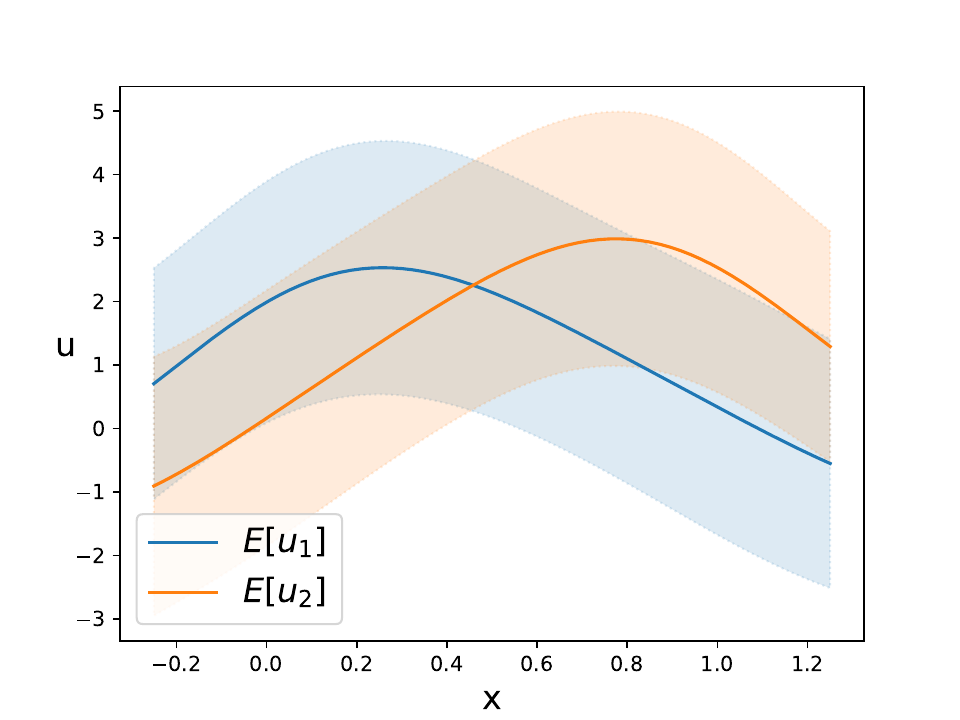}
\includegraphics[width=0.495\textwidth]{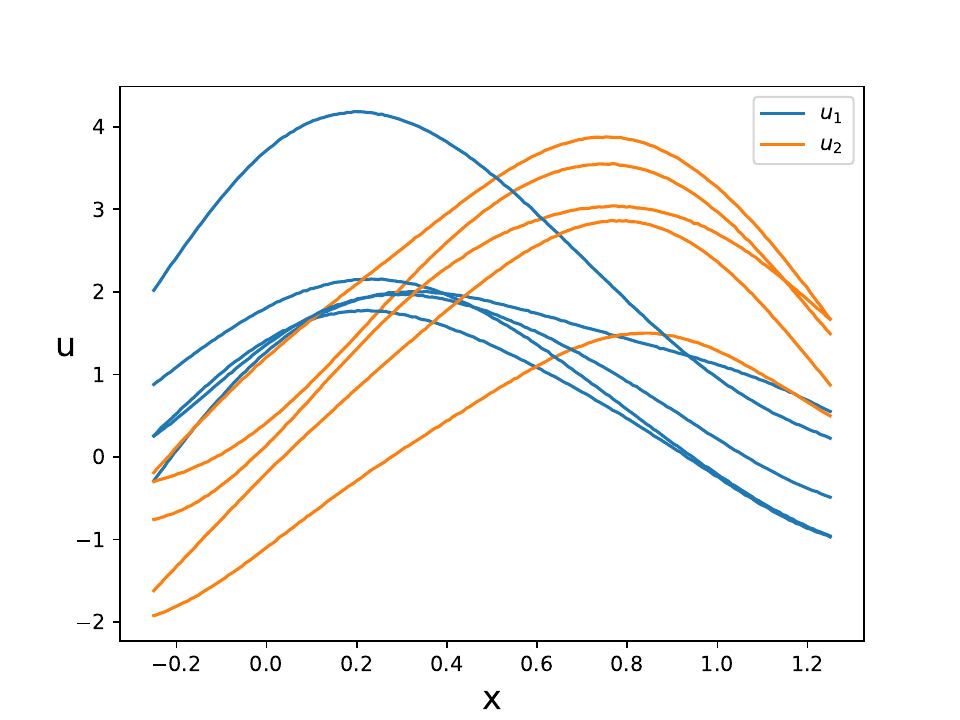}
\caption{Learned utilities from pseudo-rational choice data using Model \ref{model:choice} with rational likelihood.  Left: mean and credible region of the predicted posterior utilities. Right:  5 samples from the posterior.}
\label{fig:choicecupcakeest1}
\end{figure}
   
\begin{figure}[h]
\centering
\includegraphics[width=0.495\textwidth]{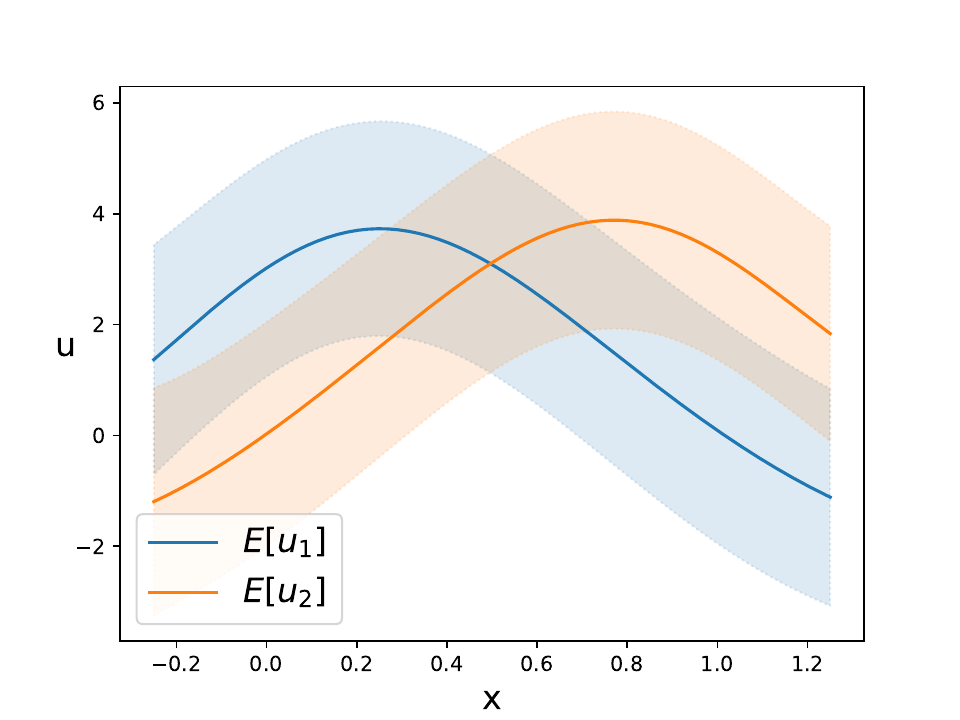}
\includegraphics[width=0.495\textwidth]{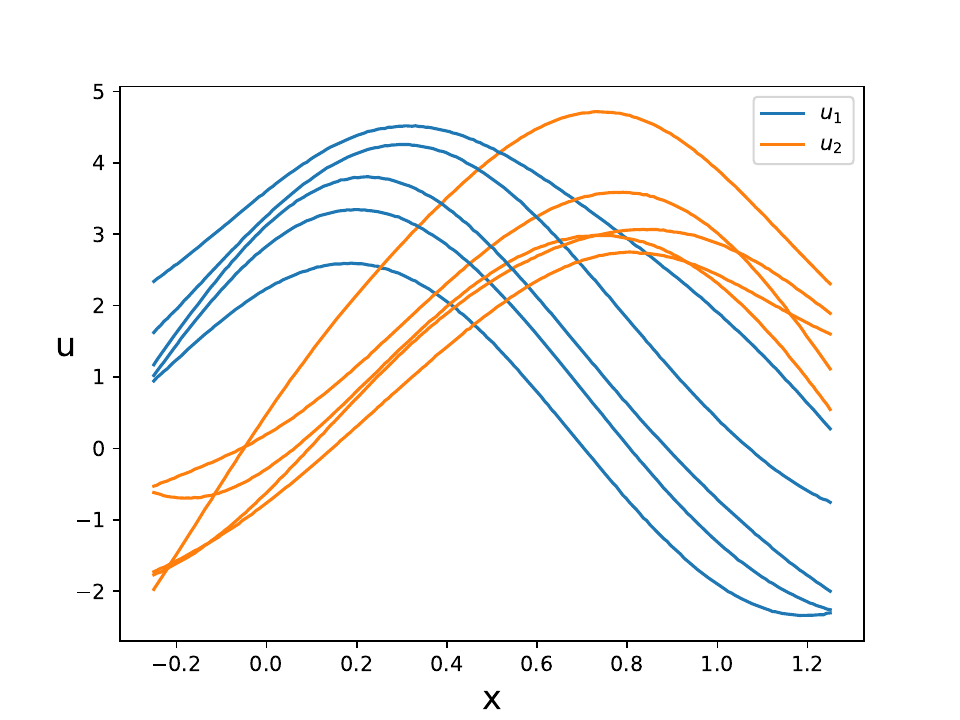}
\caption{Learned utilities from pseudo-rational choice data using Model \ref{model:choice} with pseudo-rational likelihood.  Left: mean and credible region of the predicted posterior utilities. Right:  5 samples from the posterior.}
\label{fig:choicecupcakeest2}
\end{figure}

\end{example}

\section{Computational complexity in choice function learning}
As in label preference, we have $d$ utility functions and a GP for each utility function. The key difference is that evaluating the likelihood is more demanding, as it involves a product of $C_\sharp(A_k)$ terms, i.e., the number of all possible 2-combinations (without repetition) of elements in the set $C(A_k)$. Consequently, the computational complexity also depends on the size of $A_k$, which may vary depending on the application. In this case, to ensure scalability, one can apply stochastic variational inference \citep{hensman2015scalable}, and perform subsampling along  the dimensions $m,d$ and the number of elements of each $C_\sharp(A_k),R(A_k)$.

\chapter{Applications}
\label{sec:applications}
Hereafter, we will apply the models introduced earlier to real datasets.

\section{Transportation mode preference}
\label{sec:trasport}
First, we consider a dataset including the choices made by subjects regarding their preferred transport mode for the Montreal-Toronto corridor \citep{forinash1993application,croissant2020estimation}. 
There are four transport modes (air, bus, car and train) and we have the following information about the journey: \textit{dist}  the distance
of the trip, \textit{cost}  the monetary cost, \textit{ivt}  the in-vehicle-time, \textit{ovt}  the out-of-vehicle
time, \textit{freq} the frequency and \textit{noalt}  the number of available alternatives. There are also  subject specific variables: \textit{income}  the household income and \textit{urban}  an indicator variable for trips which have a large city at the origin or the destination. 
Table \ref{tab:canada} shows a subset of the dataset for \textit{noalt=4}, namely those for which four transportation modes are available. Each \textit{case} corresponds to a different choice set for an individual. 
The dataset includes  4221 different  transport modes and  4322 cases (leading to 15520 rows). This dataset will allow us to explore how the various models we have previously introduced can be applied to address diverse choice tasks.

{
 \rowcolors{2}{gray!25}{white}
\begin{table}[h]
\centering
\footnotesize
\begin{tabular}{rlrrrrrrrr}
\rowcolor{gray!50}
\toprule
 case &   alt &  choice &  dist &   cost &  ivt &  ovt &  freq &  income &  urban \\
\midrule
 129 & train &       0 &   387 &  58.25 &  316 &   59 &     2 &      55 &      0 \\
  129 &   air &       0 &   387 & 145.80 &   56 &   85 &     9 &      55 &      0  \\
  129 &   bus &       0 &   387 &  26.67 &  301 &   58 &     8 &      55 &      0  \\
  129 &   car &       1 &   387 &  73.53 &  251 &    0 &     0 &      55 &      0  \\
  130 & train &       0 &   387 &  58.25 &  316 &   59 &     2 &      45 &      0  \\
  130 &   air &       0 &   387 & 145.80 &   56 &   85 &     9 &      45 &      0  \\
  130 &   bus &       0 &   387 &  26.67 &  301 &   58 &     8 &      45 &      0  \\
  130 &   car &       1 &   387 &  73.53 &  251 &    0 &     0 &      45 &      0  \\
  131 & train &       0 &   404 &  60.25 &  316 &   74 &     2 &      70 &      0  \\
  131 &   air &       0 &   404 & 146.20 &   56 &   94 &     9 &      70 &      0  \\
  131 &   bus &       0 &   404 &  26.67 &  301 &   93 &     8 &      70 &      0  \\
  131 &   car &       1 &   404 &  76.76 &  265 &    0 &     0 &      70 &      0  \\
  132 & train &       1 &   365 &  60.85 &  237 &   94 &     4 &      70 &      1  \\
  132 &   air &       0 &   365 & 151.20 &   56 &   77 &     9 &      70 &      1  \\
  132 &   bus &       0 &   365 &  27.82 &  301 &   83 &     8 &      70 &      1  \\
  132 &   car &       0 &   365 &  69.35 &  245 &    0 &     0 &      70 &      1  \\
\bottomrule
\end{tabular}
\caption{Each case is a different choice task.}
\label{tab:canada}
\end{table}}

Table~\ref{tab:canada}, shows four cases where all transportation choices were available (\textit{noalt=4}). 
For instance in case $129$,  a subject (income=55, urban=0) selected the \textit{car}-mode (with features dist=387, cost=73.53, ivt=251, ovt=0, freq=0).
In this first analysis, we will disregard subject-specific variables (income and urban), approaching the problem as an object-preference problem. Therefore, we will select the subset of subjects with income=70 and urban=1 (a total of $679$ cases).
Since subjects make always a transportation-mode choice, instances where the same individuals select both transportation modes cannot occur. In simpler terms, we will not observe instances of indifference (for example, if several transportation modes have closely ranked utilities for the subject), nor will we observe cases of incompatibility (where a subject makes both choices). 

In this first analysis,  we will interpret a choice, for instance \textit{car}, as the set of binary preferences over the covariates of the transportation modes, that is ${\bf x}_{car} \succ {\bf x}_{air}$, ${\bf x}_{car} \succ {\bf x}_{bus}$, ${\bf x}_{car} \succ {\bf x}_{train}$.  We will assume independent pairwise comparisons and use Model \ref{Model:Probit} (which is equivalent to Model \ref{model:Gaussian} under independent noises). We  first consider a linear utility in the covariates:
\begin{equation}
\label{eq:linearutil}
u({\bf x})=\beta_1 \texttt{cost}
+\beta_2 \texttt{ivt}+\beta_3 \texttt{ovt}+\beta_4 \texttt{freq},
\end{equation}
with ${\bf x}=[\texttt{cost},\texttt{ivt},\texttt{ovt},\texttt{freq}]^\top$,\footnote{We did not include \texttt{dist} as covariate, because it is always equal for the 4 transportation modes to be compared.} but we  estimate the unknown coefficients $\beta_i$ through Bayesian inference under the independent priors $\beta_i \sim N(0,\sigma_i^2)$. 

\begin{figure}
\centering
\includegraphics[width=0.495\textwidth]{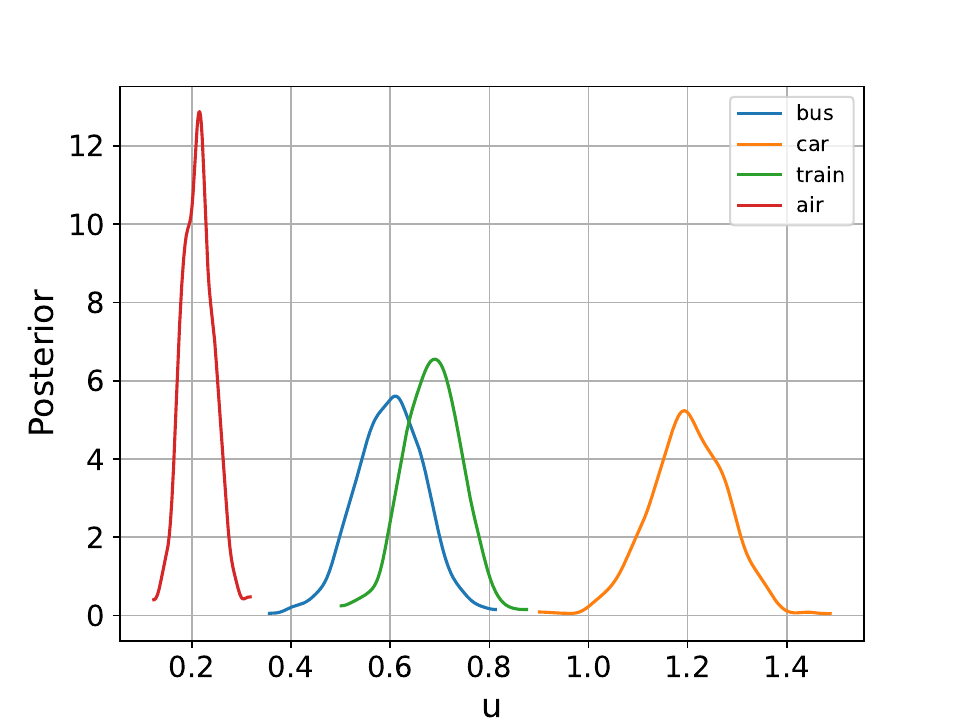}
\includegraphics[width=0.495\textwidth]{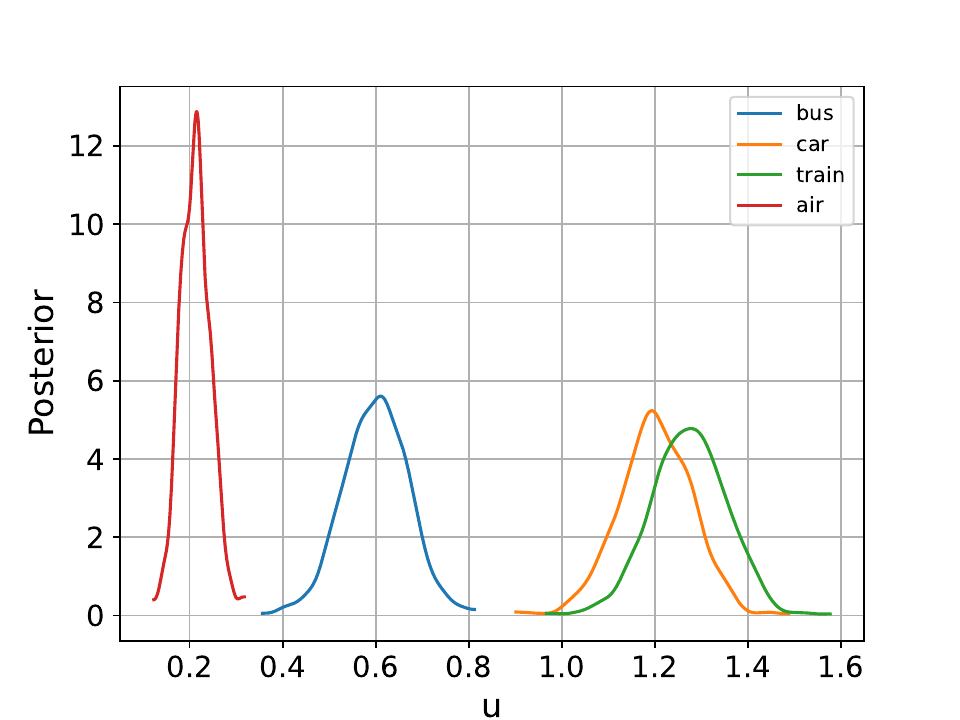}
\caption{Predicted utilities for the Transportation modes in Table \ref{tab:predicted175} for the linear kernel (left). The right plot shows the predicted utility for a new transportation mode.}
\label{fig:predtraspmode}
\end{figure}

Note that, for any two objects ${\bf x},{\bf x}'$, we have that $E[u({\bf x})]=E[u({\bf x}')]=0$ and 
\begin{equation}
\label{eq:linearkernel}
E[u({\bf x})u({\bf x}')]=k({\bf x},{\bf x}')=\sum_{i=1}^5 \sigma^2_i x_i x'_i,
\end{equation}
which is known as \textit{linear kernel}. Therefore, using the kernel \eqref{eq:linearkernel},  Model \ref{Model:Probit} is equivalent to a classical Bayesian linear utility model with Probit likelihood. In Model \ref{Model:Probit}, the variances $\sigma^2_i$ are inferred by maximising the marginal likelihood computed by using the Laplace's approximation.  
The estimated  variances for the training data (after standardising the data) are:
$$
\sigma^2_{\texttt{cost}}\approx 0.13,~
\sigma^2_{\texttt{ivt}}\approx 1.94,~
\sigma^2_{\texttt{ovt}}\approx 0.31,~
\sigma^2_{\texttt{freq}}\approx 0,
$$
which highlights that the feature \textit{freq} is not relevant. 
By randomly splitting  the dataset in 70\% training and 30\% testing ($10$ repetitions), we computed the model's predictive accuracy for subjects' choices in the test set: 84\%. For one of the splitting, we plotted the posterior predicted utilities in Figure \ref{fig:predtraspmode} (left), for the transportation modes listed in Table \ref{tab:predicted175}, taken from the testing set. Assume, for instance, that we wish to predict the effect of a reduction of the (ivt,ovt) for the train to $(30,40)$ minutes with an increasing in cost to $50$. Figure \ref{fig:predtraspmode} (right) shows that \textit{car} and the \textit{new train} transportation mode have now similar utilities.

{\rowcolors{2}{gray!25}{white}
\begin{table}[h]
\centering
\footnotesize
\begin{tabular}{rrrrrr}
\rowcolor{gray!50}
\toprule
 alt& dist &   cost &   ivt &   ovt &  freq \\
\midrule
bus & 175 &  18.43 & 119 &  75 &   4 \\
car & 175 &  33.25 & 115 &   0 &   0 \\
train & 175 &  38.15 & 109 &  65 &   5 \\
air & 175 & 155.90 &  60 & 118 &   6 \\
\bottomrule
\end{tabular}
\caption{Four transportation modes}
\label{tab:predicted175}
\end{table}}

\begin{remark}
A linear utility function is a common assumption for random utility models developed in the field of economics. The key difference compared to the model discussed above (based on a linear kernel) is that our model adopts a Bayesian approach for inference, whereas traditional linear random utility models in economics are typically fitted using maximum likelihood estimation. For example, in the transportation mode preference problem above, the maximum likelihood estimates of the coefficients  $\beta_i$ of the linear utility model in \eqref{eq:linearutil}, as returned by the R package `mlogit', are shown hereafter:
$$
\beta_{cost}=-0.65,~~
\beta_{ivt}=-2.24,~~
\beta_{ovt}=-0.95,~~
\beta_{freq}=0.13.
$$
The relative order of magnitude of the estimated parameters is comparable to the estimated variances in the GP  with a linear kernel.
 The coefficient for \textit{freq} is not zero, but the p-value for the Wald-test for $\beta_{freq}$ is 0.23 (above the standard significance threshold of 0.05). This implies that we cannot reject the null hypothesis that \textit{freq} has no effect on the utility, and therefore its contribution to the model is not statistically significant. A similar conclusion was obtained from the GP model based on the linear kernel. Another key difference between the parametric model and the GP-based model with a linear kernel is that the latter computes a posterior distribution directly over the utility function $u({\bf x})$ rather than over the parameters $\beta_i$. For a detailed explanation of this distinction, we refer the reader to \citep[Chapter 2]{rasmussen2006gaussian}.
\end{remark}

\begin{figure}
\centering
\includegraphics[width=0.495\textwidth]{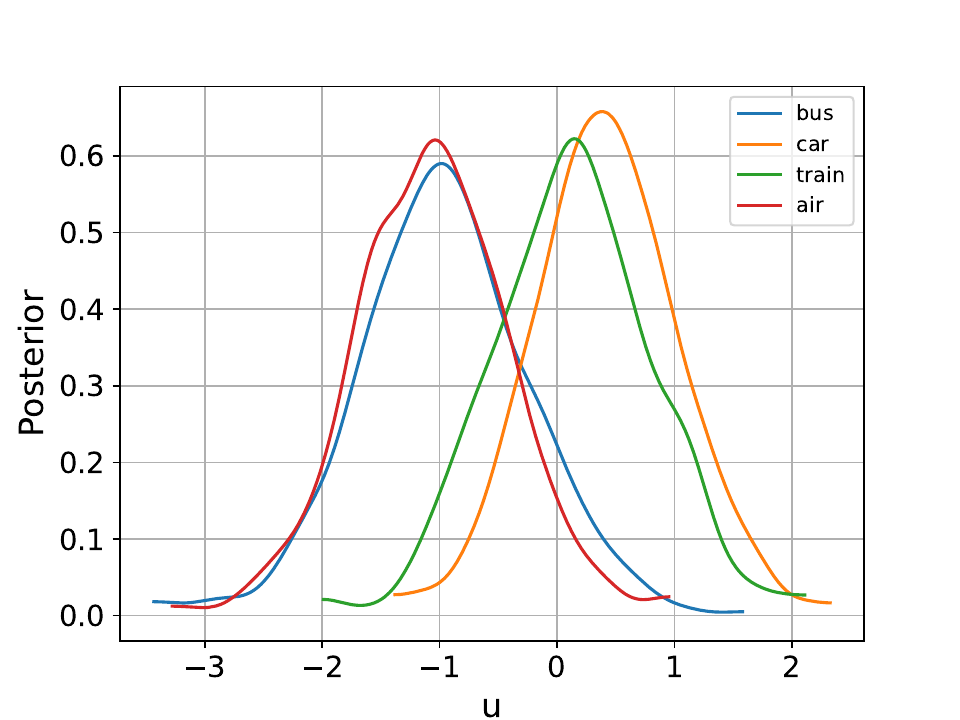}
\includegraphics[width=0.495\textwidth]{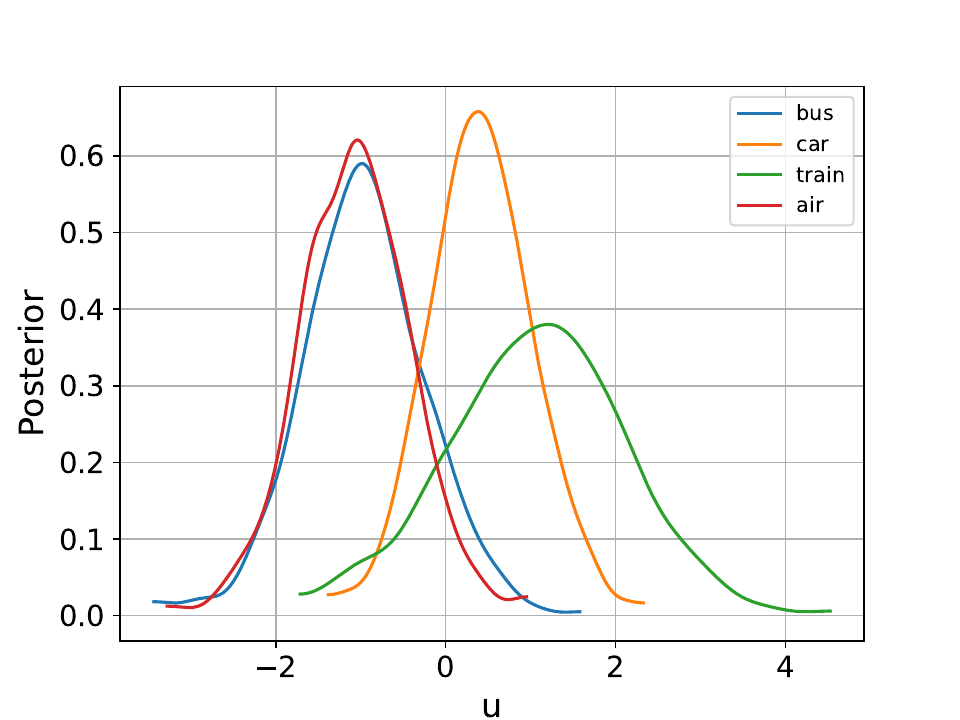}
\caption{Predicted utilities for the Transportation modes in Table \ref{tab:predicted175} for the SE kernel (left). The right plot shows the predicted utility for a new transportation mode.}
\label{fig:predtraspmodeSE}
\end{figure}

The advantage of working with GPs is that we can move from linear to nonlinear utility in the covariates by simply changing the kernel. For instance, we have considered the Squared-Exponential kernel:
\begin{equation}
\label{eq:SEkernel}
E[u({\bf x})u({\bf x}')]=k({\bf x},{\bf x}')=\sigma^2e^{-\sum_{i=1}^5 \frac{(x_i-x_i')^2}{2\ell_i^2}},
\end{equation}
and estimated (after standardising the data) its $6$  hyperparameters. 
$$
\ell_{\texttt{cost}}\approx0.70,~
\ell_{\texttt{ivt}}\approx 1.57,~
\ell_{\texttt{ovt}} \approx 1.70,~
\ell_{\texttt{freq}}\approx 0.26,~
\sigma^2=5.7,
$$
By randomly splitting  the dataset in 70\% training and 30\% testing ($10$ repetitions), we computed the model's predictive accuracy for subjects' choices in the test set:  90\%. This is considerably higher than the accuracy of the linear model. Figure \ref{fig:predtraspmodeSE} (left) shows the posterior predicted utilities for the transportation modes  in Table \ref{tab:predicted175}, and Figure \ref{fig:predtraspmodeSE} (right) reports the posterior for the new train transportation mode. 

Finally, instead of interpreting  a choice as the set of binary preferences over the covariates of the other transportation modes (example: ${\bf x}_{car} \succ {\bf x}_{air}$, ${\bf x}_{car} \succ {\bf x}_{bus}$, ${\bf x}_{car} \succ {\bf x}_{train}$), we could directly consider choice functions $A=\{{\bf x}_{air},{\bf x}_{bus},{\bf x}_{car},{\bf x}_{train}\}$ and $C(A)=\{{\bf x}_{car} \}$ using Model \ref{model:choice}. However, since each subject always selects a single transportation mode, this model is equivalent to Model \ref{Model:Probit}. Table \ref{tab:resCanada} summarises the accuracy results.
{
\rowcolors{2}{gray!25}{white}
\begin{table}
\centering
\small
\begin{tabular}{|c|c|c|c|}
\rowcolor{gray!50}
 \hline
 & Model \ref{Model:Probit} Linear Kernel & Model \ref{Model:Probit} SE Kernel & Model \ref{model:choice} SE Kernel\\
 \hline
Accuracy & 0.84 & 0.90 & 0.90  \\
  \hline
\end{tabular}
\caption{Prediction accuracy for the transportation dataset.}
\label{tab:resCanada}
\end{table}}

There are various approaches to incorporating the subjects' covariates \textit{income,urban} into this analysis and join preference data from different income groups. For example, in the SE kernel, one could employ distinct variance parameters for each income group (using instead the same lengthscales across groups). In light of the interpretation of Model \ref{Model:Probit}, this would suggest that individuals belonging to different income groups may have varying notions of close-utility. We considered preference data from three income groups (with incomes of $35,45,55$, all having similar dataset sizes) and assessed the performance of Model \ref{Model:Probit}  using a block diagonal SE kernel. Each income group had its own block, sharing a common lengthscale but with either shared variance or different variance parameter. The predictive accuracy  achieved were 86\% and 88\%, respectively.

\section{Ranked data for gaming platforms}
For label-preference, we consider the results of a survey among 91 Dutch students \citep{fok2012rank}. This survey considered the purchase of a new gaming platform and students were required to rank six  options: the Xbox, PlayStation, GameCube, PlayStation Portable, Game Boy, and a regular PC. Additionally, we know (1) their age; (2) which of the 6 platforms the student owns; (3) the average number of hours that each student spends on gaming each week. For example, Table \ref{tab:game} shows the results of the survey for three students with age 33, 19 and 18.

{
 \rowcolors{2}{gray!25}{white}
\begin{table}[h]
\centering
\footnotesize
\begin{tabular}{rrlrrr}
\rowcolor{gray!50}
\toprule
 age &  hours &    platform &  choice &  own &  choceid \\
\midrule
  33 &   2.00 &     GameBoy &   6 &    0 &     1 \\
  33 &   2.00 &    GameCube &   5 &    0 &     1 \\
  33 &   2.00 &          PC &   4 &    1 &     1 \\
  33 &   2.00 & PlayStation &   1 &    1 &     1 \\
  33 &   2.00 &  PSPortable &   3 &    0 &     1 \\
  33 &   2.00 &        Xbox &   2 &    0 &     1 \\
  19 &   3.25 &     GameBoy &   6 &    0 &     2 \\
  19 &   3.25 &    GameCube &   5 &    0 &     2 \\
  19 &   3.25 &          PC &   1 &    1 &     2 \\
  19 &   3.25 & PlayStation &   2 &    1 &     2 \\
  19 &   3.25 &  PSPortable &   3 &    0 &     2 \\
  19 &   3.25 &        Xbox &   4 &    0 &     2 \\
  18 &   4.00 &     GameBoy &   6 &    0 &     3 \\
  18 &   4.00 &    GameCube &   4 &    0 &     3 \\
  18 &   4.00 &          PC &   5 &    1 &     3 \\
  18 &   4.00 & PlayStation &   1 &    1 &     3 \\
  18 &   4.00 &  PSPortable &   2 &    0 &     3 \\
  18 &   4.00 &        Xbox &   3 &    0 &     3 \\
\bottomrule
\end{tabular}
\caption{Dataset: ranked data for gaming platforms }
\label{tab:game}
\end{table}}

\begin{figure}
    \centering
    \includegraphics[width=0.495\textwidth]{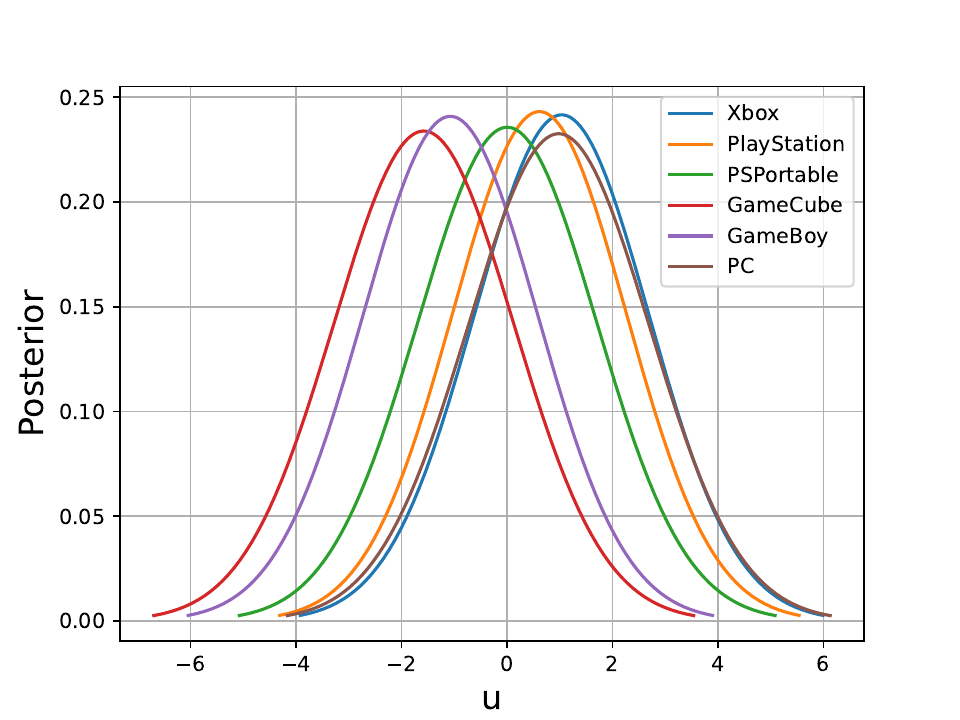}
    \includegraphics[width=0.495\textwidth]{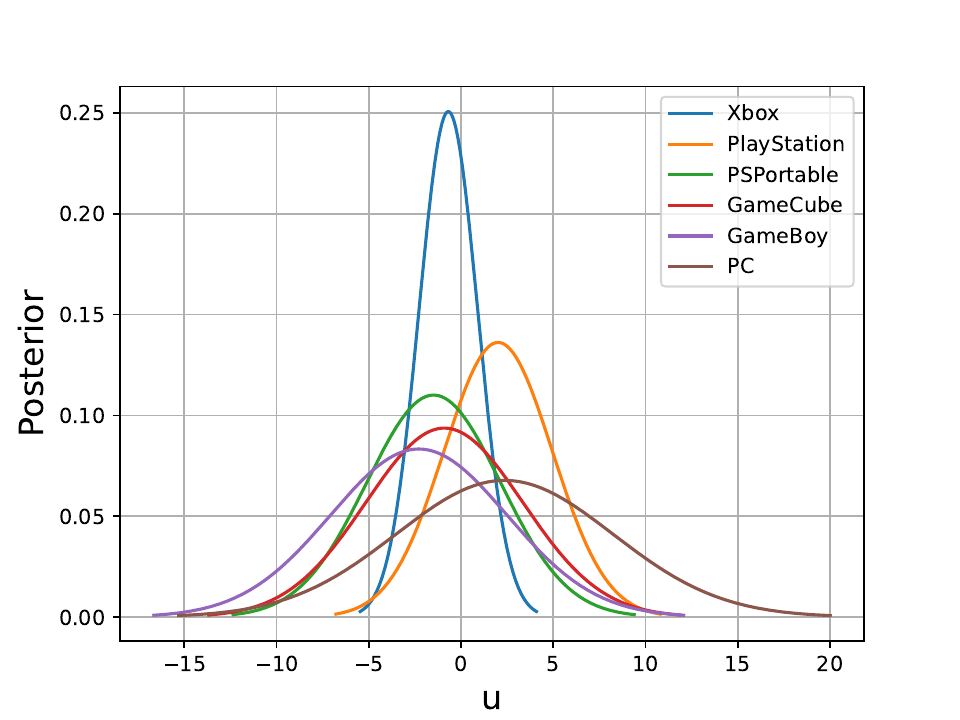}
    \caption{Predictive posterior for the utilities of the platforms. Covariates \texttt{ownPC}=1, \texttt{age}=18 and \texttt{hours}=2 (left), and for covariates \texttt{ownPlayStation}=1, \texttt{ownPC}=1 (right).}
    \label{fig:gaming}
\end{figure}

Since these are ranked data, we can use Model \ref{model:PL} to predict their ranking. There are 6 different labels (Xbox, PlayStation, GameCube, PlayStation Portable, Game Boy) and the subject's covariates are \textit{age}, \textit{hours}  and \textit{own}. For instance, for the first subject (age=33, hours=2, own=$\{$PC,PlayStation$\}$) in Table \ref{tab:game}, the ranking is  PlayStation, Xbox, PlayStation Portable, PC, GameCube and Game Boy. Given the covariates of a subject 
$$
\begin{aligned}
{\bf x}=[&\texttt{ownGameBoy},
\texttt{ownGameCube},
\texttt{ownPC},
\texttt{ownPlayStation}, \\
&\texttt{ownPSPortable},
\texttt{ownXbox},
\texttt{age},
\texttt{hours}
],
\end{aligned}
$$
the goal is to learn their gaming platforms ranking. This can be accomplished by learning the utilities 
{\small
$$
\begin{aligned}
{\bf u}({\bf x})=[u_\texttt{GameBoy}({\bf x}),
u_\texttt{GameCube}({\bf x}),
u_\texttt{PC}({\bf x}),
u_\texttt{PlayStation}({\bf x}),
u_\texttt{PSPortable}({\bf x}),
u_\texttt{Xbox}({\bf x})
].
\end{aligned}
$$}
In this case, we have directly used a nonlinear utility in the covariates by setting a SE kernel for each GP (one for each label). We have partitioned the dataset  into 90\% for training and 10\% for testing, and we computed the predicted ranking for the students in the test set. Figure \ref{fig:gaming} shows the posterior marginal distributions of the predicted utilities for the 6 labels. In particular, the left-plot shows the posteriors for a student with covariates \texttt{ownPC}=1,
\texttt{age}=18 and \texttt{hours}=2.
To produce an ordering of the gaming platforms, we can sample from the joint distribution of the 6 utilities and compute the most probable ordering. We have reported the predicted ordering and the observed one in the first and, respectively, second row of the following table:

\begin{center}
{\rowcolors{2}{gray!25}{white}
\small
\begin{tabular}{lrrrrrr}
\rowcolor{gray!50}
\toprule
{} &  Xbox &  PlayStation &  PSPortable &  GameCube &  GameBoy &   PC \\
\midrule
\text{predicted} &  2 &  3 &  4 &  6 &  5 &  1 \\
\text{observed} &  1 &  2 &  3 &  4 &  6 &  5 \\
\bottomrule
\end{tabular}}
\end{center}

Kendall-$\tau$ is a measure of correspondence between two rankings. It is defined as
$$
\tau=\frac{n_c-n_d}{n_0}
$$
where $n_0=n(n-1)/2$ is the number of possible pairs for $n$ elements, $n_c$ is the number of concordant pairs, $n_d$ is the number of discordant pairs. It is a number between $[-1,1]$, where $1$ indicates complete agreement and $-1$ complete disagreement. We can scale the Kendall-$\tau$ as $\tau'=(\tau+1)/2$, which is a number in $[0,1]$. A random predictor achieves   $\tau'=0.5$. For the two orderings in the above table, we obtain  $\tau'=0.66$. 
Figure \ref{fig:gaming}-right shows the marginal posterior utilities for a student with covariates \texttt{ownPlayStation}=1, \texttt{ownPC}=1,
\texttt{age}=20 and \texttt{hours}=15. The predicted and true ordering are:
\begin{center}
{\rowcolors{2}{gray!25}{white}
\small
\begin{tabular}{lrrrrrr}
\rowcolor{gray!50}
\toprule
{} &  Xbox &  PlayStation &  PSPortable &  GameCube &  GameBoy &   PC \\
\midrule 
\text{predicted} &  3 &  2 &  4 &  5 &  6 &  1 \\
\text{observed} &  3 &  2 &  5 &  4 &  6 &  1 \\
\bottomrule
\end{tabular}}
\end{center}
which gives $\tau'=0.93$. The average $\tau'$ computed using 10-fold cross-validation is 69\%.

\section{Modelling skills and predicting tennis match results}
\label{sec:tennis}
In the previous section, we discussed an application of preference learning regarding gaming console preferences.  A well-known application of preference learning in online gaming is ranking players. This problem is crucial  for gaming platforms  because matching players with similar skills makes the game more entertaining.  
 A famous method known as TrueSkill \citep{herbrich2006trueskill}, introduced by Microsoft, was designed to rank players for the Xbox  online gaming system. The basic idea of the algorithm is to assume that each player $i$ has a latent or true underlying skill level $u_i$. These skill levels can evolve over time according to a random walk $ u_i(t) \sim N(u_{i}(t-1), \gamma^2) $. 
 In any given game, the performance 
of player $i$, denoted as $q_i$, is defined as
$p(q_i|u_i(t))=N(q_i;u_i(t),\beta^2)$ and the probability that player $i$ wins against player $j$ is determined as
\begin{align}
\nonumber
    p(\text{i wins}|u_i(t),u_j(t)) &= \int I_{\{q_i>q_j\}}N(q_i;u_i(t),\beta^2)N(q_i;u_i(t),\beta^2)dq_idq_j\\
    &=\Phi\left(\frac{u_i(t)-u_j(t)}{\sqrt{2}\beta}\right),
\end{align}
which is the derivation discussed in Section \ref{sec:gaussiannoise}. TrueSkill  also works  for teams and deﬁnes the performance of a team to be the sum of the performance of its constituent players.
In the case of matches between two players, assuming independent matches, Model \ref{model:pairwiselabel} can be used to generalise TrueSkill.\footnote{In case of dependent matches due to common noise, we should use Model \ref{model:thurston}.} In this case, each player is represented as a label, and the features ${\bf x}_i$ include the time of the match and other relevant features of the match. 

To demonstrate this application of preference learning, we will focus on learning the skills of tennis player and predicting tennis match results. Specifically, we will use ATP tennis matches from 2022 and up to June 2024 to train Model \ref{model:pairwiselabel} and  predict the  outcomes of the 1st round of the Australian  and French Open 2024.\footnote{This is done in a rollout manner, meaning that we use all tennis results up to the starting date of the corresponding Grand Slam tournament to train the model. For example, for the Australian Open, we train the model on all  tennis results before January 14th, 2024.} The dataset\footnote{Source is \url{http://www.tennis-data.co.uk/alldata.php}.} includes approximately 4400  ATP tournament match results. In particular, Table \ref{tab:tennisdata} shows an entry in the dataset, including the name of \textit{Tournament}, the   \textit{Date} of  match, the \textit{Series} (such as Grand Slam, Masters, International or International Gold), the type of \textit{Court} (outdoors or indoors)
the type of \textit{Surface} (clay, hard, carpet or grass)
the \textit{Round} of match, the maximum number of sets playable in match \textit{Best of}, the match \textit{Winner}, the match \textit{Loser}, the ATP rank of the winner \textit{WRank}, the ATP rank of the loser \textit{LRank}, the \textit{Comment}  on the match (Completed, won through retirement of loser, or via Walkover), the average odds of match winner \textit{AvgW}
and the average odds of match loser  \textit{AvgL}.
We can use \textit{AvgW} and \textit{AvgL} as predictor for the match result: the bookmakers' predictions are correct whenever \textit{AvgW}$<$\textit{AvgL}.

\begin{table}
    \centering
    {\tiny
    \begin{tabular}{ll}
\toprule
\textbf{Tournament} & Adelaide International 1 \\
\textbf{Date} & 2023-01-02 \\
\textbf{Series} & ATP250 \\
\textbf{Court} & Outdoor \\
\textbf{Surface} & Hard \\
\textbf{Round} & 1st Round \\
\textbf{Best of} & 3 \\
\textbf{Winner} & Kecmanovic M. \\
\textbf{Loser} & O Connell C. \\
\textbf{WRank} & 29 \\
\textbf{LRank} & 78 \\
\textbf{Comment} & Completed \\
\textbf{AvgW} & 1.58 \\
\textbf{AvgL} & 2.36 \\
\bottomrule
\end{tabular}}
\caption{Dataset entry for one of the 1st Round match for ATP  Adelaide International tournament.}
\label{tab:tennisdata}
\end{table}
We consider  \textit{Date}, the ATP point of the  \textit{Series},  \textit{Court}, \textit{Surface} and \textit{Best of} as features. Therefore, the probability of the match result in  Table \ref{tab:tennisdata} is
$$
P(winner=\text{Kecmanovic})=\Phi(u_{\text{Kecmanovic}}({\bf x})-u_{\text{OConnell}}({\bf x})),
$$
with ${\bf x}=[2023-01-02,250,Outdoor,Hard, 3]$. Note that, in the code, we perform standard preprocessing of the features, such as scaling and one-hot encoding, not shown here.

We trained the model only on matches involving players who played at least one match in 2024, and we removed from the test set all matches involving players with an ATP rank lower than 200. This approach helps to exclude players for whom we have only a few available matches or players coming off a break due to injuries.

We used Model \ref{model:pairwiselabel} with an additive kernel, where one component has time as input and the other component uses the remaining covariates as inputs. Both kernels are Mat\'ern with smoothness parameter $\nu=3/2$ \citep[see chap.~4,][]{rasmussen2006gaussian}. We keep track of the different behaviour of the players in different conditions (e.g. surfaces, tournaments) by using a multiplicative index kernel defined over the ID of the player. Given $n_{players}$, the kernel is defined as $k(i,j) = (BB^T + diag(\mathbf{v}))_{i,j}$, where $B$ is a rank 2 matrix and $\mathbf{v}$ is a non-negative vector of size $n_{players}$. Model~\ref{model:pairwiselabel} is implemented in GPyTorch \citep{Gardner_2018_gpytorch} and trained with a variational sparse inducing point approximation and stochastic gradient descent on mini-batches. Note that the mini-batch size here is the number of pairwise preferences considered at each gradient step.

Table \ref{tab:australian} reports the prediction for the 1st Round matches in the Australian Open. The column \textit{Wprob} (respectively \textit{LProb}) represents the average number of scenarios (out of 1000 posterior samples) where our model gives the winner a higher chance of winning (or losing, respectively). For instance, in the first match between Arnaldi and Walton, Model \ref{model:pairwiselabel} predicts  that Arnaldi will win in 69\% of the sampled scenarios. Arnaldi is also the favourite according to the bookmakers AvgW=1.21 $<<$ AvgL=4.48.
The blue-rows highlight the matches where both the bookmakers and Model \ref{model:pairwiselabel} returned wrong predictions, the red-rows are the matches wrongly predicted only by the bookmakers and the orange-rows are the matches wrongly predicted   only by Model \ref{model:pairwiselabel}. Overall, bookmakers and Model \ref{model:pairwiselabel} have the same accuracy, ~71.7\%.
It is interesting to notice how Model \ref{model:pairwiselabel} makes predictions which are against the difference in the ATP rank such as for instance in the match between Monfils (ATP rank 76) and Hanfmann (ATP rank 52) showing that the learned player ``skill'' does not correspond to the ATP rank.

Similar comments hold for the French Open in Table \ref{tab:french}.  Overall, bookmakers and Model \ref{model:pairwiselabel} have the same accuracy:~72.4\%.

\begin{table}
\centering
{\tiny
\begin{tabular}{ll||rrrr||rr}
\textbf{Winner} & \textbf{Loser} & \textbf{WRank} & \textbf{LRank} & \textbf{AvgW} & \textbf{AvgL} & \textbf{Wprob} & \textbf{Lprob} \\
\hline
  Arnaldi M. &  Walton A. &  41 &  174 &  1.21 &  4.48 &  0.69 &  0.31 \\
\color{blue} Munar J. & \color{blue} Shevchenko A. & \color{blue} 82 & \color{blue} 48 & \color{blue} 3.00 & \color{blue} 1.40 & \color{blue} 0.39 & \color{blue} 0.61 \\
 Sinner J. &  Van De Zands. B. &  4 &  59 &  1.05 &  10.80 &  0.65 &  0.35 \\
 Kotov P. &  Rinderknech A. &  65 &  94 &  1.67 &  2.23 &  0.51 &  0.49 \\
 Rublev A. &  Seyboth Wild T. &  5 &  77 &  1.02 &  16.12 &  0.74 &  0.26 \\
 Machac T. &  Mochizuki S. &  75 &  136 &  1.27 &  3.82 &  0.65 &  0.35 \\
\color{orange} O Connell C. & \color{orange} Garin C. & \color{orange} 68 & \color{orange} 86 & \color{orange} 1.41 & \color{orange} 2.95 & \color{orange} 0.46 & \color{orange} 0.54 \\
\color{orange} De Jong J. & \color{orange} Cachin P. & \color{orange} 161 & \color{orange} 74 & \color{orange} 1.37 & \color{orange} 3.12 & \color{orange} 0.40 & \color{orange} 0.60 \\
 Fritz T. &  Diaz Acosta F. &  12 &  90 &  1.03 &  14.76 &  0.62 &  0.38 \\
\color{blue} Galan D.E. & \color{blue} Kubler J. & \color{blue} 87 & \color{blue} 113 & \color{blue} 2.46 & \color{blue} 1.55 & \color{blue} 0.40 & \color{blue} 0.60 \\
\color{red} Halys Q. & \color{red} Harris L. & \color{red} 110 & \color{red} 167 & \color{red} 3.34 & \color{red} 1.33 & \color{red} 0.54 & \color{red} 0.46 \\
 Djokovic N. &  Prizmic D. &  1 &  187 &  1.01 &  21.89 &  0.76 &  0.24 \\
 Tiafoe F. &  Coric B. &  17 &  40 &  1.54 &  2.48 &  0.66 &  0.34 \\
 Korda S. &  Kopriva V. &  26 &  132 &  1.08 &  8.50 &  0.55 &  0.45 \\
 Ruusuvuori E. &  Kypson P. &  53 &  183 &  1.11 &  6.76 &  0.60 &  0.40 \\
 Musetti L. &  Bonzi B. &  28 &  106 &  1.35 &  3.25 &  0.55 &  0.45 \\
\color{blue} Cobolli F. & \color{blue} Jarry N. & \color{blue} 100 & \color{blue} 18 & \color{blue} 4.24 & \color{blue} 1.23 & \color{blue} 0.41 & \color{blue} 0.59 \\
\color{red} Gaston H. & \color{red} Carballes Baena R. & \color{red} 97 & \color{red} 63 & \color{red} 2.29 & \color{red} 1.63 & \color{red} 0.52 & \color{red} 0.48 \\
 Shelton B. &  Bautista Agut R. &  16 &  72 &  1.42 &  2.90 &  0.68 &  0.32 \\
 Davidovich Fokina A. &  Lestienne C. &  24 &  99 &  1.19 &  4.82 &  0.73 &  0.27 \\
 Popyrin A. &  Polmans M. &  43 &  154 &  1.17 &  5.03 &  0.50 &  0.50 \\
 Mannarino A. &  Wawrinka S. &  19 &  56 &  1.43 &  2.84 &  0.55 &  0.45 \\
\color{orange} Van Assche L. & \color{orange} Duckworth J. & \color{orange} 79 & \color{orange} 95 & \color{orange} 1.82 & \color{orange} 2.00 & \color{orange} 0.44 & \color{orange} 0.56 \\
 Tsitsipas S. &  Bergs Z. &  7 &  129 &  1.09 &  7.42 &  0.72 &  0.28 \\
\color{orange} Mensik J. & \color{orange} Shapovalov D. & \color{orange} 142 & \color{orange} 114 & \color{orange} 1.67 & \color{orange} 2.22 & \color{orange} 0.45 & \color{orange} 0.55 \\
 Khachanov K. &  Altmaier D. &  15 &  50 &  1.21 &  4.48 &  0.67 &  0.33 \\
 Thompson J. &  Vukic A. &  47 &  64 &  1.35 &  3.24 &  0.54 &  0.46 \\
\color{blue} Kovacevic A. & \color{blue} Tabilo A. & \color{blue} 101 & \color{blue} 49 & \color{blue} 2.32 & \color{blue} 1.62 & \color{blue} 0.30 & \color{blue} 0.70 \\
 Zhang Zh. &  Coria F. &  54 &  84 &  1.06 &  9.65 &  0.63 &  0.37 \\
\color{red} Eubanks C. & \color{red} Daniel T. & \color{red} 35 & \color{red} 58 & \color{red} 2.37 & \color{red} 1.59 & \color{red} 0.58 & \color{red} 0.42 \\
\color{blue} Etcheverry T. & \color{blue} Murray A. & \color{blue} 32 & \color{blue} 44 & \color{blue} 2.30 & \color{blue} 1.62 & \color{blue} 0.41 & \color{blue} 0.59 \\
 Monfils G. &  Hanfmann Y. &  76 &  52 &  1.51 &  2.59 &  0.67 &  0.33 \\
 Humbert U. &  Goffin D. &  20 &  112 &  1.39 &  3.00 &  0.55 &  0.45 \\
\color{blue} Borges N. & \color{blue} Marterer M. & \color{blue} 69 & \color{blue} 96 & \color{blue} 2.12 & \color{blue} 1.74 & \color{blue} 0.34 & \color{blue} 0.66 \\
\color{red} Grenier H. & \color{red} Muller A. & \color{red} 178 & \color{red} 73 & \color{red} 3.10 & \color{red} 1.38 & \color{red} 0.58 & \color{red} 0.42 \\
\color{orange} Kecmanovic M. & \color{orange} Watanuki Y. & \color{orange} 60 & \color{orange} 102 & \color{orange} 1.31 & \color{orange} 3.49 & \color{orange} 0.39 & \color{orange} 0.61 \\
 Auger-Aliassime F. &  Thiem D. &  30 &  92 &  1.45 &  2.80 &  0.54 &  0.46 \\
\color{orange} Struff J.L. & \color{orange} Hijikata R. & \color{orange} 25 & \color{orange} 71 & \color{orange} 1.61 & \color{orange} 2.33 & \color{orange} 0.34 & \color{orange} 0.66 \\
 Lehecka J. &  Zapata Miralles B. &  23 &  78 &  1.06 &  9.98 &  0.65 &  0.35 \\
 Norrie C. &  Varillas J. P. &  22 &  81 &  1.13 &  6.18 &  0.64 &  0.36 \\
\color{blue} Cazaux A. & \color{blue} Djere L. & \color{blue} 122 & \color{blue} 33 & \color{blue} 2.04 & \color{blue} 1.79 & \color{blue} 0.35 & \color{blue} 0.65 \\
\color{red} Griekspoor T. & \color{red} Safiullin R. & \color{red} 31 & \color{red} 36 & \color{red} 2.57 & \color{red} 1.52 & \color{red} 0.56 & \color{red} 0.44 \\
\color{red} Zeppieri G. & \color{red} Lajovic D. & \color{red} 133 & \color{red} 51 & \color{red} 2.00 & \color{red} 1.82 & \color{red} 0.60 & \color{red} 0.40 \\
 Ruud C. &  Ramos-Vinolas A. &  11 &  85 &  1.02 &  15.52 &  0.72 &  0.28 \\
 Dimitrov G. &  Fucsovics M. &  13 &  70 &  1.16 &  5.36 &  0.59 &  0.41 \\
 Draper J. &  Giron M. &  55 &  66 &  1.24 &  4.06 &  0.66 &  0.34 \\
 Rune H. &  Nishioka Y. &  8 &  61 &  1.12 &  6.39 &  0.59 &  0.41 \\
 Shang J. &  Mcdonald M. &  140 &  42 &  1.71 &  2.15 &  0.51 &  0.49 \\
 Paul T. &  Barrere G. &  14 &  83 &  1.12 &  6.15 &  0.70 &  0.30 \\
\color{blue} Kokkinakis T. & \color{blue} Ofner S. & \color{blue} 80 & \color{blue} 37 & \color{blue} 1.99 & \color{blue} 1.82 & \color{blue} 0.46 & \color{blue} 0.54 \\
 Sonego L. &  Evans D. &  46 &  38 &  1.80 &  2.01 &  0.52 &  0.48 \\
 Zverev A. &  Koepfer D. &  6 &  62 &  1.06 &  9.55 &  0.71 &  0.29 \\
 Alcaraz C. &  Gasquet R. &  2 &  131 &  1.02 &  17.45 &  0.74 &  0.26 \\
\end{tabular}}
\caption{Predictions for Australian Open 1st Round: the blue-rows highlight the matches where both the bookmakers and Model \ref{model:pairwiselabel} returned wrong predictions; red-rows are the matches wrongly predicted only by the bookmakers; the orange-rows are the matches wrongly predicted  our only by Model \ref{model:pairwiselabel}.}
\label{tab:australian}
\end{table}

\begin{table}
\centering
{\tiny
\begin{tabular}{ll||rrrr||rr}
\textbf{Winner} & \textbf{Loser} & \textbf{WRank} & \textbf{LRank} & \textbf{AvgW} & \textbf{AvgL} & \textbf{Wprob} & \textbf{Lprob} \\
\hline
Nakashima B. &  Moreno De A. N. &  84 &  130 &  1.54 &  2.49 &  0.59 &  0.41 \\
 Rublev A. &  Daniel T. &  6 &  80 &  1.05 &  11.40 &  0.66 &  0.34 \\
\color{orange} Martinez P. & \color{orange} Tirante T.A. & \color{orange} 48 & \color{orange} 109 & \color{orange} 1.30 & \color{orange} 3.58 & \color{orange} 0.41 & \color{orange} 0.59 \\
\color{blue} Sonego L. & \color{blue} Humbert U. & \color{blue} 49 & \color{blue} 16 & \color{blue} 2.59 & \color{blue} 1.51 & \color{blue} 0.37 & \color{blue} 0.63 \\
 Zhang Zh. &  Vukic A. &  44 &  89 &  1.35 &  3.22 &  0.60 &  0.40 \\
\color{blue} De Jong J. & \color{blue} Draper J. & \color{blue} 176 & \color{blue} 39 & \color{blue} 4.12 & \color{blue} 1.24 & \color{blue} 0.39 & \color{blue} 0.61 \\
 Hurkacz H. &  Mochizuki S. &  8 &  162 &  1.04 &  12.27 &  0.74 &  0.26 \\
 Dimitrov G. &  Kovacevic A. &  10 &  87 &  1.09 &  7.50 &  0.64 &  0.36 \\
\color{blue} Marterer M. & \color{blue} Thompson J. & \color{blue} 101 & \color{blue} 36 & \color{blue} 2.42 & \color{blue} 1.57 & \color{blue} 0.41 & \color{blue} 0.59 \\
 Alcaraz C. &  Wolf J.J. &  3 &  107 &  1.01 &  23.24 &  0.71 &  0.29 \\
\color{red} Gasquet R. & \color{red} Coric B. & \color{red} 124 & \color{red} 73 & \color{red} 3.10 & \color{red} 1.37 & \color{red} 0.52 & \color{red} 0.48 \\
\color{orange} Muller A. & \color{orange} Nardi L. & \color{orange} 90 & \color{orange} 72 & \color{orange} 1.56 & \color{orange} 2.44 & \color{orange} 0.49 & \color{orange} 0.51 \\
\color{blue} Bergs Z. & \color{blue} Tabilo A. & \color{blue} 104 & \color{blue} 24 & \color{blue} 3.12 & \color{blue} 1.37 & \color{blue} 0.26 & \color{blue} 0.74 \\
 Marozsan F. &  Kukushkin M. &  43 &  136 &  1.26 &  3.93 &  0.57 &  0.42 \\
 Korda S. &  Mayot H. &  28 &  122 &  1.19 &  4.71 &  0.62 &  0.38 \\
\color{orange} Altmaier D. & \color{orange} Djere L. & \color{orange} 83 & \color{orange} 52 & \color{orange} 1.56 & \color{orange} 2.44 & \color{orange} 0.49 & \color{orange} 0.51 \\
\color{red} Moutet C. & \color{red} Jarry N. & \color{red} 79 & \color{red} 19 & \color{red} 2.81 & \color{red} 1.44 & \color{red} 0.51 & \color{red} 0.49 \\
 Ofner S. &  Atmane T. &  45 &  121 &  1.54 &  2.50 &  0.62 &  0.38 \\
\color{orange} Wawrinka S. & \color{orange} Murray A. & \color{orange} 98 & \color{orange} 75 & \color{orange} 1.55 & \color{orange} 2.46 & \color{orange} 0.43 & \color{orange} 0.57 \\
 Shevchenko A. &  Karatsev A. &  59 &  82 &  1.58 &  2.41 &  0.57 &  0.43 \\
 Sinner J. &  Eubanks C. &  2 &  46 &  1.02 &  17.10 &  0.74 &  0.26 \\
 Shelton B. &  Gaston H. &  15 &  88 &  1.40 &  2.96 &  0.64 &  0.36 \\
 Auger-Aliassime F. &  Nishioka Y. &  21 &  70 &  1.19 &  4.87 &  0.54 &  0.46 \\
 Baez S. &  Heide G. &  20 &  174 &  1.17 &  5.18 &  0.63 &  0.37 \\
\color{red} Arnaldi M. & \color{red} Fils A. & \color{red} 35 & \color{red} 38 & \color{red} 2.07 & \color{red} 1.77 & \color{red} 0.69 & \color{red} 0.31 \\
 Tsitsipas S. &  Fucsovics M. &  9 &  54 &  1.11 &  6.77 &  0.56 &  0.44 \\
\color{blue} Kovalik J. & \color{blue} Giron M. & \color{blue} 145 & \color{blue} 50 & \color{blue} 2.09 & \color{blue} 1.77 & \color{blue} 0.37 & \color{blue} 0.63 \\
 Khachanov K. &  Nagal S. &  18 &  95 &  1.13 &  6.11 &  0.70 &  0.30 \\
\color{blue} Kotov P. & \color{blue} Norrie C. & \color{blue} 56 & \color{blue} 33 & \color{blue} 2.85 & \color{blue} 1.43 & \color{blue} 0.45 & \color{blue} 0.55 \\
 Cerundolo F. &  Hanfmann Y. &  27 &  85 &  1.32 &  3.47 &  0.54 &  0.46 \\
 Tiafoe F. &  Bellucci M. &  26 &  173 &  1.31 &  3.55 &  0.50 &  0.50 \\
 Musetti L. &  Galan D.E. &  30 &  106 &  1.15 &  5.65 &  0.59 &  0.41 \\
 Paul T. &  Cachin P. &  14 &  108 &  1.10 &  7.14 &  0.69 &  0.31 \\
 Shapovalov D. &  Van Assche L. &  118 &  103 &  1.61 &  2.34 &  0.56 &  0.44 \\
\color{orange} Kecmanovic M. & \color{orange} Monteiro T. & \color{orange} 57 & \color{orange} 86 & \color{orange} 1.84 & \color{orange} 1.97 & \color{orange} 0.43 & \color{orange} 0.57 \\
 Lajovic D. &  Safiullin R. &  61 &  42 &  1.64 &  2.29 &  0.61 &  0.39 \\
\color{red} Monfils G. & \color{red} Seyboth Wild T. & \color{red} 37 & \color{red} 58 & \color{red} 2.27 & \color{red} 1.65 & \color{red} 0.60 & \color{red} 0.40 \\
\color{red} Fognini F. & \color{red} Van De Zands. B. & \color{red} 93 & \color{red} 102 & \color{red} 2.65 & \color{red} 1.49 & \color{red} 0.59 & \color{red} 0.41 \\
 Medvedev D. &  Koepfer D. &  5 &  65 &  1.20 &  4.66 &  0.59 &  0.41 \\
\color{orange} Etcheverry T. & \color{orange} Cazaux A. & \color{orange} 29 & \color{orange} 77 & \color{orange} 1.20 & \color{orange} 4.54 & \color{orange} 0.48 & \color{orange} 0.52 \\
 Ruud C. &  Meligeni Alves F. &  7 &  137 &  1.05 &  10.54 &  0.70 &  0.30 \\
 Machac T. &  Borges N. &  34 &  47 &  1.60 &  2.38 &  0.60 &  0.40 \\
 Davidovich Fokina A. &  Vacherot V. &  32 &  116 &  1.22 &  4.27 &  0.55 &  0.45 \\
\color{orange} Munar J. & \color{orange} Bautista Agut R. & \color{orange} 64 & \color{orange} 81 & \color{orange} 1.69 & \color{orange} 2.20 & \color{orange} 0.37 & \color{orange} 0.63 \\
 De Minaur A. &  Michelsen A. &  11 &  60 &  1.10 &  7.13 &  0.53 &  0.47 \\
 Rinderknech A. &  Walton A. &  69 &  96 &  1.17 &  5.13 &  0.68 &  0.32 \\
 Fritz T. &  Coria F. &  12 &  71 &  1.13 &  6.09 &  0.70 &  0.30 \\
\color{blue} Goffin D. & \color{blue} Mpetshi G. & \color{blue} 115 & \color{blue} 66 & \color{blue} 2.24 & \color{blue} 1.66 & \color{blue} 0.44 & \color{blue} 0.56 \\
\color{red} Cobolli F. & \color{red} Medjedovic H. & \color{red} 53 & \color{red} 135 & \color{red} 2.53 & \color{red} 1.53 & \color{red} 0.54 & \color{red} 0.46 \\
 Struff J.L. &  Burruchaga R. &  41 &  144 &  1.14 &  5.99 &  0.51 &  0.49 \\
 Rune H. &  Evans D. &  13 &  62 &  1.09 &  7.84 &  0.60 &  0.40 \\
 Zeppieri G. &  Mannarino A. &  148 &  22 &  1.23 &  4.28 &  0.56 &  0.44 \\
 Darderi L. &  Hijikata R. &  40 &  78 &  1.08 &  8.14 &  0.52 &  0.48 \\
\color{red} Kokkinakis T. & \color{red} Popyrin A. & \color{red} 100 & \color{red} 51 & \color{red} 1.98 & \color{red} 1.83 & \color{red} 0.54 & \color{red} 0.46 \\
 Carballes Baena R. &  Lestienne C. &  63 &  91 &  1.15 &  5.44 &  0.58 &  0.42 \\
 Griekspoor T. &  Mcdonald M. &  25 &  74 &  1.12 &  6.58 &  0.66 &  0.34 \\
 Djokovic N. &  Herbert P.H. &  1 &  142 &  1.02 &  17.79 &  0.69 &  0.31 \\
\color{orange} Bublik A. & \color{orange} Barrere G. & \color{orange} 17 & \color{orange} 112 & \color{orange} 1.90 & \color{orange} 1.92 & \color{orange} 0.49 & \color{orange} 0.51 \\
\end{tabular}
}
\caption{Predictions for French Open 1st Round: the blue-rows highlight the matches where both the bookmakers and Model \ref{model:pairwiselabel} returned wrong predictions; red-rows are the matches wrongly predicted only by the bookmakers; the orange-rows are the matches wrongly predicted  our only by Model \ref{model:pairwiselabel}.}
\label{tab:french}
\end{table}

\section{Choose your preferred ellipse}

To illustrate an application for learning choice functions, we designed a simple experiment based on visual choices between geometric shapes. Although academic in nature, this setup mirrors real-world scenarios, such as online shopping, where users choose between items (e.g., clothes or shoes) based on visual features. In our case, the items are ellipses that vary in shape and orientation, allowing us to explore how individual choices can be learned and interpreted when the ground truth is known.

\begin{figure}
    \centering
    \includegraphics[width=\textwidth]{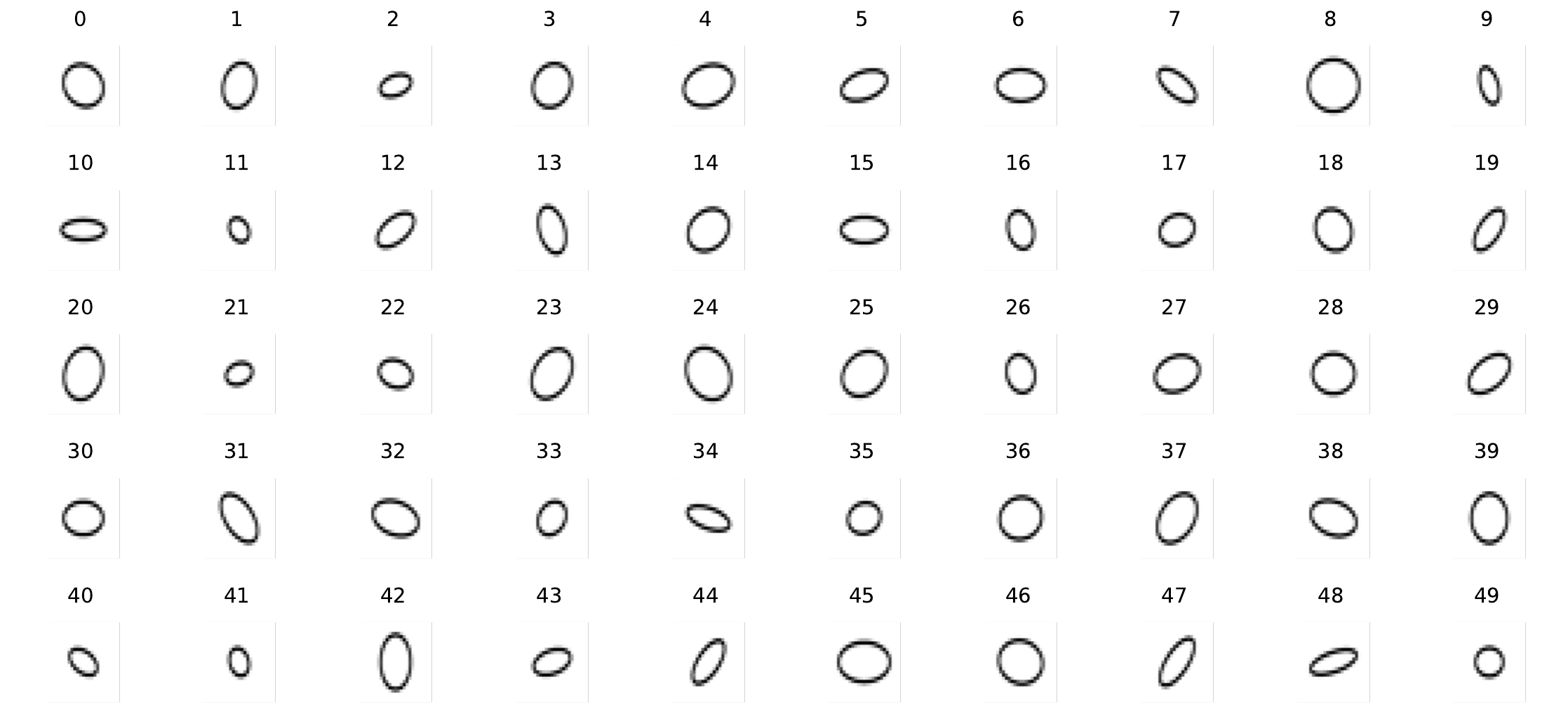}
    \caption{Thumbnail images of the 50 ellipses.}
    \label{fig:allellipses}
\end{figure}

\begin{figure}
    \centering
    \includegraphics[width=0.6\textwidth]{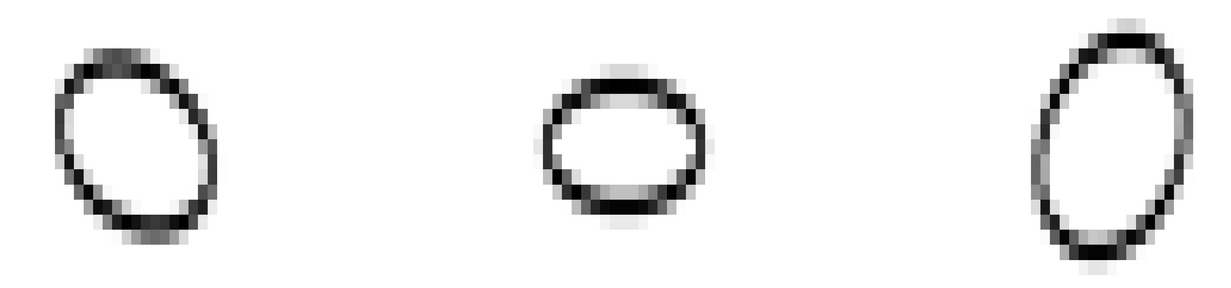}
    \caption{Ellipses shown to the student (ID=32, 15, 25 from left to right). The student selected 15.}
    \label{fig:showntostud}
\end{figure}

In particular, we implemented a gallery demonstration on a computer, showcasing three distinct images of ellipses to a student. These images were randomly selected from the 50 images shown in Figure \ref{fig:allellipses}. As it can be seen, these ellipses differ in  dimension and orientation relative to the coordinate axes. The task assigned to the student involved selecting ellipses based on two criteria: (i) small eccentricity (being close to a circular shape); (ii) alignment of their axes with the Cartesian axes. It was emphasised to the student that, in case of incomparability due to conflicting criteria, multiple choices were acceptable. Figure \ref{fig:showntostud} shows a screenshot of the image shown to the student and the student's choice in the caption. Note that, we decreased the resolution of the image to make this task harder.  The student repeated this task 160 times generating a dataset including 160 choices, for example $A_1=\{6,36,47\},C(A_1)=\{6\},~A_2=\{37,42,47\},C(A_2)=\{37,42\}$.
 
\begin{figure}
     \centering
     \begin{subfigure}[b]{0.3\textwidth}
         \centering
    \includegraphics[width=4.0cm]{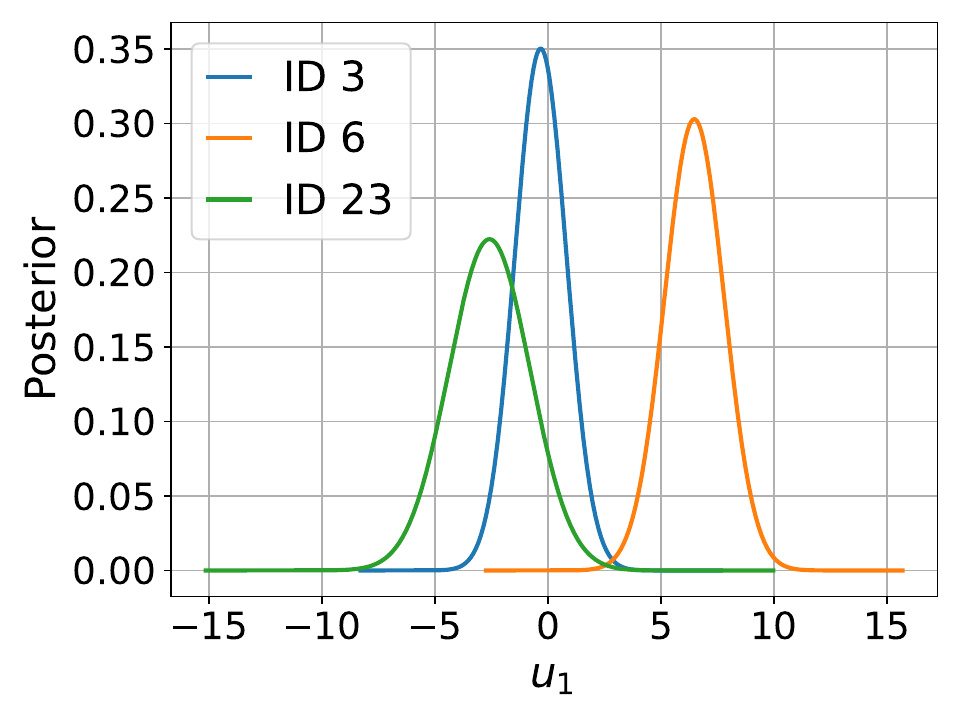}
    \includegraphics[width=4.0cm]{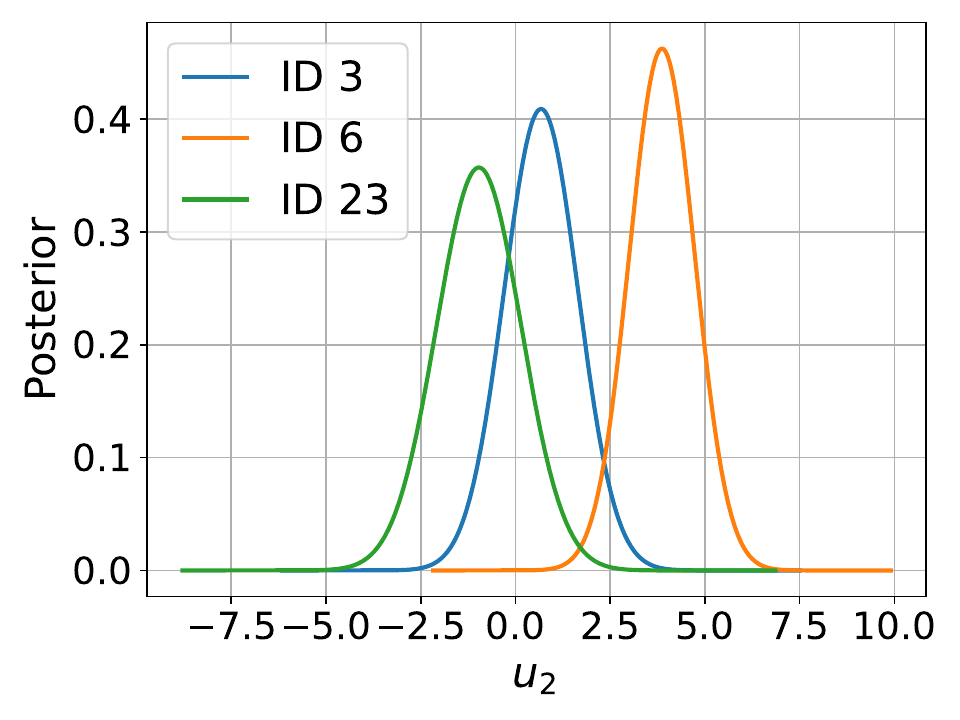}
    \caption{}
         \label{fig:outellips1}
     \end{subfigure}\hfill     
     \begin{subfigure}[b]{0.3\textwidth}
         \centering
    \includegraphics[width=4.0cm]{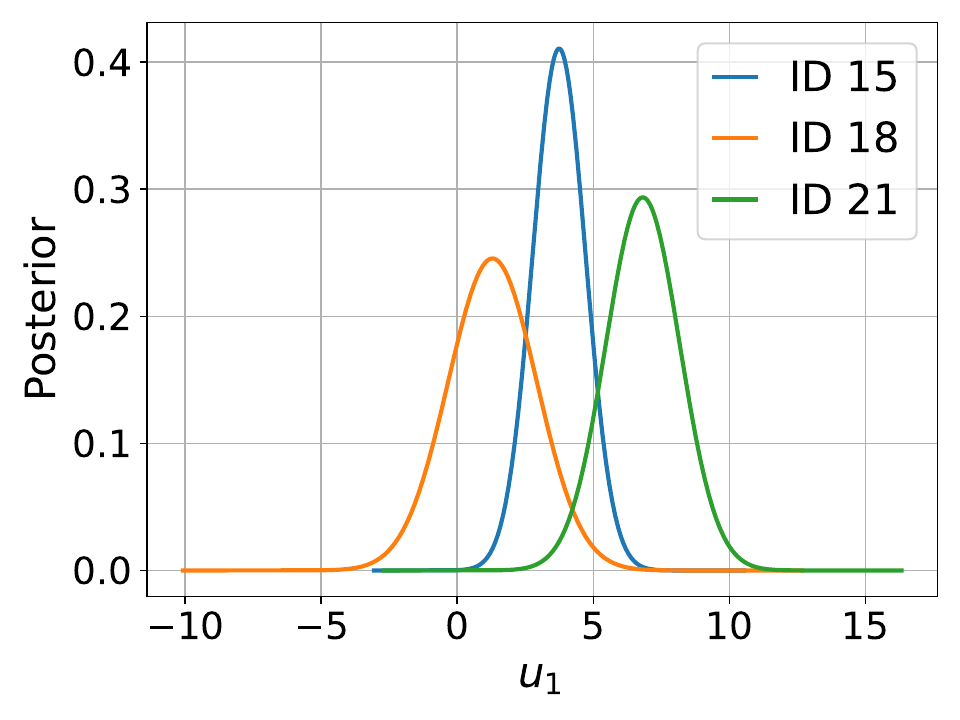}
    \includegraphics[width=4.0cm]{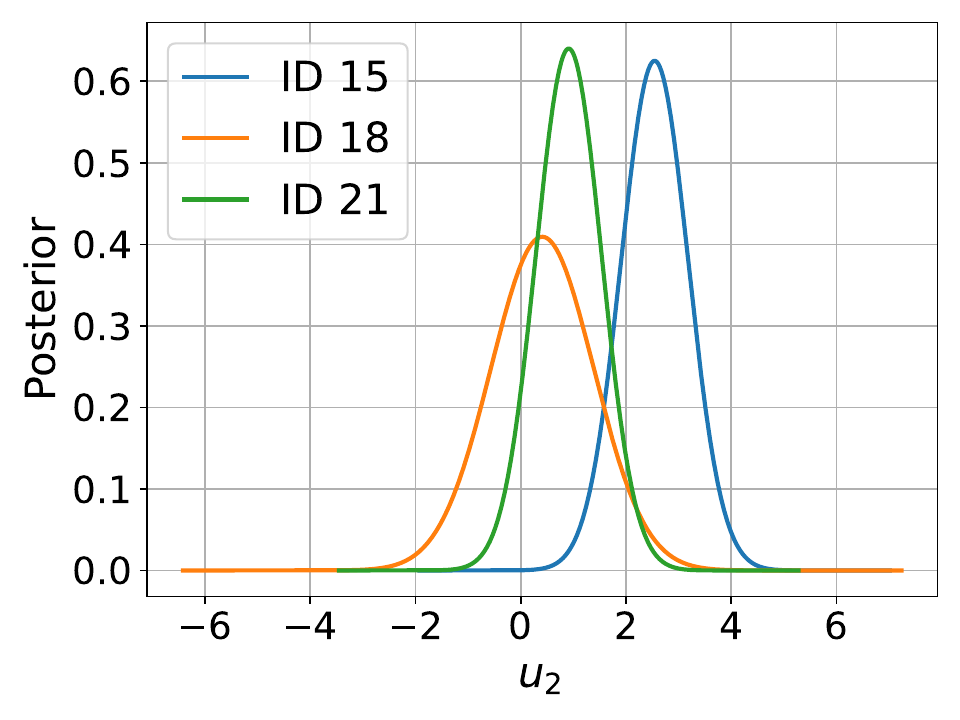}
    \caption{}
         \label{fig:outellips2}
     \end{subfigure}\hfill     
     \begin{subfigure}[b]{0.3\textwidth}
         \centering
    \includegraphics[width=4.0cm]{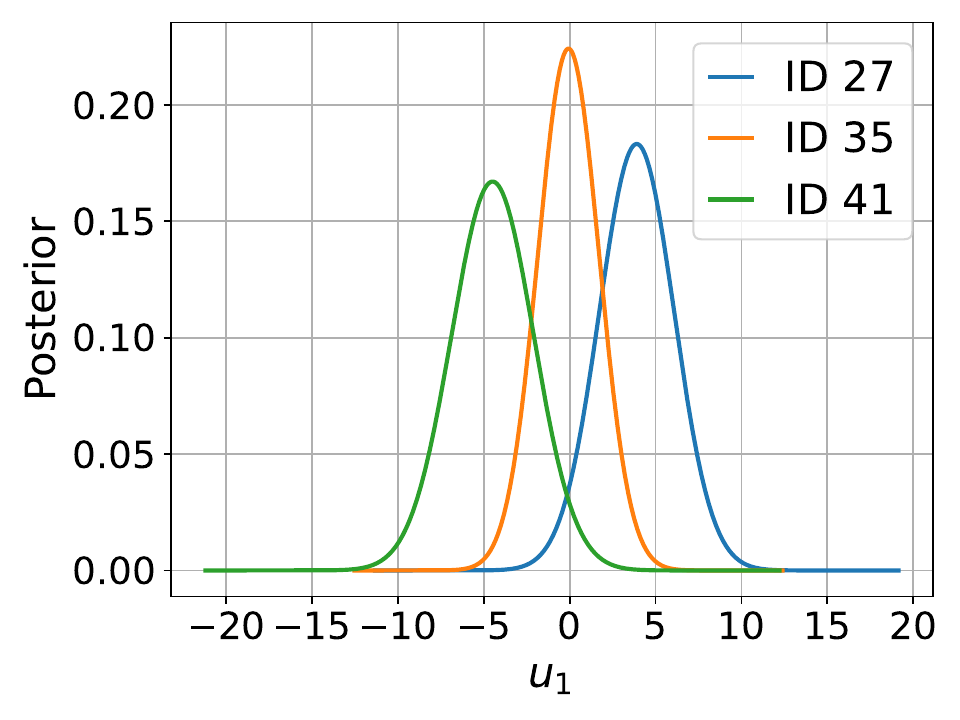}
    \includegraphics[width=4.0cm]{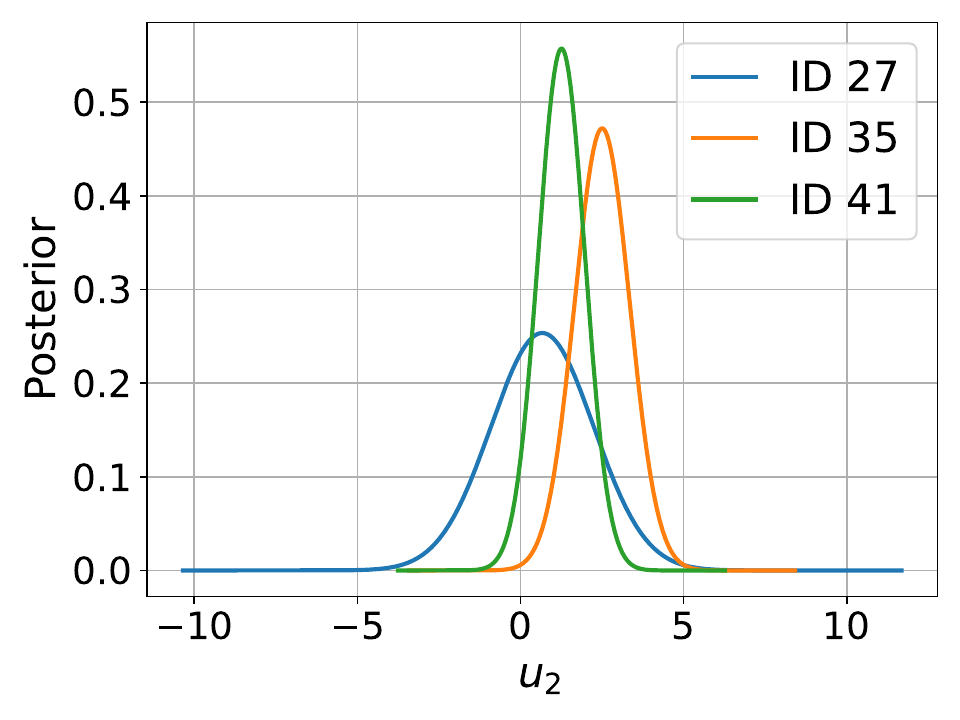}
    \caption{}
         \label{fig:outellips3}
     \end{subfigure}
        \caption{Predictive posterior for the two latent utilities (top and bottom plot) for the ellipses with (ID=3, 6, 23), (ID=15, 18, 21) and (ID=27, 35, 41).}
        \label{fig:three graphs}
\end{figure}

We use this dataset ($50$ images $25\times 25$ pixels and $160$ choices) to learn the latent utilities through Model \ref{model:choice} with pseudo-rational likelihood (since we do not know if the student used Pareto-dominance). The dataset was partitioned into 90\% for training and 10\% for testing using 10-fold cross-validation, and we computed the accuracy in predicting the student's choices. More precisely, given a choice set $A$, we computed the most probable choice $\hat{C}(A),\hat{R}(A)$ using the posterior samples of the utility computed according to Model  \ref{model:choice}. Then, given the observed choice $C(A),R(A)$, the accuracy was computed as
$$
acc= \frac{1}{|A|} \left(\sum_{o \in C(A)} I_{\hat{C}(A)}(o)+\sum_{v \in R(A)} I_{\hat{R}(A)}(v)\right).
$$
The model achieved 84\% accuracy.
Figure \ref{fig:outellips1} depicts the learned utilities for the ellipses with ID $3, 6, 23$, Figure \ref{fig:allellipses}. As expected, ellipse ID=6 dominates the other ellipses (ID=3 and 23) in both latent utilities. It is worth noting that we do not have explicit labels for the two learned utilities, and their mapping to specific criteria (small eccentricity and alignment to the Cartesian axes) remains unknown. This is analogous to the arbitrary labels in clustering.

Figure \ref{fig:outellips2} showcases the utilities for a case of incomparability. Ellipse ID=21 dominates ID=15 in terms of the first latent utility, whereas ID=15 dominates ID=21 in terms of the second latent utility. The  model arrives at a decision that matches the student's selection, who selected the ellipses $C(A)=\{15,21\}$. 

Mostly of the errors result from the ellipses having nearly identical eccentricities and similar aligning to  the axes, as shown in Figure \ref{fig:outellips3}. For $A=\{27,35,41\}$, the model predicts $\hat{C}(A)=\{27,35\}$, while the student selected $\hat{C}(A)=\{35\}$. In these cases, the student's choices are often wrong.  Therefore, the choice model appears to work quite well overall. 
{\color{red}
}

\section{Connection to recommender systems: movielens dataset}
\label{sec:movie}

In this section, we will explore the relationship between preference and choice learning within recommendation systems. Additionally, we will discuss how to perform preference and choice learning when dealing with millions of preference pair data.

A recommender system \citep{Aggarwal_2016} is an algorithm that, given user's feedback, provides one or more suggestions regarding items that might interest a specific user. Often, those systems are based on user-item interactions and use collaborative filtering methods to provide recommendations. In this section we outline how a preference model, such as Model~\ref{model:pairwiselabel}, 
 can be used as a recommender system.    

Let us consider a system with $d$ items $c_1, \ldots, c_d$ and $m$ users ${\bf x}^{(k)}$, $k=1, \ldots, m$. We can assume that we either have explicit interactions (e.g., user's rankings) or implicit interactions (e.g., browsing history). The recommendation problem consists in providing a suggestion for a new user ${\bf x}^*$ or a new (unobserved) suggestion $c_i$ to user ${\bf x}^{(k)}$ for some $k$. We can view this task either as the problem of learning the ranking of the $d$ labels for a user ${\bf x}$, or the problem of learning a user's preferences.   

{
\rowcolors{2}{gray!25}{white}
\begin{table}[h]
\centering
\footnotesize
(a)~~~\begin{tabular}{lrll}
\rowcolor{gray!50}
\toprule
 & age & sex & profession   \\
\midrule
\rowcolor{blue!15}
1 & 24 & M & technician \\
\rowcolor{orange!15}
2 & 53 & F & other  \\
3 & 23 & M & writer  \\
4 & 24 & M & technician  \\
5 & 33 & F & other  \\
$\cdots$ & $\cdots$ & $\cdots$ & $\cdots$  \\
\bottomrule
\end{tabular}\\
(b)~~~\begin{tabular}{lrcclll}
\rowcolor{gray!50}
\toprule
 & age & sex & technician & other & writer & movie \\
\midrule
\rowcolor{blue!10}
1 & -0.65 &  1. &  1. &   0.    &  0. & 0. \\
\rowcolor{orange!10}
2 & 1.90  &  0. &  0. &   1.    &  0. & 0. \\
$\cdots$ & $\cdots$ & $\cdots$ & $\cdots$ & $\cdots$ & $\cdots$ & $\cdots$ \\
\rowcolor{blue!25}
51 & -0.65 &  1. &  1. &   0.    &  0. & 7. \\
\rowcolor{orange!25}
52 & 1.90  &  0. &  0. &   1.    &  0. & 1. \\
$\cdots$ & $\cdots$ & $\cdots$ & $\cdots$ & $\cdots$ & $\cdots$ & $\cdots$ \\
\bottomrule
\end{tabular}
\caption{MovieLens example: (a) users' features; (b) tiled users' features with additional feature for pairwise preferences.}
\label{tab:movieUsers}
\end{table}
}

{
\rowcolors{2}{gray!25}{white}
\begin{table}[h]
\centering
\footnotesize
\begin{tabular}{cc}
\rowcolor{gray!50}
\toprule
 movie 1 & movie 2 \\
\midrule
\cellcolor{blue!10} 0 &  \cellcolor{blue!25} 350 \\
\cellcolor{orange!10} 1  &  \cellcolor{orange!25} 51 \\
 $\cdots$ & $\cdots$  \\
\bottomrule
\end{tabular}
\caption{MovieLens example, preference encoding:  movie 1 $\succ$ movie 2.}
\label{tab:movieUsersPref}
\end{table}
}

We can frame this task as  a preference learning task by rearranging the data in the following format. We assume that, for each user $k$, we have a vector of features ${\bf x}^{(k)}$ that identifies them. Moreover, we can create a sequence of pairwise preferences $c_i \succ_{{\bf x}^{(k)}} c_j$ for $i,j \in I_{(k)}$, where $I_{(k)}$ is the set of indices identifying the items that have an interaction with user $k$. Note that since $c_i \succ_{{\bf x}^{(k)}} c_j$ is a pairwise preference, this implies that the user must have chosen the item $c_i$ over $c_j$. This is available in the data either with an explicit interaction with both $c_i$ and $c_j$, i.e. the user $k$ ranked $c_i$ better than $c_j$, or with an implicit interaction, i.e. the user $k$ viewed item $c_i$ but not item $c_j$. We can further stack all preferences for all available users. Tables~\ref{tab:movieUsers} and~\ref{tab:movieUsersPref} show an example of how the data is formatted. We start from the users' features in Table~\ref{tab:movieUsers} (a) and with the data ``user 1 prefers movie 0 over movie 7'' (not shown). We first transform the user features with normalisation and one-hot encoding, first 5 columns of Table~\ref{tab:movieUsers} (b); then we stack all user features for all movies, as shown in Table~\ref{tab:movieUsers} (b). Note that Table~\ref{tab:movieUsers} (b) repeats the features of user 1, in blue, for as many rows as the number of movies. The last column in Table~\ref{tab:movieUsers} (b) allows us to encode the block diagonal structure of Model~\ref{model:pairwiselabel}. Finally Table~\ref{tab:movieUsersPref} shows the preference encoding: each preference is encoded with a unique movie user pair. For example ``user 1 prefers movie 0 over movie 7'' is encoded as item 0 $\succ$ item 350 ($7\times 50+0$) and ``user 2 prefers movie 0 over movie 1'' is encoded as item 1 $\succ$ 51 ($1\times 50+1$). This data structure now allows us to learn the user's behaviour by using Model~\ref{model:pairwiselabel}.

We show the effectiveness of this learning procedure on the MovieLens dataset \citep{Harper_Konstan_2015}. The original datasets consists of 100,000 ratings given by $943$ users to $1682$ movies. Each user has rated at least $20$ movies and demographic information is available for each user. We consider a subset of $650$ movies and $50$ users and we generate $2.68$ million pairwise preferences with the procedure outlined above.


Model~\ref{model:pairwiselabel} is implemented in GPyTorch \citep{Gardner_2018_gpytorch} and trained with a variational sparse inducing point approximation, $800$ inducing points and stochastic gradient descent on mini-batches of size $5096$. Note that the mini-batch size here is the number of pairwise preferences considered at each gradient step. We compare the performances of this method against Bayesian Personalized Ranking (BPR, \citet{Rendle_etal_BPR_2009}), a standard Bayesian learning procedure for recommender systems, based on a k-nearest neighbours algorithm. We evaluate the performance with classification metrics: for each new item in the test set an item is correctly classified if the suggested item was preferred by the user. Model~\ref{model:pairwiselabel} achieves an accuracy of $0.90$ and a ROC AUC of $0.97$ on the test set. BPR instead achieves an accuracy of $0.71$ and ROC AUC of $0.80$ when trained on the same subset of movies. Note that BPR can be regarded as a linear parametric equivalent of our preference based model. Moreover, our GP-based approach allows us to use the users' features as covariates to improve the performance, while  BPR uses latent features.

\chapter{Conclusions}
\label{sec:Conclusions}
In this tutorial, we provided an overview of object and label-preference learning using Gaussian Processes. We delved into nine models in detail, exploring their approaches to modelling rational preferences and addressing challenges to deal with irrational preferences such as noisy utilities, comparisons among alternatives with similar utilities, and the presence of conflicting utilities. Lastly, we examined the ``duality'' between preferences and choices, and reviewed a model designed to learn directly from choice data. We showed the application of these models to different datasets and also provided a connection with recommender systems and skill-based ranking systems.

We quickly mention here an important application of GP based models: Bayesian optimisation. Preferential Bayesian optimisation has been applied to preference data, as shown in several recent papers (see, e.g., \citet{gonzalez2017preferential}, \citet{benavoli2021preferential}, \citet{benavoli2023d,takeno2023towards}, \citet{Adachi_etal_2024_LoopExplain}), however research on Bayesian optimisation methods is still very active and work is still required to tailor standard acquisition function to this setting. 

The models shown in this tutorial show how to apply Gaussian process models to many preference learning settings. An interesting research direction to be explored is how to learn from a mix of preference data and other types of data sources. For example, a mixed likelihood was proposed in \citet{benavoli2021} for data coming from evaluations of the utility function and erroneous preferences. Often an objective evaluation of the utility function is not available, however the users' response times \citep[see, e.g.,][]{Ratcliff_McKoon_2008} could be recorded.  Another direction to improve preference based approaches could be to include this data into the learning process. In certain applications, it is desirable to enforce additional consistency properties on preference and choice data, such as monotonicity and convexity \citep{benavolilinearly2024}. These properties have been extensively investigated in economics, decision theory and probability theory \citep{kreps1990course,de2019interpreting,seidenfeld2010coherent,van2018coherent}. Future research directions include the extension of the models presented in this tutorial to include these other sources of information.

\section*{Acknowledgements}
This work is partially funded by the Swiss National Science Foundation (SNF),
Switzerland, grant 200021\_212164/1. This project has received funding from the
European Union’s Horizon Europe Research and Innovation Framework under grant
agreement No. 101160720. This publication has emanated from research supported
in part by a research grant from Research Ireland under the US-Ireland R\&D
Partnership Programme Grant Number 24/US/3978.

\appendix

\chapter{}
\section{Proof of Proposition \ref{prop:gumbel} and remarks}
\label{app:gumbel}
Consider the noisy utility model:
\begin{equation}
    \label{eq:noise_thurapp}
    \widetilde{u}_i = u_i + v_i,
\end{equation}
fro $i=1,\dots,d$, where we have removed the dependence on the covariates to simplify the notation.
Assume that $v_i$ is Gumbel distributed with location $0$ and scale $1$, that is its cumulative distribution is $F(v_i)=e^{-e^{-v_i}}$ and its probabiltiy density function is $p(v_i)=e^{-v_i-e^{-v_i}}$. Consider the quantity
$$
s = \arg \max_{i=1,\dots,d} u_i + v_i, ~~~v_i \sim \text{Gumbel}(0,1),
$$
then we have that
$$
\begin{aligned}
P(S=s)&=P(\tilde{u}_s>\tilde{u}_i ~\forall i\neq s)=\int_{-\infty}^{\infty} P(\tilde{u}_s>\tilde{u}_i ~\forall i\neq s |\tilde{u}_s)p(\tilde{u}_s) d\tilde{u}_s\\
&=\int_{-\infty}^{\infty} \prod_{i\neq s} P(\tilde{u}_s>\tilde{u}_i |\tilde{u}_s)p(\tilde{u}_s) d\tilde{u}_s=\int_{-\infty}^{\infty} \prod_{i\neq s} e^{-e^{u_i-\tilde{u}_s}} p(\tilde{u}_s) d\tilde{u}_s\\
&=\int_{-\infty}^{\infty} e^{-\sum_{i\neq s}  e^{u_i-\tilde{u_s}}} e^{u_s -\tilde{u_s}-e^{u_s - \tilde{u_s}}} d\tilde{u_s}=\int_{-\infty}^{\infty} e^{-\sum_{i}  e^{u_i-\tilde{u_s}}} e^{u_s -\tilde{u_s}} d\tilde{u_s}\\
&=\int_{-\infty}^{\infty} e^{-e^{-\tilde{u_s}}\sum_{i}  e^{u_i}} e^{u_s} e^{-\tilde{u_s}} d\tilde{u_s}\\
\end{aligned}
$$
Now if we define $c =\sum_{i}  e^{u_i}$ then we have that
$$
\begin{aligned}
P(S=s)&=e^{u_s}\int_{-\infty}^{\infty} e^{-ce^{-\tilde{u_s}}}  e^{-\tilde{u_s}} d\tilde{u_s}=e^{u_s}\int_{-\infty}^{\infty} e^{-\tilde{u_s}-ce^{-\tilde{u_s}}}   d\tilde{u_s}\\
&=\frac{e^{u_s}}{c}=\frac{e^{u_s}}{\sum_{i}  e^{u_i}}.
\end{aligned}
$$
This proves that
\begin{equation}
p(\widetilde{u}_s = \max(\widetilde{u}_1,\dots,\widetilde{u}_d))=\frac{a_s}{\sum_{j=1}^d a_j},    
\end{equation}
with $a_i=e^{u_i}$. 

It is worth noticing that,  under normal distributed $v_i$, an analogous formula does not hold  and also Luce’s axiom of Choice is not satisfied.

\section{Skew Gaussian Process}
\label{app:skewgp}
\begin{figure}[htp]
	\centering
	\begin{tabular}{cc}
		\includegraphics[width=0.49\linewidth]{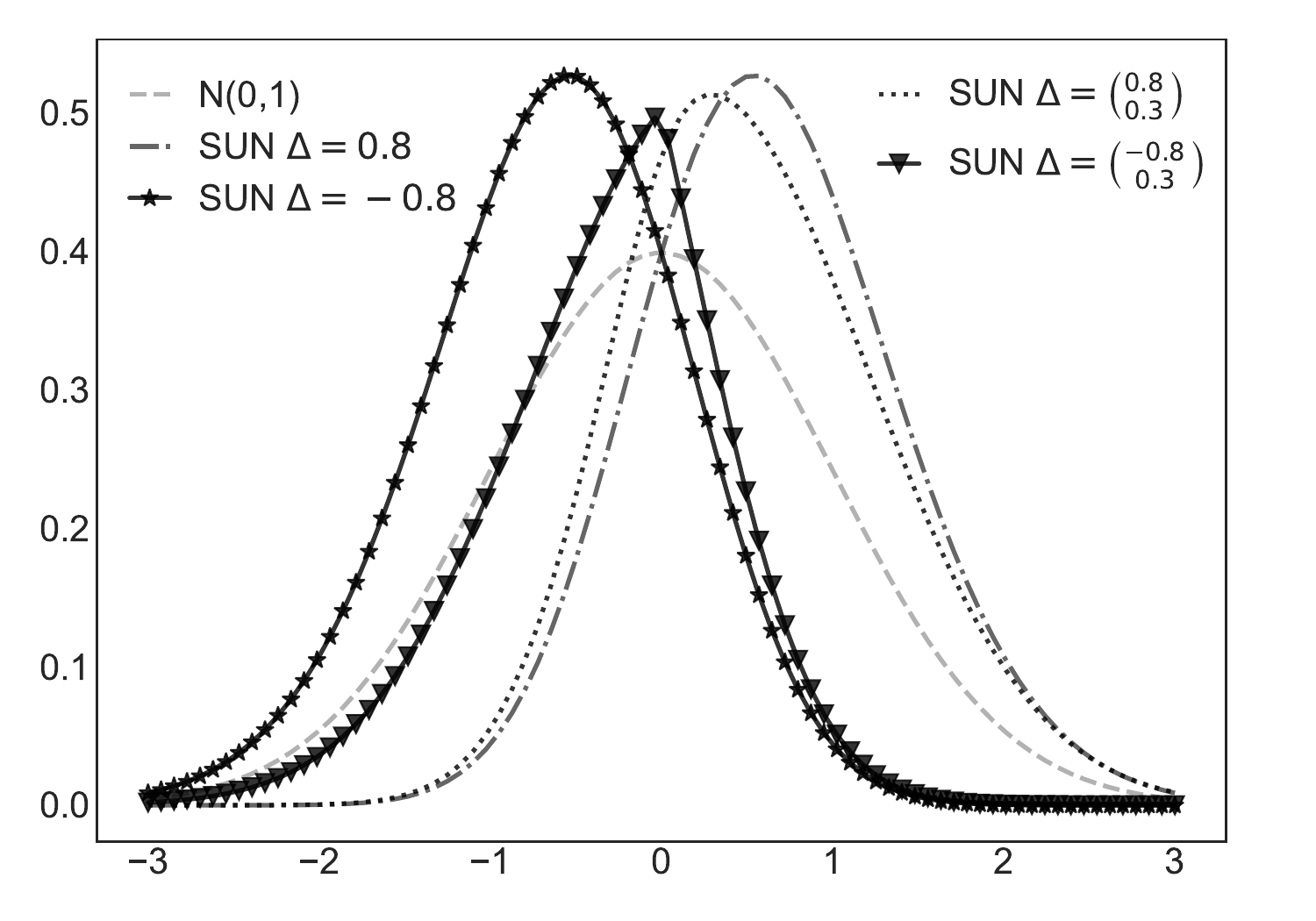} &
\includegraphics[width=.49\linewidth]{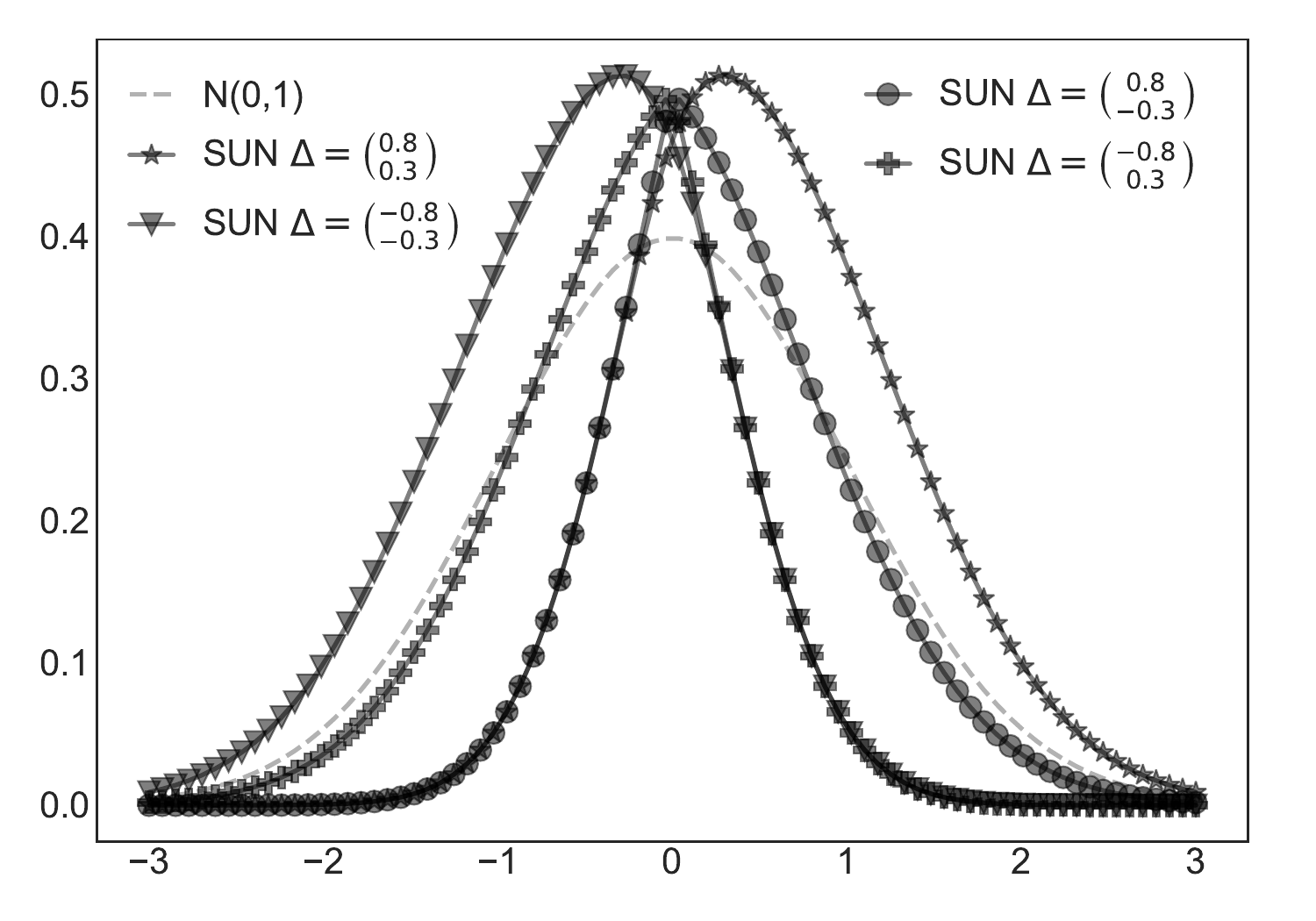} \\
		\small $s=1$, $\Gamma=1$  & \small $s=2$, $\Gamma_{0,1}=\Gamma_{1,0}=0.8$.
	\end{tabular}
	\caption{Density plots for $\text{SUN}_{1,s}(0,1,\Delta,\gamma,\Gamma)$. For all plots $\Gamma$ is a correlation matrix, $\gamma = 0$, the dashed lines without markers represent the density of $y \sim N_1(0,1)$.}
	\label{fig:SUN1d}
\end{figure}

The unified skew-normal distribution \citep{arellano2006unification,azzalini2013skew,durante2018conjugate,alodat2020gaussian,anceschi2023bayesian}  generalises the normal distribution by introducing skewness. A skew-normal distrbution can be built from  the marginal of a truncated multivariate normal, see \cite[Ch.7]{azzalini2013skew}. Consider two vectors ${\bf z}_0\in \mathbb{R}^{s},{\bf z}_1\in \mathbb{R}^{p}$ such that:
\begin{align}
\label{eq:jointnormal}
\begin{bmatrix}
{\bf z}_1\\
{\bf z}_0
\end{bmatrix}\sim N({\bf 0}_{s+p}, M), ~M=\begin{bmatrix}
\Omega  & \Delta\\
\Delta^\top & \Gamma
\end{bmatrix},
\end{align}
where $M$ is a full-rank covariance matrix. We then define $\boldsymbol{\zeta}$ to be distributed as $({\bf z}_1 |{\bf z}_0 + \bgamma > {\bf 0}_{s})$, where $\bgamma \in \mathbb{R}^s$  and the inequality
${\bf z}_0 + \bgamma > {\bf 0}_{s}$ is intended to be component-wise.\footnote{In the standard construction of the SUN distribution \cite[Ch.7]{azzalini2013skew}, the matrix $M$ is a correlation matrix. However, we can obtain the standard construction from \eqref{eq:jointnormal} by a change of variables.}
Then the  vector ${\bf z}=\bxi+\boldsymbol{\zeta}\in \mathbb{R}^{p}$ is distributed as a \textit{multivariate
unified skew-normal distribution} with latent skewness dimension $s$. We denote it as $ \bz \sim \text{SUN}_{p,s}(\bxi,\Omega,\Delta,\bgamma,\Gamma )$ and its Probability Density Function (PDF) is given by:
\begin{align}
\label{eq:sun}
p(\bz) &= \phi_{p}(\bz-\bxi;\Omega)\frac{\Phi_s\left(\bgamma+\Delta^\top{\Omega}^{-1}(\bz-\bxi);\Gamma-\Delta^\top{\Omega}^{-1}\Delta\right)}{\Phi_s\left(\bgamma;\Gamma \right)}, 
\end{align}
where $\phi_p(\bz-\bxi;\Omega)$ denotes the PDF of a multivariate normal distribution with mean $\bxi \in \mathbb{R}^{p}$ and covariance $\Omega\in \mathbb{R}^{p\times p}$. $\Phi_s({\bf a};M)$ represents the Cumulative Distribution Function (CDF) of a $s$-dimensional multivariate normal distribution with zero mean and covariance matrix $M$ evaluated at ${\bf a}\in \mathbb{R}^s$. 
The following parameters $\bgamma \in \mathbb{R}^s, \Gamma \in \mathbb{R}^{s\times s},\Delta \in \mathbb{R}^{p \times s}$ control the skewness of the distribution. In particular, the parameter $\Delta$ is called \textit{skewness matrix}. When $\Delta=0$, eq.~\eqref{eq:sun}  reduces  to  $\phi_p(\bz-\bxi;\Omega)$, i.e. a skew-normal with zero skewness matrix is a normal distribution. Moreover, we assume that $\Phi_0(\cdot)=1$, so that, for $s=0$, eq.~\eqref{eq:sun} becomes a multivariate normal distribution. Figure~\ref{fig:SUN1d} shows the density of a univariate SUN distribution with latent dimensions $s=1$ and $s=2$.  

A SkewGP \citep{benavoli2020skew, benavoli2021} is a generalisation of a SUN distribution to a stochastic process which becomes a GP when skewness is zero. To define a SkewGP, we consider  a location function $\xi: \mathbb{R}^d \rightarrow \mathbb{R}$, a scale (kernel) function $\Omega: \mathbb{R}^d \times \mathbb{R}^d \rightarrow \mathbb{R}$, a skewness vector function $\Delta: \mathbb{R}^d \rightarrow \mathbb{R}^{s}$ and the parameters $\bgamma \in \mathbb{R}^s, \Gamma \in \mathbb{R}^{s \times s}$. 
A real function $f: \mathbb{R}^d \rightarrow \mathbb{R}$ is said to be SkewGP-distributed with latent dimension $s$, if for any sequence of $n$ points $\bx_1, \ldots, \bx_n \in \mathbb{R}^d$, the vector $[f(\bx_1), \ldots, f(\bx_n)]^\top \in \mathbb{R}^n$ is SUN distributed with parameters $\bgamma, \Gamma$ and location, scale and skewness matrices,
respectively, given by
\begin{equation}
\begin{array}{rl}
\xi(X)&=\left[\begin{smallmatrix}
\xi(\bx_1), \xi(\bx_2),\dots, \xi(\bx_n)
\end{smallmatrix}\right]^\top,\vspace{0.1cm}\\
\Omega(X,X)&=\left[
\begin{smallmatrix}
\Omega(\bx_1,\bx_1) & \Omega(\bx_1,\bx_2) &\dots & \Omega(\bx_1,\bx_n)\\
\Omega(\bx_2,\bx_1) & \Omega(\bx_2,\bx_2) &\dots & \Omega(\bx_2,\bx_n)\\
\vdots & \vdots &\dots & \vdots\\
\Omega(\bx_n,\bx_1) & \Omega(\bx_n,\bx_2) &\dots & \Omega(\bx_n,\bx_n)\\
\end{smallmatrix}\right],\vspace{0.1cm}\\
\Delta(X)&=\left[\begin{smallmatrix}\Delta(\bx_1),\Delta(\bx_2),\dots, \Delta(\bx_n)\\
\end{smallmatrix}\right],
\end{array}
\end{equation}
where $X=[\bx_1, \bx_2, \dots, \bx_n]^\top$.  

In this case, we write $f({\bf x}) \sim \text{SkewGP}_s(\xi({\bf x}), \Omega({\bf x},{\bf x}),\Delta({\bf x}),\bgamma, \Gamma)$.
SkewGPs are conjugate with both the normal and affine probit likelihood and, more in general, with their product. This allows us  to derive their posterior for nonparametric regression, classification, preference learning and mixed problems.

In particular,  consider the affine-probit-normal product  likelihood:  
\begin{equation}
\begin{aligned}
p(Y,Z,W \mid f(X)) &= \Phi_{m_a}(Z+Wf(X); \Sigma) \phi_{m_r}(Y-Cf(X);R).  
\end{aligned}
\label{eq:mixedlike}
\end{equation}
where $m_r$ (the subscript $r$ stands for regression) denotes the number of regression-type observations  and $m_a$ the number of binary/preference-type observations (the subscript $a$ stands for affine). Therefore, we have that
$Y \in \mathbb{R}^{m_r}, C \in \mathbb{R}^{m_r \times n}, W \in \mathbb{R}^{m_a \times n}, Z \in \mathbb{R}^{m_a \times 1}$. The matrices $ \Sigma \in \mathbb{R}^{m_a \times m_a},R \in  \mathbb{R}^{m_r \times m_r}$ are covariance matrices.  
This likelihood encompasses all the standard likelihood functions used in regression, classification and preference-learning. 
For instance, a standard regression  is obtained by setting $C=I_{m_r}$,  $R=\sigma^2 I_{m_r}$ and $m_a=0$; classification is obtained for $W=diag(2Y-1)$, $Z={\bf 0}_{m_a}$, $\Sigma=I_{m_a}$ and $m_r=0$, where $Y$ is the vector containing the observed class values $Y_i \in\{0,1\}$. 
Preference learning is obtained with $Z={\bf 0}_{m_a}$, $\Sigma=I_{m_a}$ and $m_r=0$ and $W \in \mathbb{R}^{m_a \times n}$   whose s-th row is all zero apart from $W_{i}=1,W_{j}=-1$ if the data includes the preference ${\bf x}_i\succ {\bf x}_j$. 

We  now include this result from \cite[Theorem 3]{benavoli2021}.

\begin{proposition}
	\label{prop:postmixed}	
	Let us assume a SkewGP prior, \\ i.e.~$f({\bf x}) \sim \text{SkewGP}_s(\xi({\bf x}), \Omega({\bf x},{\bf x}),\Delta({\bf x}),\bgamma, \Gamma)$, the likelihood \eqref{eq:mixedlike}, then a-posteriori $f({\bf x})$ is SkewGP with mean, covariance and skewness functions:
 \vspace{-0.15cm}
   \begin{align}
   \nonumber
	\tilde{\bxi}({\bf x})  &=\bxi({\bf x})+\Omega({\bf x},X) C^T(C\Omega(X,X) C^T+R)^{-1}(Y-C\xi(X)),\\
 \nonumber
	\tilde{\Omega}({\bf x},{\bf x}) &= \Omega({\bf x},{\bf x})
 -\Omega({\bf x},X) C^T(C\Omega(X,X) C^T+R)^{-1}C\Omega(X,{\bf x}),\\
	\nonumber
\tilde{\Delta}({\bf x}) &=\begin{bmatrix}
\Delta(\bx)~ & \Omega(\bx,X)W^T
\end{bmatrix}\\ \nonumber
&-\Omega({\bf x},X)C^T (C\Omega(X,X) C^T+R)^{-1} C 
 \begin{bmatrix}
\Delta(X)~ & \Omega(X,X)W^T
\end{bmatrix},\\
\nonumber
\tilde{\bgamma} &= \bgamma_p+\begin{bmatrix}
\Delta(X)~ & \Omega(X,X)W^T
\end{bmatrix}^T {\Omega}(X,X)^{-1}(\tilde{\bxi}(X)-\bxi(X))\\
\nonumber
\tilde{\Gamma} &=  \Gamma_p- \begin{bmatrix}
\Delta(X)~ & \Omega(X,X)W^T
\end{bmatrix}^T\Omega^{-1}(X,X)\begin{bmatrix}
\Delta(X)~ &\Omega(X,X)W^T
\end{bmatrix}\\ \nonumber
&~~~+\Delta_p^T\tilde{\Omega}(X,X)^{-1}\Delta_p,\\
\nonumber
\Delta_p&=\tilde{\Omega}(X,X){\Omega}^{-1}(X,X)\begin{bmatrix}
\Delta(X)~ & \Omega(X,X)W^T
\end{bmatrix},\\
\nonumber
\bgamma_p &=[\bgamma,~~Z+W\xi(X)]^T, \\
\nonumber
\Gamma_p&=\begin{bmatrix}
	\Gamma & ~~\Delta(X)^T  W^T \\
	W  \Delta(X) & ~~(W \Omega(X,X) W^T + \Sigma) \end{bmatrix}.
\end{align}
\end{proposition}
The computation of predictive inference (posterior mean, credible intervals etc.) can be achieved by sampling. 
 Note that \cite[Ch.7]{azzalini2013skew} that $\mathbf{z} \sim \operatorname{SUN}_{p,s}(\bxi,\Omega, \Delta, \bgamma, \Gamma)$ can be written as $\mathbf{z} = \bxi + \mathbf{r}_0 + \Delta \Gamma^{-1}\mathbf{r}_{1,-\gamma}$ with $\mathbf{r}_0\sim \phi_p(0; \bar{\Omega}-\Delta \Gamma^{-1}\Delta^T)$ and $\mathbf{r}_{1,-\gamma}$ is the truncation below $\gamma$ of $\mathbf{r}_{1} \sim \phi_s(0;\Gamma)$. Samples of the vector $\mathbf{r}_0$ can be achieved efficiently with standard methods, and  $\mathbf{r}_{1,-\gamma}$  can be obtained efficiently using the methods discussed in Section \ref{sec:consistent}. Similarly to GPs, the functions and matrices defining a SkewGP, $\text{SkewGP}_s(\xi({\bf x}), \Omega({\bf x},{\bf x}),\Delta({\bf x}),\bgamma, \Gamma)$ may depend on hyperparameters  ${\boldsymbol \theta}$. These parameters  are usually estimated by maximising the marginal likelihood, which is equal to:
 \begin{equation}
  \label{eq:ml_normal_mix}
  p(Y)=\phi_{m_r}(Y-C\xi(X);C\Omega(X,X)C^T+R)\frac{ \Phi_{s+m_a}(\tilde{\bgamma};~\tilde{\Gamma})}{\Phi_{s}(\bgamma;~\Gamma)},
 \end{equation}
 with $\tilde{\bgamma},\tilde{\Gamma}$ are defined in  Proposition \ref{prop:postmixed}. This involves the computation of a high-dimensional multivariate CDFs $ \Phi_{s+m_a}(\cdot), \Phi_{s}(\cdot)$. In this tutorial, we use the Laplace's approximation to approximate this integral.

\section{PrefGP library}
\label{app:GPpref}
This tutorial is supported by a Python software library, \textit{prefGP} \citep{prefGP}, that implements Models 1-9 and reproduces the results presented in the paper via notebooks. Further models can be developed by modifying the likelihood, as the inference process relies on generic Laplace's and Variational approximations (implemented using either Jax or PyTorch).

\bibliographystyle{plainnat}
\bibliography{biblio}

\end{document}